\documentclass{article}

\usepackage{PRIMEarxiv}

\usepackage[utf8]{inputenc} 
\usepackage[T1]{fontenc}    
\usepackage{hyperref}       
\usepackage{url}            
\usepackage{booktabs}       
\usepackage{float}
\usepackage{amsfonts}       
\usepackage{amsmath}
\usepackage{amssymb}
\usepackage{nicefrac}       
\usepackage{microtype}      
\usepackage{lipsum}
\usepackage{todonotes}
\usepackage{fancyhdr}       
\usepackage{graphicx}       
\usepackage{subcaption}
\graphicspath{{media/}}     
\usepackage{orcidlink}
\usepackage{algorithm}
\usepackage{algpseudocode}
\usepackage{soul}

\usepackage[pagewise]{lineno}
\usepackage{todonotes}
\nolinenumbers

\usepackage{amsthm}

\newtheorem{assumption}{Assumption}
\newtheorem{definition}{Definition}
\newtheorem{proposition}{Proposition}

\title{ENFORCE: Nonlinear Constrained Learning with Adaptive-depth Neural Projection
}

\author{
  Giacomo Lastrucci\orcidlink{0009-0005-3475-9351}\\
  Process Intelligence Research Group\\
  Department of Chemical Engineering\\
  Delft University of Technology\\
   \And
  Artur M. Schweidtmann\orcidlink{0000-0001-8885-6847}\\
  Process Intelligence Research Group\\
  Department of Chemical Engineering\\
  Delft University of Technology \\
  \texttt{a.schweidtmann@tudelft.nl} \\
}

\begin{document}
\maketitle

\begin{abstract}
Ensuring neural networks adhere to domain-specific constraints is crucial for addressing safety and trustworthiness while also enhancing inference accuracy. Despite the nonlinear nature of most real-world tasks, the majority of existing methods are limited to affine (equality) or convex (inequality) constraints. We introduce ENFORCE, a neural network architecture that uses an adaptive projection module (AdaNP) to enforce nonlinear equality and inequality constraints in the predictions up to a specified tolerance $\varepsilon$, and exactly in the affine-in-$y$ case.
For affine constraint sets, we prove that the associated projection mapping is non-expansive (1-Lipschitz), ensuring stable gradient propagation. For nonlinear constraints, we establish local convergence analysis under standard regularity conditions. We evaluate ENFORCE on multiple tasks, including function fitting, real-world engineering case studies, and learning optimization problems. For the latter, we introduce a class of scalable optimization problems as a benchmark for nonlinear constrained learning. In the benchmarks, the predictions of our architecture satisfy nonlinear equality and inequality constraints up to a prescribed tolerance $\varepsilon$, while maintaining scalability with tractable computational complexity at training and inference time.
\end{abstract}


\vspace{.25cm}
\makebox[\linewidth][c]{%
    \raisebox{-0.2\height}{\includegraphics[height=1em]{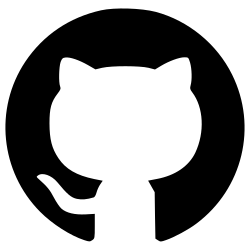}}%
    \hspace{1em}\url{https://github.com/process-intelligence-research/ENFORCE}%
    \hspace{2em}\raisebox{-0.2\height}{\includegraphics[height=1em]{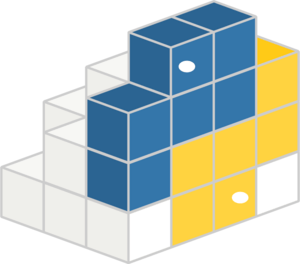}}%
    \hspace{1em}\texttt{pip install enforce-nn}%
}

\begin{figure}[H]
    \centering
    \includegraphics[width=0.75\textwidth]{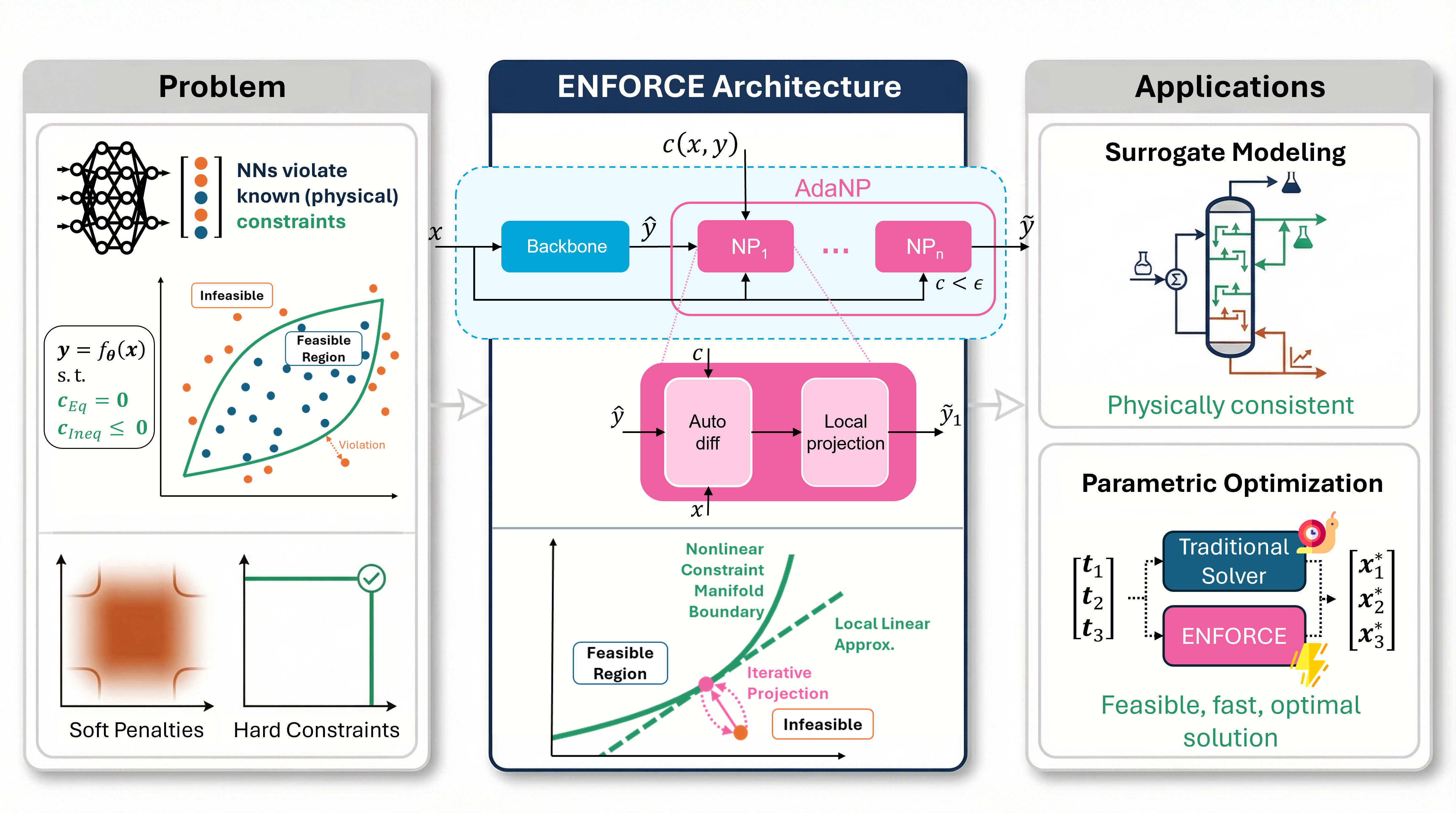}
    \caption{Neural network predictions may violate known (physical) constraints, leading to infeasible outputs. ENFORCE performs feasibility recovery through an adaptive-depth neural projection module that projects predictions toward the feasible manifold defined by nonlinear equality and inequality constraints, enabling hard-constrained learning for surrogate modeling and parametric optimization.}
    \label{fig:graph-abstr}
\end{figure}

\section{Introduction}
Neural networks (NNs) are the backbone of many recent advancements in artificial intelligence (AI), excelling in tasks such as natural language processing, image analysis, and scientific discovery due to their modularity, simplicity, and strong generalization capabilities. However, their ability falls short when strict adherence to domain-specific constraints is required. Depending on the task, prior knowledge about the system (e.g., from physics, safety, or operational constraints) is often available and is typically leveraged by humans in decision-making processes. In contrast, data-driven methods such as NNs rely solely on data. Thus, trained NNs may achieve strong predictive accuracy yet still violate known constraints, leading to inconsistent or infeasible outputs, thereby leading to unreliable decisions in downstream applications. Moreover, when domain knowledge is available as analytical equations, ensuring that NNs adhere to this information is crucial to avoid a suboptimal utilization of expert insights and potentially reduce data demand~\cite{ESamadi2022_trainingstrategyhybrid}. From a computational perspective, this problem can be interpreted as enforcing feasibility of learned predictions with respect to analytically defined constraint sets.\\
Enforcing strict constraints in NNs is a promising area of research for many fields. For example, in AI for Science, integrating first-principle laws ensures physically consistent models, enabling insightful scientific discovery~\cite{Wang2023_Scientificdiscoveryage, Xu2021_Artificialintelligencepowerful} or system modeling in engineering~\cite{Schweidtmann2024_reviewperspectivehybrida}. A prominent line of research focuses on leveraging NNs to safely accelerate computationally intensive tasks. Two major application areas have emerged: (i) the development of surrogate models that approximate complex high-fidelity simulations, thereby significantly reducing computational cost~\cite{Lastrucci2025_PicardKKThPINN} and (ii) learning to solve parametric constrained optimization problems (often referred to as proxy optimization), where the NN is trained to predict optimal solutions directly, either in a supervised or unsupervised manner~\cite{Kotary2021_EndEndConstrained, Donti2021_DC3learningmethod, DiVito2024_LearningSolveDifferential, Schweidtmann2018_DeterministicGlobalOptimization}. More broadly, constraint-aware prediction is critical in settings where learned models are embedded within decision-making pipelines. For instance, in engineering and scientific applications, physically inconsistent predictions can propagate infeasibility into downstream optimization or control problems. Ensuring that model outputs satisfy analytically defined constraints is therefore essential for reliable and computationally robust deployment.
Beyond feasibility at inference time, incorporating such structural knowledge can also improve learning behavior by reducing the hypothesis space, which may mitigate overfitting in data-scarce regimes~\cite{Min2024_HardConstrainedNeural}.\\
Enforcing constraints in NNs is not straightforward. The majority of existing approaches rely on incorporating penalty terms into the loss function to minimize constraint violations~\cite{Raissi2019_Physicsinformedneural}. Yet, these penalty-based methods offer no guarantees of constraint satisfaction (\textit{soft-constrained}). In contrast, other methods aim to ensure strict adherence to analytical constraints by design (\textit{hard-constrained}). For instance, one can use sigmoid functions to bound outputs. To enforce analytical constraints expressed as mathematical equations, recent studies incorporate feasibility-recovery operators in the form of (i) projection or repair layers, (ii) predict-and-complete (implicit) layers, or (iii) constrained optimization schemes applied during training. For a comprehensive literature review on hard-constrained neural networks, the reader is referred to Section~\ref{sec:related-work}. A variety of efficient methods have been proposed to enforce constraints defined by affine relationships between input and output variables or by convex regions~\cite{Chen2024_PhysicsInformedNeural, Chen2021_Theoryguidedhard, Min2024_HardConstrainedNeural}. However, although many real-world applications are inherently governed by nonlinear constraints~\cite{Mize2019_BestPracticesEstimating, Nicolis1995_IntroductionNonlinearScience}, this setting remains largely unexplored. Existing approaches for handling nonlinear constraints predominantly rely on repeatedly solving nonlinear systems or constrained optimization subproblems during training and inference~\cite{Donti2021_DC3learningmethod, Mukherjee2024_developmentsteadystate}. These methods can introduce significant computational overhead: implicit or root-finding layers solve a nonlinear system in each forward pass and differentiate through it during backpropagation, whereas solver-based approaches train the network via constrained optimization algorithms and often solve an additional subproblem at inference time, limiting scalability with respect to network size and constraint dimension.\\
We propose ENFORCE, a neural network architecture that integrates differentiable projection layers to drive predictions toward feasibility with respect to nonlinear equality and inequality constraints. ENFORCE is positioned within the class of projection-based feasibility restoration methods, embedding an SQP-inspired correction step directly into the forward pass of a learned model. The method is trained using standard unconstrained optimization techniques and leverages an adaptive-depth neural projection (AdaNP) module that iteratively linearizes and projects onto the resulting local approximation of the constraint manifold until a prescribed feasibility tolerance is reached. Although this work uses a neural network as the backbone predictor, the AdaNP module could potentially be coupled with other differentiable machine learning models. Unlike external solver-based approaches, ENFORCE performs feasibility recovery internally and in a computationally efficient manner, with tractable per-iteration complexity and depth-adaptive runtime. We evaluate ENFORCE on multiple problems, including real-world engineering problems and a scalable class of nonlinear parametric optimization problems that we introduce as a benchmark for nonlinear constrained learning.\\
Thus, the main contributions of this work are (i) the integration of adaptive projection within neural architectures to achieve $\varepsilon$-feasibility for nonlinear equality and inequality constraints, (ii) a theoretical analysis of stability in the affine case and local convergence in the nonlinear case, and (iii) an empirical evaluation on scalable and relevant constrained learning benchmarks.\\

\section{Related work}
\label{sec:related-work}
This section reviews different approaches to enforcing constraints in NNs, with a focus on existing hard-constrained methods.

\subsection{Soft-constrained neural networks}
One of the earliest approaches to embedding domain knowledge into NNs involves the use of \textit{soft} constraints. Soft constraints are incorporated as penalty terms appended to the loss function, penalizing residuals of algebraic~\cite{Erichson2019_PhysicsinformedAutoencoders, Pfrommer2020_ContactNetsLearningDiscontinuousa} or differential equations underlying the system~\cite{ Wang2021_Learningsolutionoperator}. Physics-informed Neural Networks (PINNs)~\cite{Raissi2019_Physicsinformedneural} represent a widely used framework designed to solve partial differential equations (PDEs) with deep learning by employing soft constraints and collocation points.\ Although the soft-constrained approach places no restrictions on the complexity of the constraints, it has the drawback of not guaranteeing strict adherence to them. Furthermore, the interplay of multiple loss terms -- particularly when they differ in scale or physical interpretation -- can complicate optimization and affect predictive accuracy~\cite{Wang2020_Understandingmitigatinggradient, Wang2020_WhenwhyPINNs}. Nevertheless, soft-constrained formulations have proven effective in practice and remain a dominant paradigm in scientific machine learning, particularly for solving PDEs and modeling complex physical systems.

\subsection{Hard-constrained neural networks}
\label{subsec:related-work-hard}
Hard-constrained neural networks refer to methodological approaches ensuring that neural network predictions adhere to analytical constraints by construction. These constraints, explicitly encoded within the architecture, act as inductive biases, guiding the learning process toward compliance with domain knowledge or restrictions~\cite{Karniadakis2021_Physicsinformedmachine}. Architectures such as convolutional neural networks (CNNs)~\cite{LeCun1989_BackpropagationAppliedHandwritten} and graph neural networks (GNNs)~\cite{Bronstein2017_GeometricDeepLearning, Wu2021_ComprehensiveSurveyGraph} encode inductive biases by guaranteeing invariance with respect to patterns and symmetries. Simple analytical constraints can be enforced using differentiable functions, such as sigmoids or ReLU for output bounding and softmax for simplex constraints. Recent literature includes significant contributions for enforcing analytical inequality constraints, such as convex polytopes and convex sets more generally~\cite{Frerix2020_HomogeneousLinearInequality, Donti2021_DC3learningmethod, Wang2024_LinSATNetPositiveLinear, Tordesillas2023_RAYENImpositionHard, Konstantinov2023_NewComputationallySimple}. One can also constrain the neural network to guarantee specific functional characteristics, such as Lipschitz continuity~\cite{Anil2018_SortingoutLipschitz} or Lyapunov stability~\cite {Manek2020_LearningStableDeepa}. Nevertheless, this falls outside the scope of this work. Since this paper focuses on analytical equality and inequality constraints, the following literature review considers existing methods for this specific case.

\paragraph{Projection methods}
Many methods for encoding hard equality constraints utilize projection techniques, which correct preliminary neural network predictions by appending non-trainable layers to the output. Projections can be formulated as optimization problems (i.e., distance minimization) or derived from geometric principles. For example, in~\cite{Chen2021_Theoryguidedhard} neural network predictions of physical systems governed by PDEs are projected to ensure solutions satisfy the finite difference discretization of the underlying linear PDEs. A more general approach is the KKT-hPINN~\cite{Chen2024_PhysicsInformedNeural}, which enforces linear equality constraints in the inputs and outputs. Recently, HardNet~\cite{Min2024_HardConstrainedNeural} was introduced to enforce equality and inequality constraints affine in the output, without input restrictions, via a closed-form projection step. Moreover, Iftakher et al.~\cite{Iftakher2025_PhysicsInformedNeural} proposed a method called KKTHardnet to enforce nonlinear constraints leveraging log-exponential reformulation and a Newton method.

\paragraph{Predict-and-complete}
NNs can also predict a subset of output variables, $y_P \in \mathbb{R}^{N_O - N_{Eq}}$, where $N_O$ is the dimension of the target vector and $N_{Eq}$ is the dimension of the equality constraint system, to then complete the prediction by solving the system of constraints based on this partial output (null-space methods). This approach ensures that the constraints are always satisfied. For instance, Beucler et al. introduced this concept to simulate physical systems such as climate modeling~\cite{Beucler2019_EnforcingAnalyticConstraints}. However, when the constraints are not available in explicit form, solving the system requires a root-finding solver. Similar approaches have been proposed within the hybrid modeling community, particularly in the \textit{serial} configuration, where a fully data-driven method is used to predict unknown inputs to a mechanistic model~\cite{Schweidtmann2024_reviewperspectivehybrida}. While studies like DC3~\cite{Donti2021_DC3learningmethod} have developed efficient backpropagation techniques, scenarios involving implicit nonlinear constraints can be computationally expensive to tackle with predict-and-complete methods. Moreover, predict-and-complete approaches can suffer training instabilities and result in decreased prediction accuracy~\cite{Beucler2019_EnforcingAnalyticConstraints}.

\paragraph{Constrained optimization}
To enforce analytical constraints, researchers leveraged constrained optimization to deploy specialized layers or directly train the neural network. OptNet~\cite{Amos2017_OptNetDifferentiableOptimization} is an optimization layer developed to solve quadratic programs. Angrawal et al.~\cite{Agrawal2019_DifferentiableConvexOptimization} expand the methodology to convex programs. They develop efficient differentiation techniques through such layers. Min et al.~\cite{Min2024_HardConstrainedNeural} leveraged such optimization layers to develop HardNet-Cvx, a neural network enforcing convex constraints. However, the forward pass always requires the solution of a constrained optimization problem. Recently, Mukherjee and Bhattacharyya~\cite{Mukherjee2024_developmentsteadystate} approached the constrained learning paradigm by training a neural network using a constrained optimization solver such as IPOPT~\cite{Waechter2005_implementationinteriorpoint} instead of standard unconstrained optimization algorithms. However, these approaches pose severe limitations in terms of NNs and dataset size.

\paragraph{Other methods}
Other methods have been proposed for constrained learning in NNs, mostly considering affine or convex regions~\cite{Tao2023_ArchitecturePreservingProvable, Tao2024_ProvableEditingDeep}. Many of them consider constraints only dependent on the input of the NN~\cite{Schweidtmann2021_Obeyvaliditylimits, Tordesillas2023_RAYENImpositionHard, Balestriero2022_POLICEProvablyOptimal, Brosowsky2020_SampleSpecificOutput}, others design strategies to include the dependence on both inputs and outputs~\cite{Konstantinov2023_NewComputationallySimple, Lastrucci2025_PicardKKThPINN}. Recently, contributions to enforce general logic and linear constraints have been proposed by the neuro-symbolic AI community, developing loss terms or constraining layers using logic programming~\cite{Giunchiglia2021_MultiLabelClassification, Stoian2024_HowRealisticIs,Fischer2019_DL2TrainingQuerying}. 

Overall, to the best of our knowledge, the only methods aiming to enforce nonlinear equality and inequality constraints involving both the input and output of a neural network are DC3~\cite{Donti2021_DC3learningmethod} (using predict-and-complete method) and KKT-Hardnet~\cite{Iftakher2025_PhysicsInformedNeural} (using a projection layer with a Newton solver). In this work, we compare our method with the state of the art in terms of accuracy, constraint feasibility, and computational efficiency, and identify the scenarios in which our method is most suitable.

\section{Preliminaries}
\paragraph{Problem statement}
Given a dataset $(x^*_i, y^*_i)_{i=1,...,N}$, without loss of generality we consider a neural network $f_{\theta}$ with parameters $\theta$ to approximate the underlying relationships while satisfying a set of known algebraic equality constraints $c(x,y)=0$. In general, $c$ can be a nonlinear function in the input $x$ and output $y$ of the neural network, incorporating domain knowledge or specifying critical requirements. 

\begin{assumption}
i) The constraints $c(x,y)=0$ are feasible and linearly independent, ii) $N_{Eq} < N_O$, where $N_{Eq}$ is the number of equality constraints and $N_O$ is the output dimensionality of the neural network, i.e., there are available degrees of freedom to learn.
\end{assumption}

One way to enforce the neural network prediction $\hat{y}$ to satisfy the constraints is to project it onto the feasible hypersurface (manifold) defined by $c(x,y)=0$. The projection operation can be defined as an optimization problem:

\begin{equation}
\label{eq:weighted_nonlinear_projection}
\tilde{y} = \operatorname*{arg\,min}_{y} \frac{1}{2} (y - \hat{y})^T W (y - \hat{y}) \quad
\text{s.t.} \quad  c(x, y) = 0
\end{equation}

If $W$ is the identity matrix, the prediction is corrected by an orthogonal projection onto the feasible region. This can be interpreted as finding the feasible solution that minimizes the Euclidean distance from the original prediction $\hat{y}$. A local solution to the nonlinear program in Eq.\eqref{eq:weighted_nonlinear_projection} can be found by solving the first-order necessary optimality conditions, known as Karush–Kuhn–Tucker (KKT) conditions (cf. Chapter 12.3 in~\cite{Nocedal2006_NumericalOptimization}). However, the latter is not necessarily straightforward as it may involve solving a system of nonlinear equations.

\paragraph{Quadratic projection}
When $c(x, y)$ is an affine function in the neural network input and output, then the problem results in a quadratic program (QP) and a closed-form analytical solution is available for the KKT conditions~\cite{Chen2024_PhysicsInformedNeural}.
An extension to the closed-form is available when generalizing to any function in the input $x$, as the projection operation is still a QP. Consider an affine constraint on $y$ of the form $c = C(x)y - v(x) - b = 0$, where $C(x)$ and $v(x)$ act as the linear coefficient matrix and translation vector, respectively. For a given input $x_i$ and prediction $\hat{y}_i$, any function of $x_i$ can be treated as constant with respect to the optimization problem in Eq.\eqref{eq:weighted_nonlinear_projection}, which thus reduces to a QP.\\
Enforcing affine functions is not new and is also achieved through other techniques \cite{Min2024_HardConstrainedNeural, Balestriero2022_POLICEProvablyOptimal}. However, relaxing the assumption to allow for nonlinear constraints of both the input and output of the neural network commonly results in decreased computational efficiency and stability, as it typically requires the use of constrained optimization~\cite{Mukherjee2024_developmentsteadystate} or root-finding solvers such as Newton's methods~\cite{Donti2021_DC3learningmethod}. This motivates the need for scalable and computationally efficient approaches to nonlinear constrained learning, which we address with the proposed ENFORCE architecture.

\section{ENFORCE: Methodology for nonlinear constrained learning}
We present ENFORCE, a framework for efficient nonlinear constrained learning. ENFORCE combines a backbone neural network with an adaptive-depth neural projection module (AdaNP) that iteratively applies a linearize-and-project correction for equality constraints $c(x,y)=0$ with $c\in\mathcal{C}^1$. Exact feasibility is obtained in one step for affine-in-$y$ constraints, whereas for nonlinear constraints the guarantee is local: under standard regularity assumptions and for sufficiently accurate backbone predictions, AdaNP reduces the constraint residual below a prescribed tolerance $\varepsilon$. 
We further extend the framework to handle inequality constraints $g(x,y)\le 0$ via a reformulation into an equivalent system of equalities in an augmented space, as detailed in Section~\ref{subsec:ineq_extension}.

\subsection{AdaNP: Adaptive-depth neural projection}
\label{subsec:adanp}
We locally approximate the nonlinear program in Eq.\eqref{eq:weighted_nonlinear_projection} and exploit the efficiency of quadratic projections to generalize the methodology to nonlinear constraints. Assuming $c$ is of class $\mathcal{C}^1$, we use first-order Taylor expansion to locally linearize the constraints around the neural network input $x_0$ and prediction $\hat{y}$:

\begin{equation}
\label{eq:Taylor}
c(x,y) \simeq c(x_0,\hat{y}) + \left.J_xc\right|_{x_0, \hat{y}} (x - x_0) + \left.J_yc\right|_{x_0, \hat{y}} (y - \hat{y}),
\end{equation}

where $J_xc$ and $J_yc$ are the Jacobian matrices with respect to the variable $x$ and $y$, respectively.
Since the neural network input is fixed for a given sample, the linearization is exact in $x$, thus, $x = x_0$. Considering orthogonal projection for notation simplicity, the nonlinear optimization problem in~Eq.\eqref{eq:weighted_nonlinear_projection} is locally approximated by the (linearly constrained) QP:

\begin{equation}
\label{eq:linearized_projection}
\tilde{y} = \operatorname*{arg\,min}_{y} \frac{1}{2} \| y - \hat{y} \|^2 \quad
\text{s.t.} \quad c(x,\hat{y}) + \left.J_yc\right|_{x, \hat{y}} (y - \hat{y}) = 0
\end{equation}


\begin{definition}[Projection operator $\mathcal{P}$]
\label{def:NPlayer}
Given an input $x \in \mathbb{R}^{N_I}$ to a neural network $f_{\theta}$, its prediction $\hat{y} = f_{\theta}(x) \in \mathbb{R}^{N_O}$, and a set of constraints $c\in\mathcal{C}^1(\Omega, \mathbb{R}^{N{_{Eq}}})$, with $N_{Eq}<N_O$, we define an operator $\mathcal{P}$ such that $\tilde{y}=\mathcal{P}(\hat{y})$ is the solution to the linearized quadratic program in Eq.\eqref{eq:linearized_projection}, in the domain $\Omega$ where the constraints are defined.\\
In particular,
\[
\tilde{y}=B^*\hat{y}+v^*,
\]
with
\[
B^* = I - B^T(BB^T)^{-1}B,
\]
\[
v^* = B^T(BB^T)^{-1}v,
\]
where
\[
I\in \mathbb{R}^{N_O \times N_O},
\]
\[
B = \left.J_yc\right|_{x, \hat{y}},
\]
\[
v = \left.J_yc\right|_{x, \hat{y}} \hat{y} - c(x, \hat{y}).
\]
\end{definition}

The projection operator defined above possesses favorable stability properties. 
In particular, when the constraint set is affine, the projection mapping is non-expansive.

\begin{proposition}[Non-expansiveness for affine constraint sets]
\label{prop:nonexpansive_affine}
Consider the projection operator
\begin{equation}
\tilde{y} = \operatorname*{arg\,min}_{y} \frac{1}{2} \|y - \hat{y}\|^2 
\quad \text{s.t.} \quad c(y)=0,
\end{equation}
where $c(y)$ is affine in $y$, so that the feasible set 
$\mathcal{M} = \{y \in \mathbb{R}^{N_O} : c(y)=0\}$ 
is a convex affine subspace. 

Then the projection mapping $\mathcal{P}(\hat{y}) = \tilde{y}$ 
is non-expansive, i.e., $1$-Lipschitz:
\[
\|\mathcal{P}(\hat{y}_i) - \mathcal{P}(\hat{y}_j)\|
\le
\|\hat{y}_i - \hat{y}_j\|.
\]
\end{proposition}
Hence, for affine constraint sets, the projection operator is non-expansive. 
A characterization of the gradient flow through the projection layer is provided in Appendix~C.

\textbf{Remark (nonlinear constraints).}
For nonlinear constraints, the projection step corresponds to a projection onto a locally linearized approximation of the constraint manifold. 
In this case, global non-expansiveness does not generally hold. 
Stability and convergence depend on standard regularity conditions (e.g., LICQ, smoothness) and on the proximity of the prediction to the constraint manifold.

Given the closed-form expression of the operator $\mathcal{P}$ derived in Appendix~\ref{app:clos_form}, we can define a differentiable \textit{neural projection} (NP) layer representing the operator $\mathcal{P}$. The forward and backward passes of an NP layer are computationally cheap (more details on implementation and computational cost are given in Appendix~\ref{app:implementation_details}). However, the operator $\mathcal{P}$ projects the neural network prediction onto a linear approximation of the nonlinear constraints (i.e., the tangent hyperplane). The error that we introduce is proportional to the projection displacement $e_{D}=||\tilde{y} - \hat{y}||$. From this consideration, it follows that (1) the error is mitigated as the projection displacement is small, i.e., the neural network prediction is sufficiently accurate, and (2) a single NP layer cannot ensure exact adherence to nonlinear constraints.
It is worth noting that a single NP layer guarantees strict satisfaction of equality constraints that are affine in $y$ and nonlinear in $x$, i.e., it efficiently enforces constraint classes considered in similar recent works \cite{Chen2024_PhysicsInformedNeural, Min2024_HardConstrainedNeural}.

To address the challenge of satisfying nonlinear constraints, we propose AdaNP: an adaptive-depth neural projection composition that, under certain conditions, enforces nonlinear constraint satisfaction to arbitrary tolerance $\varepsilon$. 
\begin{definition}[AdaNP module]
Given an operator $\mathcal{P}$ as defined in Def.~\ref{def:NPlayer}, AdaNP is a composition of $n$ operators $\mathcal{P}$, such that:
\begin{equation*}
    \text{AdaNP} = \mathcal{P}_1 \circ \dots \circ \mathcal{P}_n
\end{equation*}
\end{definition}

\begin{proposition}[Local convergence of AdaNP]
\label{prop:adanp_convergence}
Let $x$ be fixed and let $y^\star$ be a feasible point such that $c(x,y^\star)=0$.
Assume that $c$ is $\mathcal{C}^2$ in a neighborhood of $y^\star$ and that the constraint qualification LICQ holds at $y^\star$, i.e., $J_y c(x,y^\star)$ has full row rank.
If the initial prediction $\hat{y}$ is sufficiently close to $y^\star$, then the iterates
\[
\tilde{y}_n = (\mathcal{P}_1 \circ \dots \circ \mathcal{P}_n)(\hat{y})
\]
generated by AdaNP are well-defined and satisfy
\[
\|c(x,\tilde{y}_n)\| \to 0
\quad\text{with a local linear rate.}
\]
In particular, for any tolerance $\varepsilon>0$ there exists $n\in\mathbb{N}$ such that
$\|c(x,\tilde{y}_n)\|<\varepsilon$.
A sufficient local rate condition is given in Appendix~\ref{app:local_convergence_rate}.
\end{proposition}

\paragraph{Remark (scope of the guarantee).}
Proposition~\ref{prop:adanp_convergence} is a \emph{local} result: it requires the backbone prediction to lie in a neighborhood where the linearization is accurate and $J_yc$ is well-conditioned.
Global convergence from arbitrary initializations is not guaranteed; if the backbone prediction is too far from the constraint manifold, a nonzero residual may remain.
Empirically, we observe stable behavior across different tested regimes.

AdaNP is a differentiable stack of $n$-NP layers that can be composed on every neural network backbone. The depth $n$ adjusts adaptively during training and inference depending on the nonlinearities and the specified tolerance (cf. Algorithm~\ref{algorithm:adanp} for details on the adaptive behavior). Accurate NNs typically result in shallower AdaNP modules, since the linearization error $e_{D}$ is related to the distance $||\hat{y}-y^*||$ between the neural network prediction $\hat{y}$ and ground truth output $y^*$.
This introduces a trade-off between the complexity of the backbone and the required depth of AdaNP to satisfy the specified tolerance criteria.

Algorithm~\ref{algorithm:adanp} provides a high-level overview of the procedure underlying the AdaNP module. The depth of AdaNP (i.e., the number of projection iterations) adapts to satisfy the constraints requirements according to a specified tolerance $\varepsilon_t$.

\begin{algorithm}
\caption{AdaNP: Adaptive-depth Neural Projection}
\label{algorithm:adanp}
\begin{algorithmic}[1]
\State \textbf{Input:} input $x$, preliminary prediction $\hat{y}$, constraints $c$, tolerance $\varepsilon_t$, maximum depth $d_{\text{max}}$
\State \textbf{Initialize:} depth counter $i \gets 0$
\While{$m(c(x, \hat{y})) > \varepsilon_t$ \textbf{and} $i < d_{\text{max}}$}
    \State \textbf{Compute constraints Jacobian:} $J_{y}c$
    \State \textbf{Compute:} $B = \left.J_yc\right|_{x, \hat{y}}$
    \State \textbf{Compute:} $v = B \hat{y} - c(x, \hat{y})$
    \State \textbf{Compute:} $B^* = I - B^T(BB^T)^{-1}B$
    \State \textbf{Compute:} $v^* = B^T(BB^T)^{-1}v$
    \State \textbf{Project:} $\tilde{y} = B^* \hat{y} + v^*$
    \State \textbf{Update:} $\hat{y} = \tilde{y}$
    \State \textbf{Increment:} $i \gets i + 1$
\EndWhile
\State \Return $\tilde{y}$
\end{algorithmic}
\end{algorithm}
Here, $m(c(x, \hat{y}))$ represents some measure of the constraint residual, where $m(\cdot)$ can be the \textit{max} or the \textit{mean} operator.

\paragraph{Analogy with Sequential Quadratic Programming}
AdaNP can also be seen as an iterative method that recursively improves the solution of a linearized nonlinear program. Here, we notice the similarity to sequential quadratic programming (SQP) techniques. Specifically, AdaNP is a simple case of SQP method for which the objective function is naturally quadratic while the nonlinear constraints are linearized (in contrast to full SQP, in which the objective function is quadratically approximated). This observation allows to analyze the convergence rate of the method starting from SQP theory~\cite{Nocedal2006_NumericalOptimization, Fletcher2002_Nonlinearprogrammingpenalty, Fletcher2002_GlobalConvergenceFilter}. The reader is referred to the Appendix~\ref{app:local_convergence_rate} for a complete discussion.

\paragraph{Deviation from Newton's method}
While the KKT conditions for a nonlinear program (Eq.\eqref{eq:weighted_nonlinear_projection}) can be more generally solved using Newton's methods, our method circumvents the computational overhead associated with calculating the Hessian matrix of the constraints (cf. Appendix~\ref{app:deviation_newton}) at the cost of reduced convergence rate (i.e., full Newton's method converges quadratically).

\subsection{Extension to inequality constraints}
\label{subsec:ineq_extension}
While ENFORCE is primarily introduced for equality constraints $c(x,y)=0$, many applications require predictions to comply with inequality constraints (e.g., bounds, envelopes). In this section, we outline a practical extension that addresses problems of the form:
\begin{equation}
\label{eq:proj_ineq_problem}
\tilde{y} \in \arg\min_{y \in \mathbb{R}^{N_O}} \frac{1}{2}\|y-\hat{y}\|^2
\quad \text{s.t.} \quad
c_{\mathrm{eq}}(x,y)=0, \qquad g(x,y)\le 0,
\end{equation}
where $g(x,y)\in\mathbb{R}^{N_{Ineq}}$ is a vector of inequality functions.\\
Importantly, this extension does not impose any additional limitation on the number of inequality constraints $N_{Ineq}$ beyond the standard requirement for equality constraints, i.e., $N_{Eq} \le N_O$.
In particular, we do not require the \emph{total} number of constraints $N_C = N_{Eq}+N_{Ineq}$ to be smaller than $N_O$, which is a restriction that appears in some alternative hard-constrained architectures for mixed constraint types~\cite{Min2024_HardConstrainedNeural}.\\
\paragraph{Fischer--Burmeister inequality reformulation}
A standard first-order characterization of Eq.~\eqref{eq:proj_ineq_problem} is given by the KKT conditions. Introducing multipliers $\lambda \in \mathbb{R}^{N_Ineq}$ for the inequalities, the KKT system includes
\begin{equation}
\label{eq:kkt_ineq}
g(x,y)\le 0,\qquad \lambda \ge 0,\qquad \lambda \odot g(x,y)=0,
\end{equation}
together with stationarity and the equality constraints. Inspired by Iftakher et al.~\cite{Iftakher2025_PhysicsInformedNeural}, the complementarity conditions in Eq.~\eqref{eq:kkt_ineq} can be expressed as a system of equalities using the Fischer--Burmeister (FB) function. For scalars $(a,b)$, define
\begin{equation}
\label{eq:fb_def}
\phi_{\mathrm{FB}}(a,b) := \sqrt{a^2+b^2+\varepsilon_{\mathrm{FB}}}-a-b,
\end{equation}
with a small regularization $\varepsilon_{\mathrm{FB}}>0$ for numerical stability near $(0,0)$.
It holds that $\phi_{\mathrm{FB}}(a,b)=0$ enforces $a\ge 0$, $b\ge 0$, and $ab=0$ (up to the regularization).

For each inequality $g_i(x,y)\le 0$, we set $a=\lambda_i$ and $b=-g_i(x,y)$ and impose
\begin{equation}
\label{eq:fb_ineq}
\phi_i(x,y,\lambda_i) := \phi_{\mathrm{FB}}(\lambda_i,-g_i(x,y)) = 0,
\end{equation}
which replaces the set of inequalities in Eq.~\eqref{eq:kkt_ineq}.\\
This converts the inequality constraints into equalities in an \emph{extended} variable space. Concretely, we define the extended output
\begin{equation}
\label{eq:y_ext}
y_{\mathrm{ext}} :=
\begin{bmatrix}
y \\ \lambda
\end{bmatrix}
\in \mathbb{R}^{N_O+N_{Ineq}},
\end{equation}
and an extended constraint mapping
\begin{equation}
\label{eq:c_ext}
c_{\mathrm{ext}}(x,y_{\mathrm{ext}})
=
\begin{bmatrix}
c_{\mathrm{eq}}(x,y) \\
\phi_{\mathrm{FB}}(\lambda,-g(x,y))
\end{bmatrix}
\in \mathbb{R}^{N_{Eq}+N_{Ineq}},
\end{equation}
where $\phi_{\mathrm{FB}}(\lambda,-g)$ is applied componentwise. After this reformulation, the projection step used by ENFORCE can be applied without modification by treating $c_{\mathrm{ext}}(x,y_{\mathrm{ext}})=0$ as an equality-only system and running AdaNP in the extended space.

\paragraph{Implementation note (extended outputs)}
In our implementation, the backbone network predicts only the original outputs $y$. The additional multiplier coordinates $\lambda$ are appended as zeros before the projection step. This avoids allocating trainable parameters for multipliers while still allowing the projection step to adjust $\lambda$ as needed in the extended space. The initialization choice is consistent with the KKT conditions at strictly feasible points, where inactive inequalities satisfy $\lambda_i=0$.


\subsection{Architecture and implementation}
\label{subsection:Architecture}
\begin{figure}[t]
    \centering
    \includegraphics[width=0.60\textwidth]{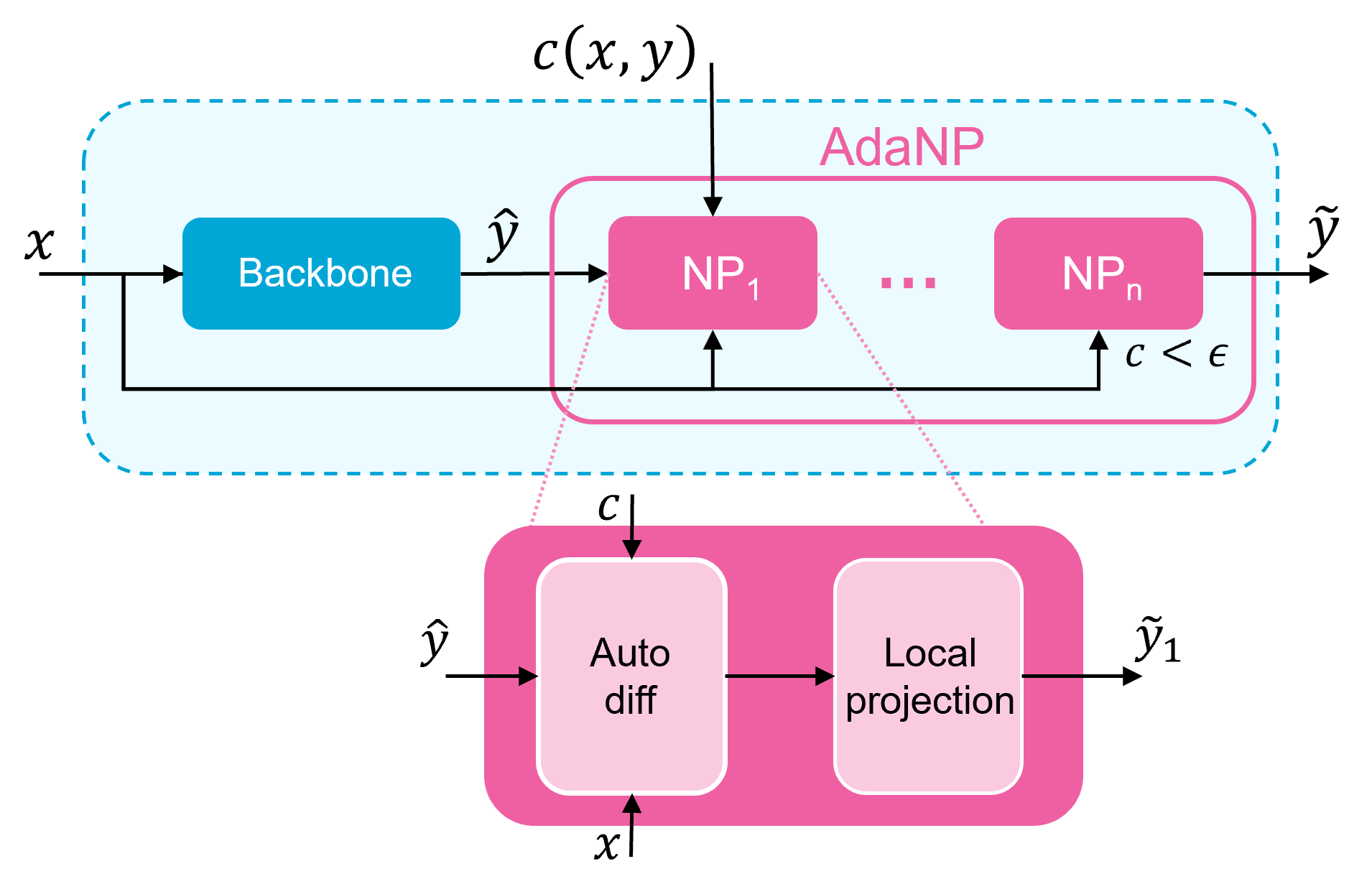}
    \caption{ENFORCE consists of a backbone neural network and an adaptive neural projection (AdaNP) module. The backbone network can be of every kind, such as fully connected, convolutional, or transformer architecture. AdaNP includes an adaptive number of neural projection (NP) layers, each composed of an auto-differentiation and a local projection step.}
    \label{fig:architecture}
\end{figure}

The architecture of ENFORCE (Fig.~\ref{fig:architecture}) is composed of (1) a neural network as backbone, which can be of any kind and complexity, and (2) an AdaNP module. Although beyond the scope of this work, we note that AdaNP is not inherently tied to neural network backbones and the same feasibility-recovery mechanism could be coupled with other differentiable machine learning models trained using standard gradient descent methods, providing a potential route toward constraint-aware machine learning models beyond neural architectures. The depth of AdaNP depends on the backbone performance and specified tolerance $\varepsilon$. Indeed, the tolerance of AdaNP can be tuned to increase training efficiency (cf. Section~\ref{sect:empirical_hyperp}). A single NP layer is composed of two steps: (1) automatic differentiation and (2) local neural projection.

\paragraph{Exact Jacobian computation} 
To compute the Jacobian of the constraint system, when not available analytically, we leverage automatic differentiation available in most deep learning libraries~\cite{paszke2019pytorchimperativestylehighperformance, tensorflow2015-whitepaper, jax2018_github}. Computing the local Jacobian $J_yc|_{x, \hat{y}}$ is computationally inexpensive, as it requires propagating derivatives only through the constraints and does not involve the neural network backbone. Furthermore, its computation can be efficiently parallelized on GPU.

\paragraph{Local neural projection}
The neural projection defined by the operator $\mathcal{P}$ in Def.~\ref{def:NPlayer} depends on individual input-prediction instances. Thus, the projection is locally defined in the neighborhood of ($x_i, \hat{y}_i$). We parallelize the computation of local neural projections using batched linear algebra operations by constructing a \textit{rank-3} tensor $\mathbf{B}$ and a \textit{rank-2} tensor $\mathbf{v}$ (Appendix~\ref{app:batch_local_projection}). This enables efficient implementation of the NP layer within standard stochastic gradient descent frameworks, with practical scalability in the batch dimension ($BS$). On modern hardware, handling up to $N_C < 10^3$ constraints results in a computational cost that remains practical, particularly when using Cholesky decomposition for matrix inversion~\cite{burden2005numerical}. Overall, the complexity of this method is \textit{equivalent} to other state-of-the-art methods such as DC3~\cite{Donti2021_DC3learningmethod} and KKT-Hardnet~\cite{Iftakher2025_PhysicsInformedNeural}. Additional insights into memory requirements are provided in Appendix~\ref{app:memory_footprint}.

\paragraph{Eigenvalue clamping for ill-conditioning in the projection step}
When constraints have disparate scales (e.g., different physical quantities or formulations), the projection step can encounter numerical ill-conditioning in finite-precision arithmetic. In ENFORCE, each linearized projection requires inverting a regularized Gram matrix of the form
\begin{equation}
\label{eq:gram_matrix}
G := B\,W^{-1}B^\top + \varepsilon I,
\end{equation}
where $B = J_y c_{\mathrm{ext}}(x,y_{\mathrm{ext}})$ is the constraint Jacobian in the (possibly extended) output space, $W\succ 0$ is the chosen weighting matrix (typically $W = I$), and $\varepsilon>0$ is a Tikhonov regularization parameter. In exact arithmetic, $G$ is symmetric positive definite; however, in \texttt{float32} computations, accumulation and scaling effects may produce small negative eigenvalues, causing Cholesky factorization to fail.

To make the projection robust, we implement the following solver strategy:
\begin{enumerate}
    \item attempt a Cholesky factorization of the symmetrized matrix $\tfrac{1}{2}(G+G^\top)$;
    \item if Cholesky fails for any batch element, fall back to an eigendecomposition
    $G = V\Lambda V^\top$ and clamp the spectrum from below,
    \begin{equation}
    \label{eq:eig_clamp}
    \Lambda \leftarrow \max(\Lambda,\varepsilon)\quad \text{(elementwise)},
    \end{equation}
    yielding the stabilized inverse $G^{-1} \approx V\Lambda^{-1}V^\top$.
\end{enumerate}
This procedure preserves the intended role of $\varepsilon$ as a minimum eigenvalue floor and avoids amplification of numerically ill-conditioned directions. Geometrically, clamping attenuates corrections along constraint directions that are poorly resolved by the local linearization (or by finite precision), resulting in a regularized projection step that remains well-defined even when $B$ exhibits large scale disparities.

\subsection{Training ENFORCE}
\label{subsection:training}
We train ENFORCE using standard unconstrained gradient descent methods (i.e., Adam~\cite{Kingma2014_AdamMethodStochastic}). We develop and use a constrained learning methodology using AdaNP to guide the neural network training to convergence, supported by the theoretical implications of orthogonal projections described in Section~\ref{subsec:adanp}.

\paragraph{Loss function}
The loss function used throughout this study takes the following general form:

\begin{equation}
\label{eq:loss}
\ell = \ell_T + \ell_D + \ell_C= \ell_T + \frac{\lambda_D}{N} \sum_{i=1}^{N} ||\hat{y}_i - \tilde{y}_i||^2 + \frac{\lambda_C}{N} \sum_{i=1}^{N} ||c(x_i, \tilde{y}_i)||,
\end{equation}

where the first term, $\ell_T$, is a task-specific loss function selected based on the target model. The second and third terms are regularization penalties that respectively minimize the projection displacement, $||\Hat{y}_i - \tilde{y}_i||$, and the constraint residual $||c(x_i, \tilde{y}_i)||$. The relative contributions of these terms are controlled by the scalar weights $\lambda_D$ and $\lambda_C$.
Minimizing the projection displacement aims to (1) ensure minor linearization error ($\varepsilon_L \sim \Delta y$) and (2) prevent the neural network from learning alternative functions whose projections onto the constraints fall within the neighborhood of the desired functions. Also, this additional loss term is suggested to reduce reliance on AdaNP, thereby lowering the computational cost during inference (i.e., by decreasing the depth of AdaNP).\\


\paragraph{Adaptive training strategy}
We propose a strategy to facilitate constrained learning during the early stages of training, guided by the theoretical insights presented in Section~\ref{subsec:adanp}. In the initial training phases, the preliminary prediction $\hat{y}$ may be inaccurate and lie far from the constraint manifold. Under such conditions, projecting onto a locally linearized approximation of the constraints can introduce substantial errors in the prediction. To mitigate this issue in practice, inspired by trust-region methods~\cite{Nocedal2006_NumericalOptimization}, we activate AdaNP only when the projection operation leads to an improvement in the prediction accuracy (e.g., quantified by a decrease in some loss measure $m_\ell$). This often leads to an unconstrained pre-training phase, followed by the activation of the AdaNP module. In other words, this serves as a heuristic to ensure that the prediction $\hat{y}$ lies sufficiently close to the constraint manifold.\\
The algorithm that we propose (cf. Algorithm~\ref{algorithm:adanp_activation}) assesses the effectiveness of the projection operation by quantifying and comparing a task-specific loss measure ($m_\ell$) computed on both the preliminary and the projected predictions. This loss measure is different from the complete training loss function. In the presented experiments, denoting by $\bar{y}$ either $\hat{y}$ or $\tilde{y}$, depending on the context, we define:
\begin{itemize}
    \item \textbf{Regression:} For supervised regression tasks, $m_\ell = \frac{1}{N} \sum_{i=1}^{N} ||y_i - \bar{y}_i||^2$  is the standard mean squared error loss.
    \item \textbf{Constrained optimization problem}: In the context of unsupervised learning for parametric optimization problems, we define the loss measure as $m_\ell = \frac{1}{N} \sum_{i=1}^{N} f_i(x, \bar{y}) + \frac{\lambda_C}{N} \sum_{i=1}^{N} ||c(x_i, \bar{y})||$. It is essential to include a penalty term for constraint violations, as strictly enforcing the constraints may inherently lead to larger values of the objective function, which is, however, acceptable.
\end{itemize}
The algorithm is applied at every forward pass during training. Specifically, it is executed after the raw backbone output (preliminary prediction $\hat{y}$) and the adaptive projection step (projected prediction $\tilde{y}$).\\
To provide an intuitive analogy, this mechanism is reminiscent of trust-region methods in constrained optimization~\cite{Nocedal2006_NumericalOptimization}, where a candidate step is only \textit{accepted} if it leads to a sufficient improvement in the objective. In our case, a task-specific cost function is used to assess whether the projection improves the prediction.

\begin{algorithm}
\caption{AdaNP activation algorithm during training}
\label{algorithm:adanp_activation}
\begin{algorithmic}[1]
\State \textbf{Input:} neural network $f_\theta$, input $x$, loss measure $m_\ell$
\State \textbf{Predict:} $\hat{y} = f_\theta(x)$
\State \textbf{Project:} $\tilde{y} = \mathcal{P}(\hat{y})$
\If{$m_\ell(\tilde{y}) > m_\ell(\hat{y})$}
    \State $\tilde{y} = \hat{y}$ \Comment{Discard the projection and use original prediction}
\Else
    \State $\tilde{y}_n = \operatorname{AdaNP}(\tilde{y})$ \Comment{Activate AdaNP}
\EndIf
\State \Return $\tilde{y}_n$
\end{algorithmic}
\end{algorithm}

\section{Experiments and discussion}
We evaluate the proposed method on multiple tasks: (i) illustrative function fitting with nonlinear equality and inequality constraints, (ii) learning solutions to scalable nonlinear parametric optimization problems, and (iii) real-world engineering case studies.
In the first experimental part, we provide in-depth discussion and analysis of the ENFORCE performance and restrict the comparison against unconstrained and soft-constrained methods.
In the following sections, we benchmark ENFORCE against the most relevant hard-constrained neural network baselines for nonlinear constraints, in particular DC3 and KKT-HardNet. Methods tailored to purely linear/convex constraint sets are not included, as they fall outside the scope of this work.\\
All experiments were conducted using an NVIDIA A100 Tensor Core GPU 80 GB, while the nonlinear programming solver runs on a CPU (11th Gen Intel(R) Core(TM) i7, 4 Core(s), 8 Logical Processor(s)).

\subsection{Function fitting with equality constraint}
\label{subsec:func_fitting}
We aim to fit the illustrative oscillating function $y:\mathbb{R}\rightarrow \mathbb{R}^2$ defined by
\begin{equation}
\label{eq:toy-problem}
\begin{aligned}
    y_1(x) &= 2\sin(fx),\\
    y_2(x) &= -\sin^2(fx) - x^2,
\end{aligned}
\end{equation}
where $x\in\mathbb{R}$ is the (scalar) input and $f\in\mathbb{R}$ is a fixed frequency parameter.
Notably, the system is implicitly linked by a nonlinear constraint $c(x,y_1, y_2) = (0.5y_1)^2+x^2+y_2$, involving both input and output variables. We train an ENFORCE model consisting of a 64-neuron 1-hidden-layer fully connected ReLU neural network as a backbone and an AdaNP module to force the predictions to satisfy the constraint. The supervised task loss is the mean squared error ($\ell_T = \frac{1}{N} \sum_{i=1}^{N} ||y_i - \tilde{y}_i||^2$), while $\lambda_C$ is set to zero (i.e., the constraint is addressed exclusively by AdaNP and no soft constraint term is used). To verify the regression capabilities, we sample 100 training data points from a uniform distribution in $x=[-2,2]$ and 100,000 test points in the same domain. Every run is repeated 5 times using different initialization seeds. We compare the method with a traditional (unconstrained) multilayer perceptron (MLP) and a soft-constrained neural network sharing the same architecture. We train all the NNs for 50,000 epochs, using Adam optimizer~\cite{Kingma2014_AdamMethodStochastic} and a learning rate of $10^{-3}$.

\subsubsection{Accuracy and constraint feasibility}
\begin{figure}[t]
    \centering
    \begin{subfigure}[t]{0.24\textwidth}
        \centering
        \includegraphics[width=\textwidth]{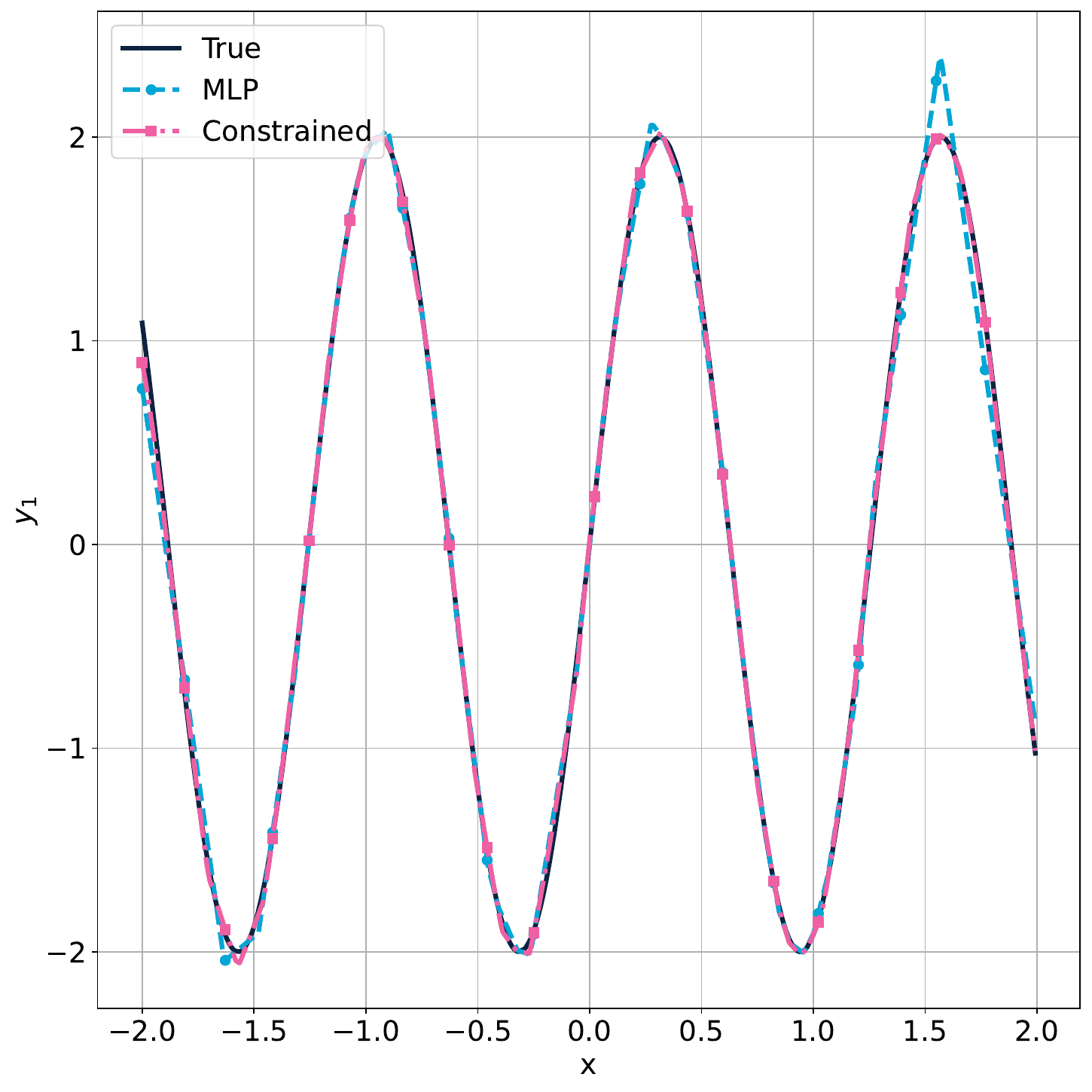}
        \caption{$y_1$}
        \label{fig:y1}
    \end{subfigure}
    \begin{subfigure}[t]{0.24\textwidth}
        \centering
        \includegraphics[width=\textwidth]{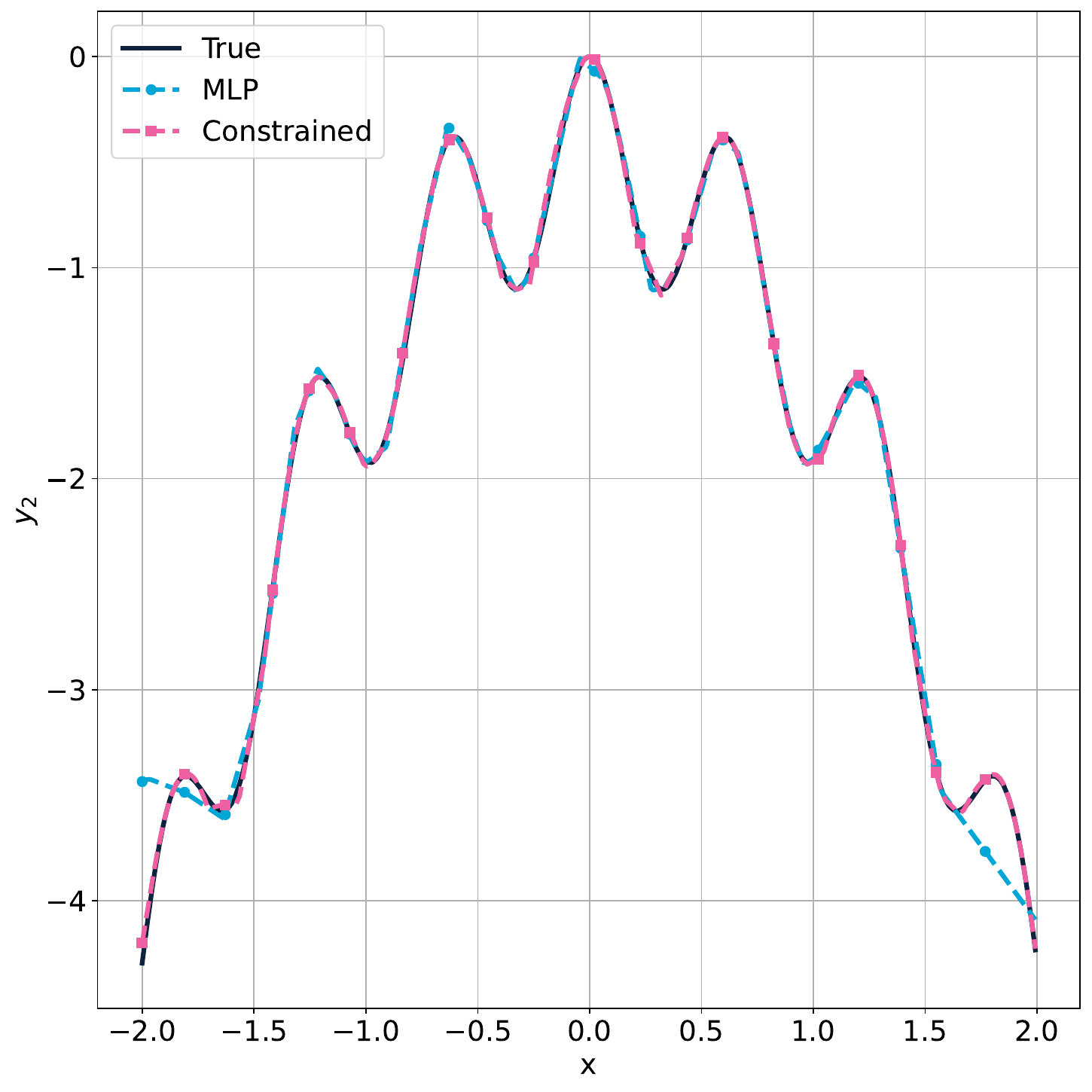}
        \caption{$y_2$}
        \label{fig:y2}
    \end{subfigure}
    \begin{subfigure}[t]{0.24\textwidth}
        \centering
        \includegraphics[width=\textwidth]{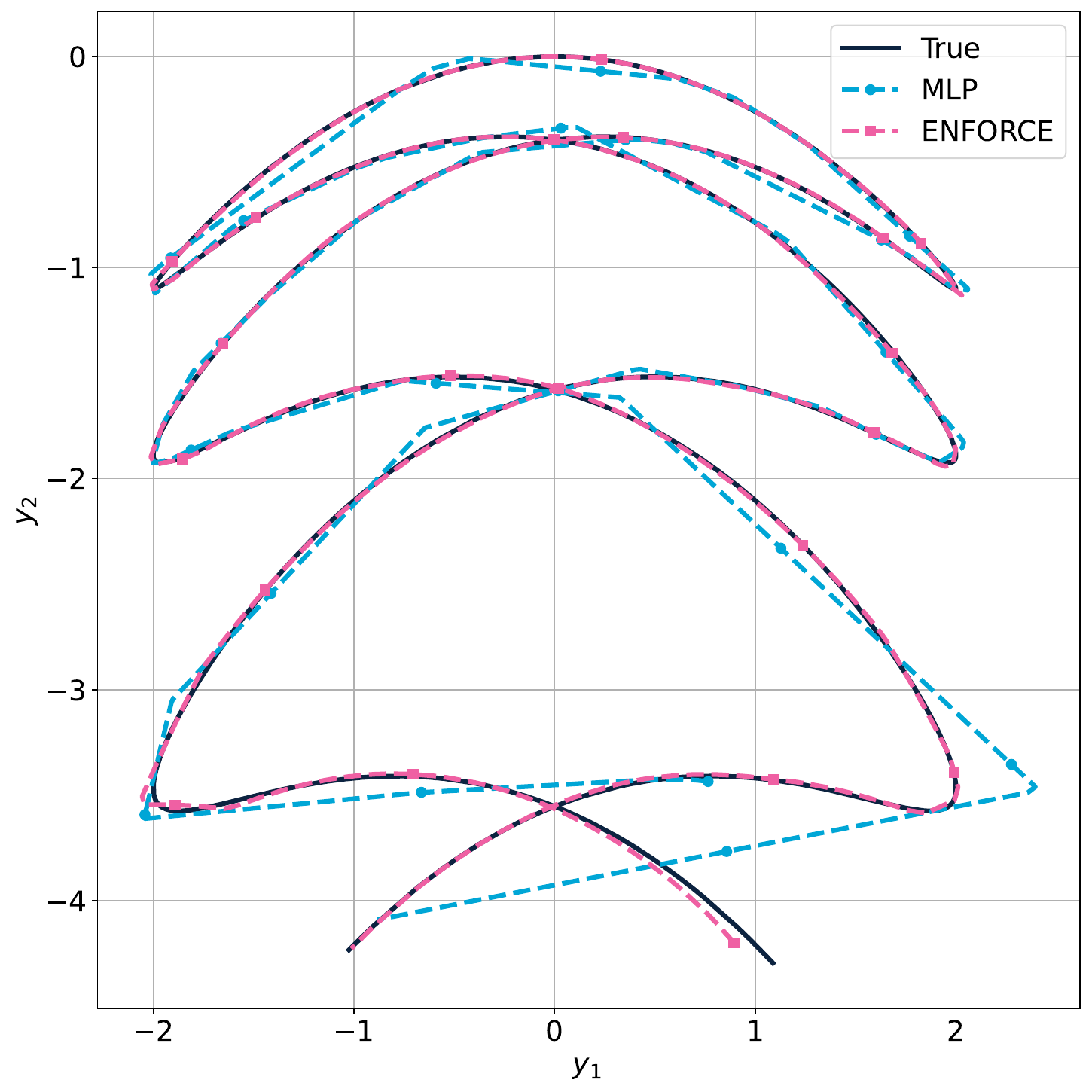}
        \caption{Constraint Satisfaction}
        \label{fig:constraint}
    \end{subfigure}
    \begin{subfigure}[t]{0.24\textwidth}
        \centering
        \includegraphics[width=\textwidth]{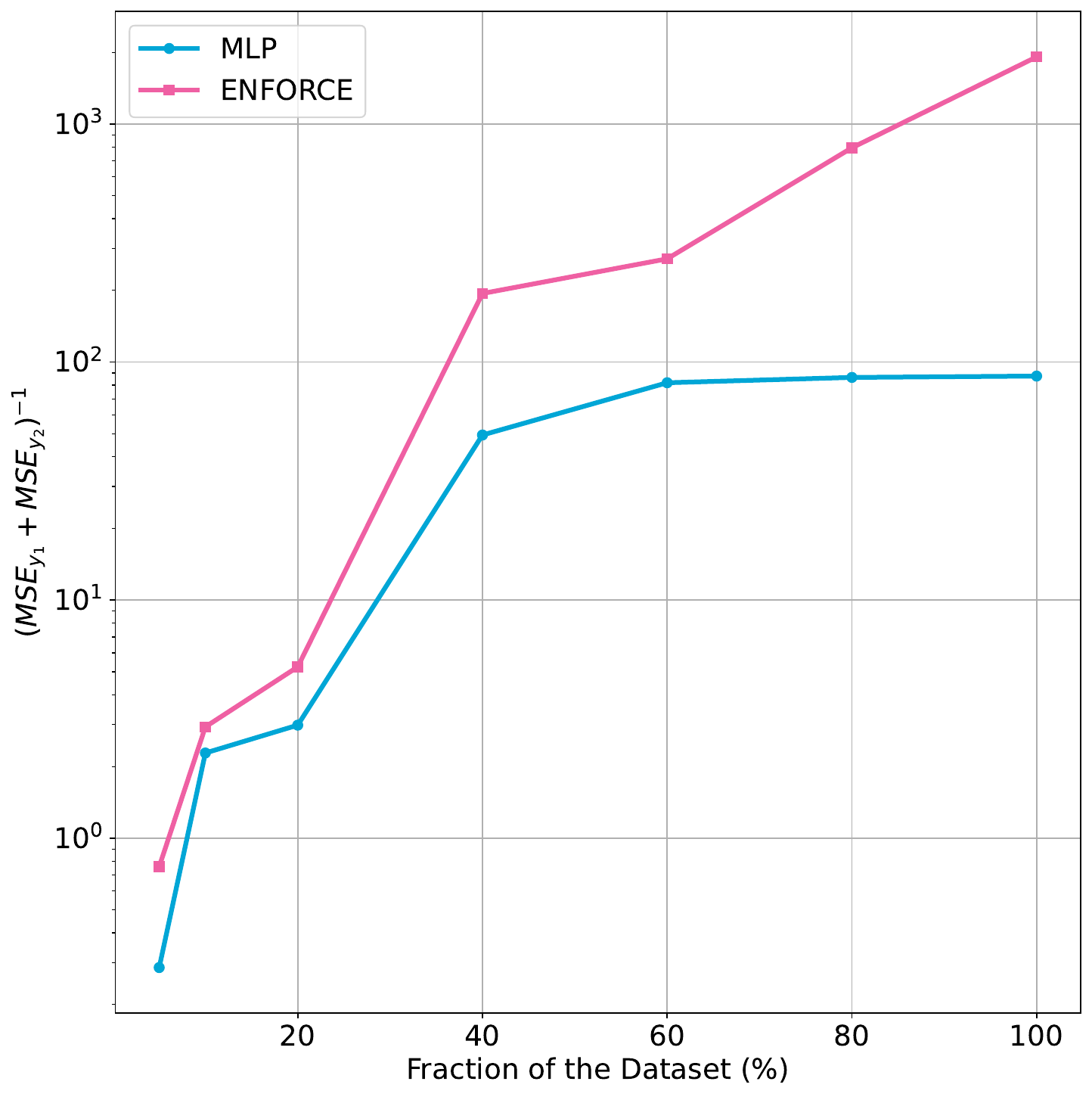}
        \caption{Data Scarcity}
        \label{fig:data-scarcity}
    \end{subfigure}
    \caption{Prediction comparison between ENFORCE ($\lambda_D = 0.5$, $\varepsilon_T = 10^{-4}$, $\varepsilon_I = 10^{-6}$) and a multilayer perceptron (MLP). ENFORCE enhances the overall accuracy and recover $\varepsilon$-feasibility with respect to nonlinear constraints. ENFORCE also performs better under data-scarcity conditions when the models are trained on uniformly sampled fractions of the dataset (Fig.~\ref{fig:data-scarcity}). This observation suggests that constrained learning may enhance data efficiency.}
    \label{fig:prediction}
\end{figure}

\begin{table}[h!]
    \centering
    \caption{Regression accuracy and constraint satisfaction of ENFORCE on 100,000 test samples when compared with a multilayer perceptron (MLP) and a soft-constrained neural network (Soft). Results for $\lambda_D=0.5$, $\varepsilon_T=10^{-4}$, and $\varepsilon_I = 10^{-6}$ are reported. We report the inference time for a batch of 1,000 samples, with $f=5$. (Note that $\text{MAPE} = \frac{100\%}{N} \sum_{i=1}^{N} \left| \frac{y^*_i - \tilde{y}_i}{y^*_i} \right|$).}
    \begin{tabular}{lccccc} 
        \hline
        Method & MAPE [\%] & $R^2$ & Mean eq. [\%] & Max eq. [\%] & Inf. time [s] \\ \hline
        MLP & 0.339 $\pm$ 0.083 & 0.994 $\pm$ 0.003 & 1.47 $\pm$ 0.33 & 17.13 $\pm$ 3.94 & 0.002 $\pm$ 0.000 \\
        Soft ($\lambda_C=1$) & 0.944 $\pm$ 0.143 & 0.972 $\pm$ 0.002 & 1.55 $\pm$ 0.16 & 7.77 $\pm$ 0.40 & 0.002 $\pm$ 0.000 \\
        ENFORCE & 0.060 $\pm$ 0.028 & 0.999 $\pm$ 0.000 & 0.00 $\pm$ 0.00 & 0.00 $\pm$ 0.00 & 0.008 $\pm$ 0.003 \\ \hline
    \end{tabular}
    \label{tab:Accuracy-Constraints}
\end{table}
One of the major concerns in constrained learning is related to keeping the expressivity and approximation capabilities of NNs~\cite{Cybenko1989_Approximationsuperpositionssigmoidal, Hornik1989_Multilayerfeedforwardnetworks} while enforcing constraints to be satisfied. Here, we evaluate the regression capabilities and robustness to spectral biases~\cite{Rahaman2018_SpectralBiasNeural} of ENFORCE on oscillating functions ($f=5$) while assessing the average and maximum constraint deviation (i.e., residual).
The main results are summarized in Table~\ref{tab:Accuracy-Constraints}. ENFORCE outperforms the soft-constrained neural network and the MLP achieving constraint satisfaction up to the prescribed inference tolerance $\varepsilon_I$ with minor computational costs (Fig.~\ref{fig:constraint}). The inference time for a batch of 1,000 samples is 6 ms longer when using ENFORCE compared to an MLP. This amount should be regarded as additive (+6 ms), not multiplicative (e.g., 4x relative to the MLP). Indeed, the computational complexity is entirely attributed to the AdaNP module, meaning that the backbone architecture has no impact. Therefore, if applied to larger backbones (e.g., transformers), the relative computational impact may become negligible.

\subsubsection{Empirical analysis of constrained learning and hyperparameters}
\label{sect:empirical_hyperp}
We report additional empirical observations on the effect of our constrained learning routine in the function fitting case study. We find that (i) ENFORCE outperforms the MLP even without projection (i.e., using only the backbone at inference), (ii) the method positively influences training dynamics and loss convergence, and (iii) the displacement weighting factor $\lambda_D$ is the dominant constrained-learning hyperparameter, while the impact of $\varepsilon_T$ is comparatively limited.

\begin{figure}[t]
    \centering
    \begin{subfigure}[t]{0.4\textwidth}
        \centering
        \includegraphics[width=\textwidth]{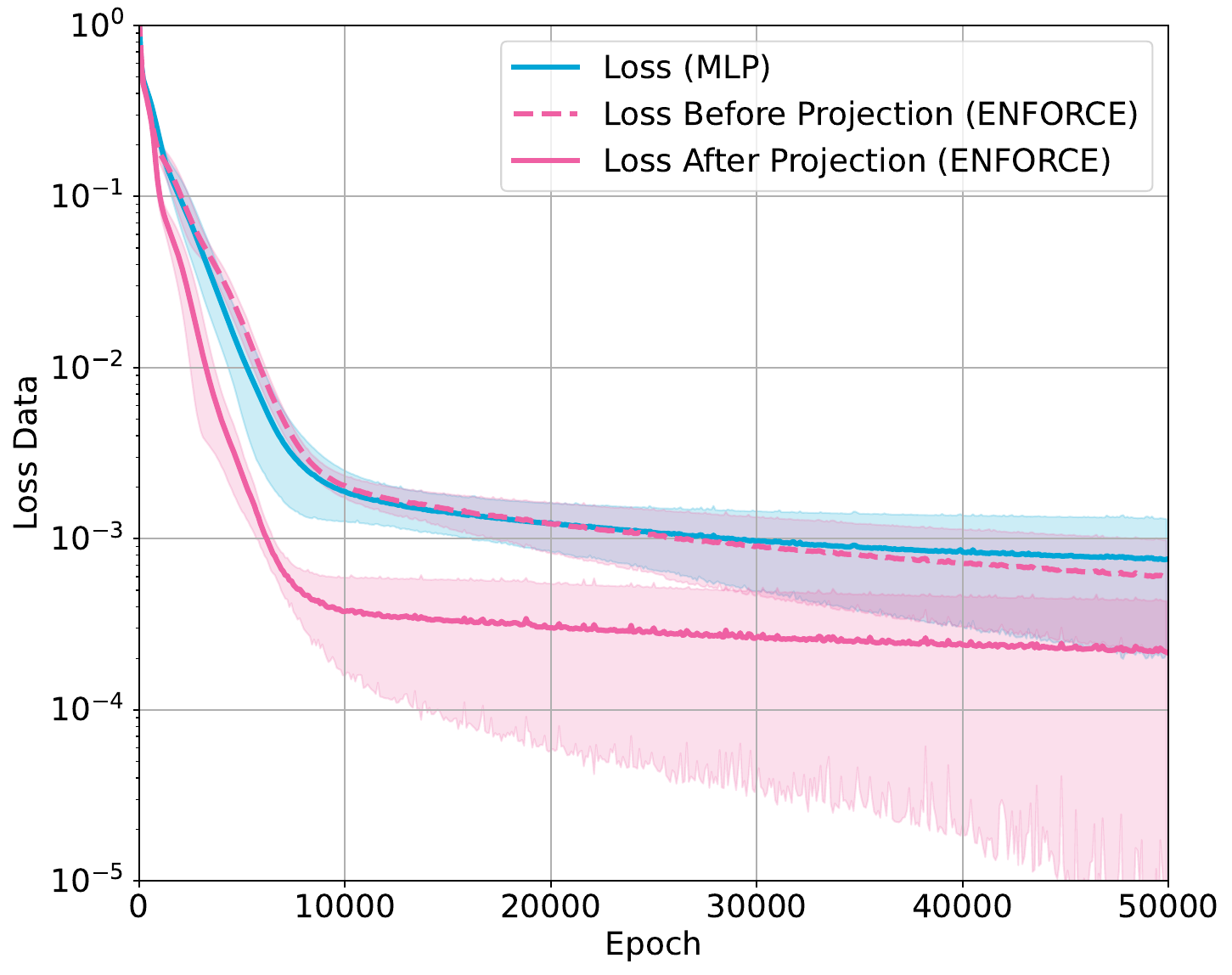}
        \caption{Loss neural network vs. ENFORCE}
        \label{fig:loss-comparison}
    \end{subfigure}
    \begin{subfigure}[t]{0.4\textwidth}
        \centering
        \includegraphics[width=\textwidth]{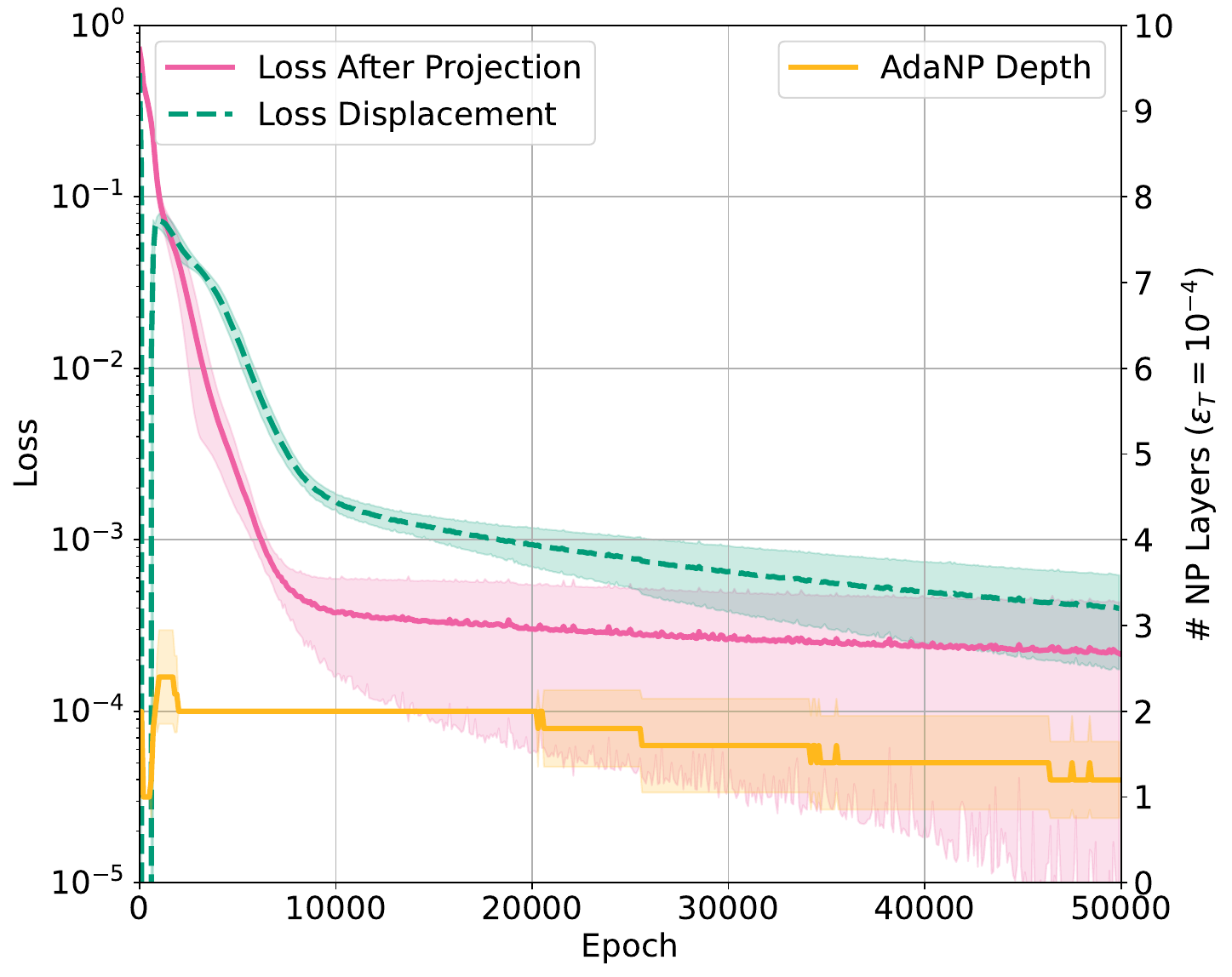}
        \caption{ENFORCE Losses and AdaNP}
        \label{fig:losses-ENFORCE}
    \end{subfigure}
    \caption{ENFORCE demonstrates improved convergence, achieving lower loss values compared to an unconstrained MLP. Enhanced training performances are reported for the backbone network of ENFORCE even before the action of AdaNP. This effect is enabled by the simultaneous minimization of the projection displacement (in green) and the action of the AdaNP module (in yellow). Note that we report average values across multiple runs, which explains why the depth of AdaNP appears as a step function with non-integer values.}
    \label{fig:loss}
\end{figure}

\paragraph{Effects of constrained learning}
Notably, a trained ENFORCE outperforms the MLP even before the projection steps, demonstrating superior performance using only the neural network backbone (Fig.~\ref{fig:loss-comparison}, dashed-pink line). This can be attributed to the structure of the hard-constrained learning process, where the predictions are adjusted via projection to satisfy underlying constraints. Unlike soft-constrained methods, which only penalize constraint violations in the loss function, the projection-based adjustments transform predictions to adhere strictly to the constraints. Consequently, after a few training steps, the model benefits from constrained learning, aligning its predictions more closely with valid regions of the solution space, resulting in improved predictions even before projection. Similar insights are also provided by Chen et al. (2021)~\cite{Chen2021_Theoryguidedhard}.
Therefore, the constrained learning approach is likely to yield improved results even when AdaNP is omitted during inference to enhance computational efficiency. However, it should be noted that in this scenario, constraint satisfaction cannot be guaranteed.

\paragraph{Training dynamics}

\begin{figure}[t]
    \centering
    \begin{subfigure}[t]{0.45\textwidth} 
        \centering
        \includegraphics[height=6cm]{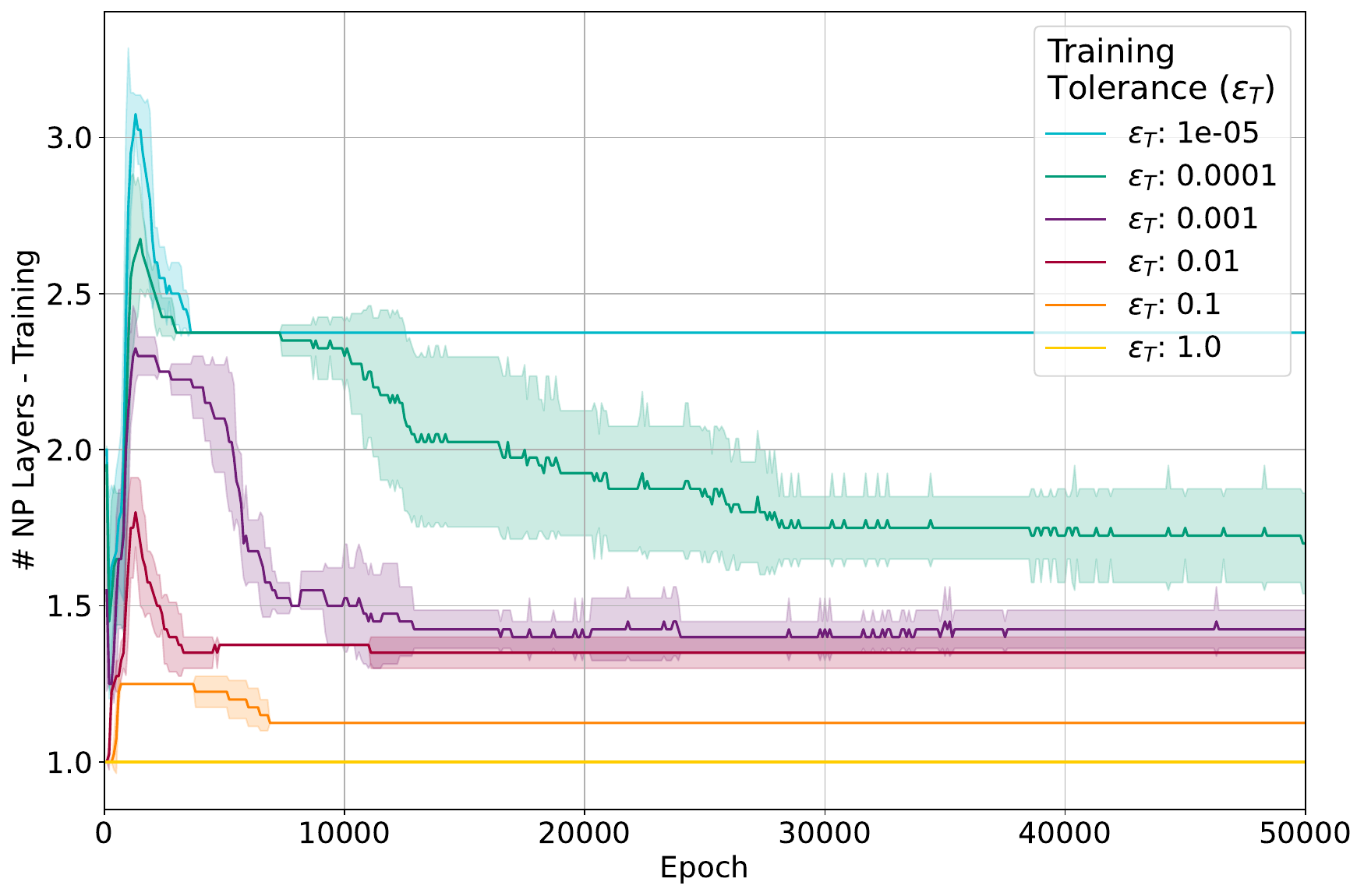}
        \caption{Depth of AdaNP (number of neural projection layers) during training.}
        \label{fig:NPs-training}
    \end{subfigure}%
    \hspace{0.08\textwidth} 
    \begin{subfigure}[t]{0.45\textwidth} 
        \centering
        \includegraphics[height=6cm]{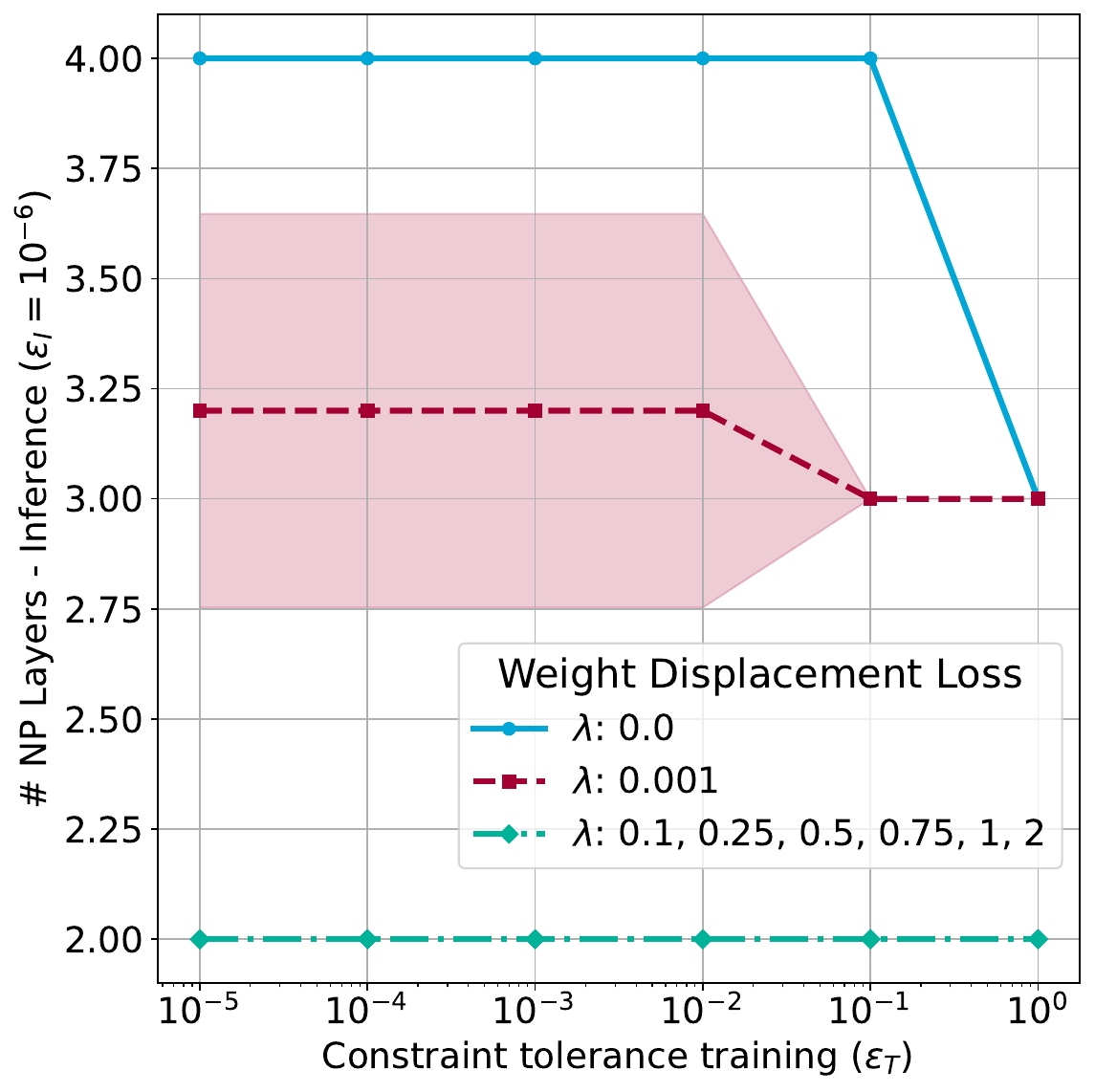}
        \caption{Depth of AdaNP (number of neural projection layers) during inference.}
        \label{fig:NPs-inference}
    \end{subfigure}
    \caption{Dynamic evolution of AdaNP during training and inference when different training hyperparameters are chosen. At training time, AdaNP is deeper as a smaller constraint tolerance $\varepsilon_T$ is chosen.  }
    \label{fig:NPs}
\end{figure}

To understand the training dynamics of ENFORCE, we analyze the loss curves shown in Fig.~\ref{fig:loss-comparison}, where the training data loss of ENFORCE is compared to the MLP. Being interested in the effect of hard-constrained learning and to ease the visualization, we do not report here the loss curve of the soft-constrained neural network. In this case study, AdaNP contributes to the learning process from the very early iterations (Fig.~\ref{fig:losses-ENFORCE}, orange line), suggesting that the projection operations positively guide the optimization process. The combination of approximated feasible predictions and minimization of projection displacement drives the learning process toward the constraints manifold. \\
The modified loss function effectively guides the training process toward smaller projection displacements (Fig.~\ref{fig:losses-ENFORCE}, green dashed line). The displacement loss decreases consistently during training due to the influence of the penalty term in the loss function. Moreover, the depth of AdaNP progressively diminishes over training iterations down to $\sim$1 layer (Fig.\ref{fig:losses-ENFORCE}, orange line), due to (1) improved overall regression accuracy and (2) smaller projection displacement (i.e., a better linear approximation of the constraints). This adaptive behavior optimizes computational resources by adjusting to the required tolerance at each iteration. Furthermore, this decay in AdaNP depth is consistently observed across different training tolerance values, as illustrated in Fig.\ref{fig:NPs-training}.

\begin{figure}[t]
    \centering
    \includegraphics[height=6cm]{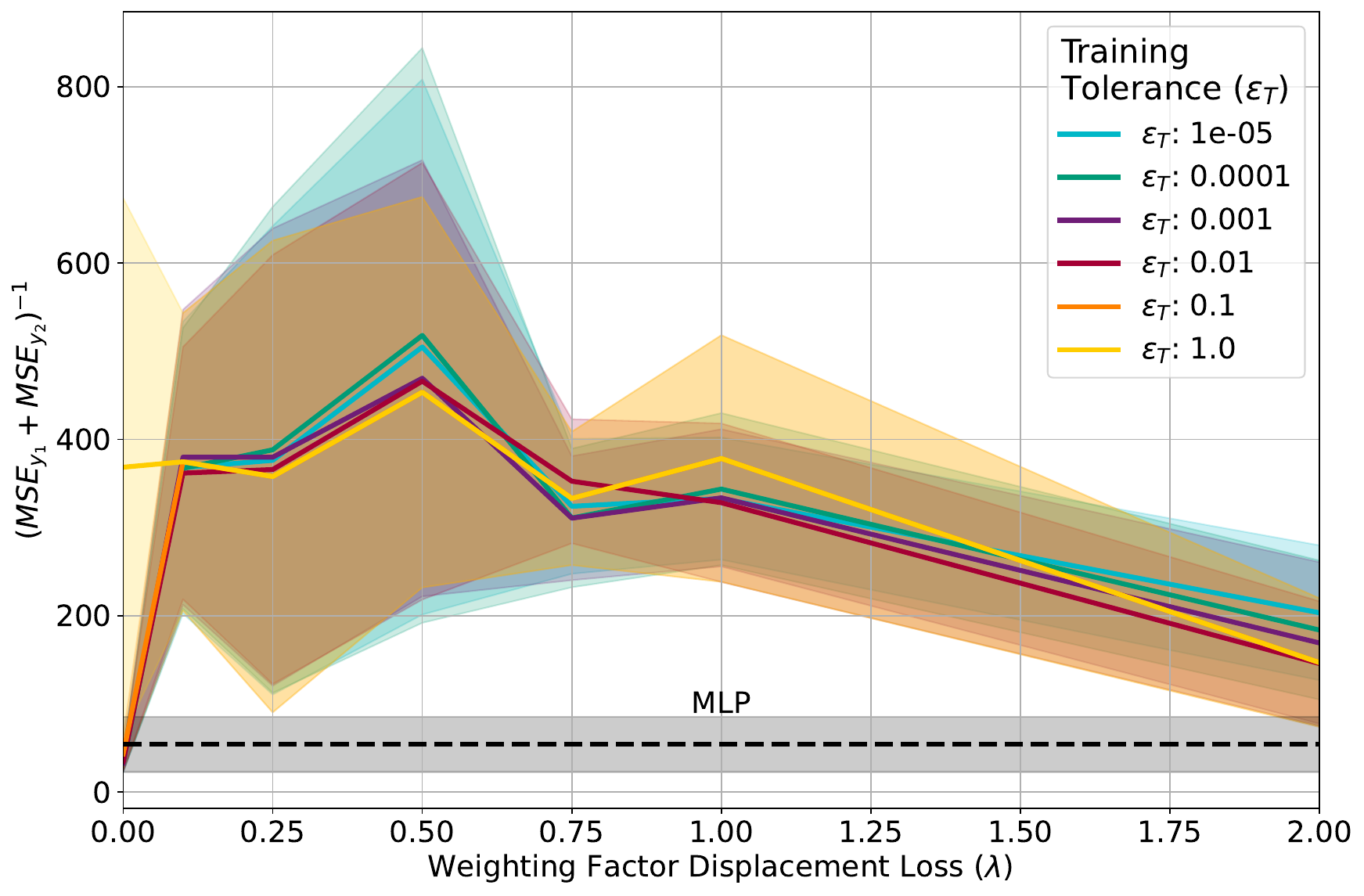}
    \caption{Influence of constrained learning hyperparameters on the accuracy of ENFORCE on the test set (note that here we plot the inverse of the mean squared error (MSE)). The weighting factor $\lambda_D$ favors the learning process if appropriately tuned. Conversely, the training tolerance $\varepsilon_T$ exhibits a small impact on performance, suggesting it can be set based on available resources.}
    \label{fig:lambda-epsilon}
\end{figure}

\paragraph{Constrained learning hyperparameters (training)}
We systematically analyze the influence of hyperparameters, such as the displacement weighting factor $\lambda_D$ and the tolerance $\varepsilon_T$, on the constrained learning process. 
Fig.~\ref{fig:lambda-epsilon} shows the influence of the hyperparameters on the accuracy of trained ENFORCE models evaluated on the test set. \\
The impact of the training tolerance $\varepsilon_T$ on the model accuracy does not exhibit a clear trend, as its effect varies unpredictably with the weighting factors. Moreover, its influence is generally small compared to the variance of different training runs (Fig.~\ref{fig:lambda-epsilon}). Intuitively, a smaller tolerance $\varepsilon_T$ necessitates deeper AdaNP modules, resulting in higher computational costs due to the increased number of neural projection layers. This effect is visible in Fig.~\ref{fig:NPs-training}, where the depth of AdaNP during training is reported (i.e., number of projection layers). The average depth of AdaNP increases to accommodate stricter tolerances. For example, it expands from one to three layers as the tolerance $\varepsilon_T$ is tightened from 1 to $10^{-5}$. Notably, in this case study, AdaNP operates with a minimum of one projection layer (i.e., when the tolerance is set to 1) and a maximum of 100. More importantly, the required depth tends to have a slower decay during training, if compared to using less strict tolerances (as visible in Fig.~\ref{fig:NPs-training}). Larger tolerances result, on average, in shallow AdaNP layers (approximately one layer). This significantly reduces the training time associated with the projection operations. Along with the minor impact on overall accuracy, this observation suggests setting the training tolerance $\varepsilon_T$ to less stringent requirements.\\
The regression accuracy is evidently affected by the choice of the displacement loss weighting factor $\lambda_D$ (Fig.~\ref{fig:lambda-epsilon}). Remarkably, unlike the challenging task of tuning weighting factors in soft-constrained methods~\cite{Wang2020_Understandingmitigatinggradient}, the constrained learning approach proposed here positively impacts accuracy regardless of the specific weighting factor chosen (as shown in Fig.~\ref{fig:lambda-epsilon}, the accuracy of ENFORCE is consistently greater than that of a standard MLP). However, an inappropriate choice of this parameter can result in suboptimal outcomes (e.g., when $\lambda_D=2$ in Fig.~\ref{fig:lambda-epsilon}). Therefore, careful tuning of this hyperparameter is warranted.\\

\paragraph{Constrained learning hyperparameters (inference)}
During inference, ENFORCE dynamically adapts the depth of AdaNP to ensure an average tolerance below $\varepsilon_I=10^{-6}$ in this case study. The required depth, however, also depends on the training parameters. Fig.~\ref{fig:NPs-inference} illustrates the number of NP layers needed to satisfy the constraint under varying $\lambda_D$ and $\varepsilon_T$. The weighting factor is shown to reduce the required number of NP layers by half, with no additional cost during training. This phenomenon can be attributed to the fact that, in the absence of a penalty for projection displacement, the neural network is free to learn a function that, although potentially far from the actual one, results in projections that fall within the vicinity of the ground truth. This approach, however, necessitates multiple projections. In contrast, the penalty term drives the model to learn a function that is sufficiently close to the ground truth, thereby reducing the number of neural projections required. Increasing the value of $\varepsilon_T$ impacts (positively) the depth of AdaNP at inference time when the displacement penalty factor is set to be small during training. This finding further supports the recommendation of employing shallow AdaNP modules during training, by relaxing the value of $\varepsilon_T$.\\
We conclude that the displacement loss weighting factor $\lambda_D$ plays an important role by balancing the contribution of the projection displacement error. On the other hand, enforcing strict satisfaction during training with an arbitrary small tolerance $\varepsilon_T$ does not necessarily improve the overall outcome.

\subsection{Function fitting with inequality constraints}
\label{subsec:sin_ineq}
We next illustrate ENFORCE on a simple regression problem with \emph{inequality}
constraints. The goal is to learn a scalar function $f:\mathbb{R}\to\mathbb{R}$
whose outputs must remain within a known envelope.
Let $x \in [0,3\pi]$ and define the ground-truth function
\begin{equation}
\label{eq:sin_ineq_true}
f(x) = h(x)\sin(x),
\qquad
h(x) = 1 + \frac{x^2}{3\pi^2}.
\end{equation}
The admissible set is the pointwise envelope $|y|\le h(x)$, equivalently described
by two inequality constraints
\begin{equation}
\label{eq:sin_ineq_constraints}
g_1(x,y) := y - h(x) \le 0,
\qquad
g_2(x,y) := -h(x) - y \le 0.
\end{equation}

We generate a training set of $N_{\mathrm{tr}}=1200$ inputs sampled uniformly
$x \sim \mathcal{U}(0,3\pi)$, with noisy labels
\begin{equation}
\label{eq:sin_ineq_noise}
\tilde{y} = f(x) + \varepsilon(x),
\qquad
\varepsilon(x)=\beta\,\mathrm{sign}(f(x)) + \eta,
\qquad
\eta \sim \mathcal{N}(0,\sigma^2),
\end{equation}
where $\sigma=0.3$ and $\beta=0.1$. The bias term $\beta\,\mathrm{sign}(f(x))$
shifts labels \emph{in the direction of the signal}, mimicking systematic measurement bias. In such conditions, an unconstrained regressor
that fits the training distribution may overshoot the envelope at peaks and troughs.
We evaluate regression accuracy using a noiseless test set of 300 samples drawn uniformly from $\mathcal{U}(0,3\pi)$ with labels $y=f(x)$.\\
We use a one-hidden-layer fully connected network with $64$ hidden neurons to predict the scalar output $y$. Specifically, we compare ENFORCE to an unconstrained MLP and a soft-constrained MLP with $\lambda_C=1$.
During projection in ENFORCE, we operate in an \emph{extended} output space that additionally
includes the multipliers $(\lambda_1,\lambda_2)$ required by the FB reformulation.
All methods are trained with Adam, a learning rate of $10^{-3}$, batch size $128$, and $500$ epochs. We use a training tolerance of $10^{-4}$ and inference tolerance
of $10^{-6}$, with a maximum of $100$ projection iterations, and $\lambda_D=0.5$.

\subsubsection{Results and comparison}
\begin{figure}
    \centering
    \includegraphics[width=0.8\linewidth]{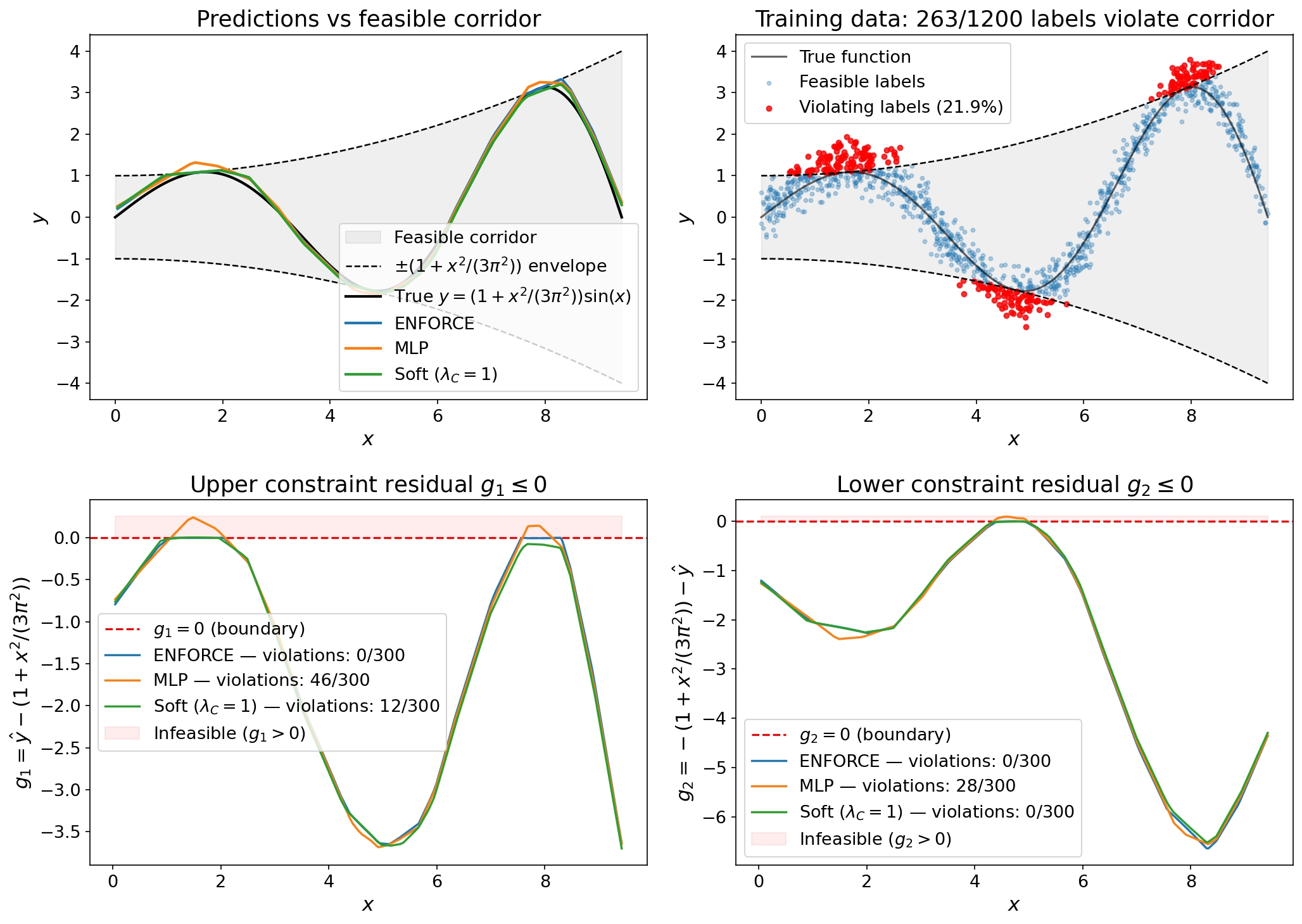}
    \caption{Function fitting with nonlinear inequality envelopes.}
    \label{fig:illustrative_ineq}
\end{figure}

Table~\ref{tab:illustrative_ineq} and Figure~\ref{fig:illustrative_ineq} report the
regression accuracy and constraint satisfaction of the three methods on 300 noiseless
test samples.
The unconstrained MLP learns a predictor biased in the direction of the training noise,
systematically overshooting the quadratic amplitude envelope at every peak and trough of
the target function: 15.33\,\% of test predictions violate the upper bound $g_1$, and 9.33\,\% violate the lower bound $g_2$.
Adding a quadratic penalty on constraint violations with weight $\lambda_C = 1$ (Soft)
reduces the upper-bound infeasibility to 4\,\% and eliminates lower-bound violations
entirely, while also improving the NRMSE from 7.84\,\% to 6.86\,\% owing to the
regularising effect of the penalty.
ENFORCE enforces both constraints to $\varepsilon\text{-feasibility}$ by construction: zero infeasible predictions
are recorded for both $g_1$ and $g_2$ across the entire test set.
The regression accuracy of ENFORCE is on par with
both baseline methods, confirming that the hard projection does not compromise
predictive quality.
The inference time of ENFORCE is approximately $100\times$ higher
than the unconstrained baselines, reflecting the cost of the AdaNP module
requiring on average 4 to 5 layers (projection iterations) to converge.

This illustrative case study highlights a practical limitation of penalty-based (soft) constraint enforcement in feasibility-critical settings: constraint satisfaction at inference remains \emph{approximate} and depends on the choice of the penalty weight $\lambda_C$. Although the penalty term encourages feasibility during training, selecting $\lambda_C$ entails an inherent trade-off: too small values may lead to systematic residual violations, while overly large values can bias training toward feasibility at the expense of data fit.
In contrast, ENFORCE leads to feasibility by construction. The neural network is trained end-to-end with a regression loss, and feasibility is subsequently recovered through a deterministic projection step that maps each raw prediction to an $\varepsilon$-feasible point with respect to the inequality set.

\begin{table}[h!]
    \centering
    \caption{%
        Regression accuracy and constraint satisfaction on 300 test samples
        for the function fitting with inequality constraints case study.
        Results for $\lambda_D = 0.5$, $\varepsilon_T = 10^{-4}$, and
        $\varepsilon_I = 10^{-6}$ are reported.
        $g_1$: upper bound violation; $g_2$: lower bound violation.
        NRMSE $= \mathrm{RMSE} / \sigma_{y}$, where $\sigma_{y}$ is the standard deviation of the target ground truth.%
    }
    \label{tab:illustrative_ineq}
    \setlength{\tabcolsep}{6pt}
    \begin{tabular}{lcccccccc}
        \hline
        Method
            & $R^2$
            & NRMSE [\%]
            & Infeas.\ $g_1$ [\%]
            & Infeas.\ $g_2$ [\%]
            & Inf.\ time [s] \\
        \hline
        MLP
            & 0.994
            & 7.84
            & 15.33
            & 9.33
            & 0.0002 \\
        Soft ($\lambda_C = 1$)
            & 0.995
            & 6.86
            & 4.00
            & 0.00
            & 0.0003 \\
        ENFORCE
            & 0.995
            & 7.16
            & 0.00
            & 0.00
            & 0.020 \\
        \hline
    \end{tabular}
\end{table}

\subsection{Constrained optimization problem}
\label{subsec:constr_opt_prob}
A relevant field in NNs research involves learning approximate solutions to constrained optimization problems as an inexpensive alternative to traditional solvers. Existing benchmarks for such a task lack scalable problems involving nonlinear equality constraints, which limits the evaluation of methods beyond the linear setting. We address this gap by introducing a new benchmark that incorporates nonlinear equality constraints while retaining scalability in problem complexity, following a state-of-the-art protocol~\cite{Donti2021_DC3learningmethod}.\\
We compare our method with alternative methods for learning (or solving) constrained optimization problems, such as MLP, soft-constrained MLP, the state-of-the-art DC3~\cite{Donti2021_DC3learningmethod}, and the deterministic nonlinear programming solver IPOPT~\cite{Waechter2005_implementationinteriorpoint}. We do not compare with baselines specialized for affine or convex constraints, such as RAYEN~\cite{Tordesillas2023_RAYENImpositionHard} and HardNet~\cite{Min2024_HardConstrainedNeural}, as they are not designed to handle nonlinear equality constraints. For every model, we use an equivalent fully-connected neural network backbone consisting of 2 hidden layers with 200 neurons. Training is performed with a batch size of 200, a learning rate of $10^{-4}$, and until model convergence (3,500 epochs for DC3; 1,000 for ENFORCE and the other baselines). Every run is repeated 5 times. Like Donti et al., we run the test inference on a single batch of 833 samples, and for the inference time of the optimizer, we assume full parallelization on 833 CPUs (cf.~\cite{Donti2021_DC3learningmethod} for more details).

\subsubsection{Nonconvex problem with linear equality constraints}
We consider the same class of nonconvex optimization problems as in Donti et al.~\cite{Donti2021_DC3learningmethod}, with the focus on equality constraints:

\begin{equation}
    \label{eq:lin_opt_prob_learning}
    \min_{y \in \mathbb{R}^{N_O}} f_{\text{obj}}(y) = \frac{1}{2} y^TQy + p^T \sin{y}, \quad \text{s.t.} \quad c^Ty = x,
\end{equation}

where $Q\in\mathbb{R}^{N_O \times N_O}\succeq0$, $p \in \mathbb{R}^{N_O}$, and $c \in \mathbb{R}^{N_C \times N_O}$ are randomly sampled constant parameters, while $x\in\mathbb{R}^{N_C}$ (with $N_C$ = $N_I$) is the variable parameter across problem instances. $Q$ is a diagonal matrix chosen to be positive semi-definite and $x$ is uniformly sampled in the interval $[-5,5]$. 
We aim to learn the optimal $y$ given an instance of $x$ in an unsupervised fashion. Rather than using a dataset of solved optimization instances, we minimize the objective in the unsupervised task loss $\ell_T = f_{\text{obj}}(y) = \frac{1}{2} y^TQy + p^T \sin{y}$.

\begin{table}
\centering
\caption{Results on a batch of 833 instances of nonconvex optimization problems with linear equality constraints involving 200 variables and 150 equality constraints. ENFORCE achieves feasibility up to the specified tolerance, is $25\times$ faster than IPOPT, and learns a $40\%$ better optimum than the state-of-the-art DC3 method. Baseline deep learning and soft-constrained methods show significant constraint violations and suboptimal predictions.}
\begin{tabular}{llllll}
\toprule
Method & Obj. value & Max eq. & Mean eq. & Inf. time [s] & Train. time [s] \\
\midrule
IPOPT~\cite{Waechter2005_implementationinteriorpoint} & $-10.64 \pm 0.00$ & $0.00 \pm 0.00$ & $0.00 \pm 0.00$ & $0.379 \pm 0.060$ & -- \\
MLP & $-52.99 \pm 0.01$ & $45.38 \pm 0.56$ & $9.14 \pm 0.02$ & $0.001 \pm 0.001$ & $540.87 \pm 14.77$ \\
Soft ($\lambda_c=1$) & $-8.18 \pm 0.18$ & $1.47 \pm 0.41$ & $0.09 \pm 0.00$ & $0.001 \pm 0.001$ & $651.17 \pm 29.16$ \\
DC3~\cite{Donti2021_DC3learningmethod} & $-6.27 \pm 0.07$ & $0.00 \pm 0.00$ & $0.00 \pm 0.00$ & $0.004 \pm 0.000$ & $1511.03 \pm 517.57$ \\
ENFORCE & $-10.59 \pm 0.00$ & $0.00 \pm 0.00$ & $0.00 \pm 0.00$ & $0.016 \pm 0.002$ & $834.78 \pm 4.14$ \\
\bottomrule
\end{tabular}
\label{tab:lin_optimizer_comparison}
\end{table}
In Table~\ref{tab:lin_optimizer_comparison}, we report the results on the constrained nonconvex task for 200 variables and 150 linear equality constraints. In Appendix~\ref{app:additional_exp} (Table~\ref{tab:scaling_metrics_linear}), we show how the performance of the methods scales with varying numbers of variables and constraints. Given the linear nature of the constraints, ENFORCE consistently guarantees the feasibility for all the test samples. ENFORCE learns a solution that is only $0.47\%$ suboptimal relative to the objective value obtained by IPOPT, while obtaining a $25\times$ acceleration. ENFORCE shows faster training convergence when compared to DC3 and an optimal objective gain of $40\%$. This improvement can be attributed to the stability of the projection mapping, in contrast to the null-space completion method. As expected, unconstrained and soft-constrained methods do not guarantee feasibility and may yield predicted optima with lower objective values than those computed by constraint-respecting solvers. However, in constrained optimization, such infeasible solutions are inadmissible, regardless of their objective value.


\subsubsection{Nonconvex problem with nonlinear equality constraints}
Extending the linear benchmark setting above, we introduce scalable problems with nonlinear equality constraints in both inputs and outputs, enabling systematic high-dimensional analysis beyond the linear case:
\begin{equation}
    \label{eq:opt_prob_learning}
    \min_{y \in \mathbb{R}^{N_O}} f_{\text{obj}}(y) = \frac{1}{2} y^TQy + p^T \sin{y}, \quad \text{s.t.} \quad y^TAy + c^Ty+d=x^3.
\end{equation}
Here, $A \in \mathbb{R}^{N_C \times N_O \times N_O}$ denotes a tensor holding $N_C$ randomly generated symmetric matrices, while the remaining parameters follow the same sampling procedure as previously described. The varying parameter $x$ is uniformly drawn from the range $[-5,5]$, and the dataset dimensionality and split remain unchanged. We consider a problem with 200 variables and 150 nonlinear constraints.\\
The results are reported in Table~\ref{tab:nonconvex_nonlinear_comparison}. 
ENFORCE successfully predicts optimal solutions satisfying the set of nonlinear constraints across the whole test set. ENFORCE consistently achieves a 25× speedup in inference compared to the nonlinear programming solver, while maintaining the optimal objective within $6\%$. Traditional deep learning and soft constraint methods perform poorly when faced with nonlinear constraints, resulting in significant infeasibility or failure to approximate an optimal solution. We do not report DC3 results for this nonlinear benchmark because the publicly available implementation is specialized to a specific hard-coded problem class (including problem-specific derivatives), and extending it to our setting would require substantial modifications. Once again, a scalability analysis across multiple problem dimensions is reported in Appendix~\ref{app:additional_exp} (Table~\ref{tab:scaling_metrics_nonlinear}).

\begin{table}[H]
\centering
\caption{Performance comparison on nonconvex problems with nonlinear equality constraints involving 200 variables and 150 constraints. ENFORCE predicts feasible and near-optimal solutions across the entire test set, achieving a $25\times$ speedup over IPOPT. Other deep learning methods exhibit significant constraint violations and suboptimal performance.}
\begin{tabular}{llllll}
\toprule
Method & Obj. value & Max eq. & Mean eq. & Inf. time [s] & Train. time [min] \\
\midrule
IPOPT~\cite{Waechter2005_implementationinteriorpoint} & $-29.45 \pm 0.00$ & $0.00 \pm 0.00$ & $0.00 \pm 0.00$ & $3.40 \pm 1.40$ & -- \\
MLP & $-53.07 \pm 0.00$ & $497.38 \pm 4.64$ & $118.39 \pm 0.07$ & $0.002 \pm 0.001$ & $10.56 \pm 0.32$ \\
Soft ($\lambda_c=1$) & $(6.03 \pm 0.04) \times 10^{4}$ & $79.30 \pm 3.70$ & $16.68 \pm 0.06$ & $0.001 \pm 0.000$ & $15.04 \pm 0.62$ \\
ENFORCE & $-27.77 \pm 0.02$ & $0.00 \pm 0.00$ & $0.00 \pm 0.00$ & $0.14 \pm 0.08$ & $69.35 \pm 23.11$ \\
\bottomrule
\end{tabular}
\label{tab:nonconvex_nonlinear_comparison}
\end{table}

\subsection{Engineering case studies}
We evaluate ENFORCE on two engineering case studies introduced by Iftakher et al.~\cite{Iftakher2025_PhysicsInformedNeural}. The first involves enforcing nonlinear equality constraints arising from thermodynamic relationships in a chemical process simulation. The second considers a pooling problem with both nonlinear equality and inequality constraints derived from mass balances and product specifications. In this section, we compare ENFORCE to the state-of-the-art KKT-Hardnet model described in~\cite{Iftakher2025_PhysicsInformedNeural} and to the accelerated \texttt{jax} implementation (KKT-Hardnet-v2). 

\subsubsection{Chemical process simulation (nonlinear equality constraints)}
We first consider the extractive distillation case study introduced by Iftakher et al.~\cite{Iftakher2025_PhysicsInformedNeural}, which models the separation of the azeotropic refrigerant mixture R-410A using the ionic liquid [EMIM][SCN]. The task is to train neural-network surrogate models on steady-state simulation data while enforcing nonlinear equality constraints arising from the underlying thermodynamic relationships. We use the dataset provided by the authors, generated from Aspen Plus simulations after filtering unconverged runs (cf.~\cite{Iftakher2025_PhysicsInformedNeural} for details on the process model, constraints, and data generation), together with the same training setup.\\
As shown in Table~\ref{tab:case_study_column}, ENFORCE satisfies the nonlinear equality constraints up to the prescribed tolerance while achieving the lowest prediction error, with an MSE of $(1.2 \pm 0.1)\times10^{-7}$. Compared with the original KKT-Hardnet, ENFORCE is about $50\times$ faster at inference and about $100\times$ faster during training. The accelerated \texttt{jax}-based KKT-Hardnet-v2 substantially reduces the computational gap with respect to the original implementation, reaching inference and training times comparable to ENFORCE. However, ENFORCE retains higher predictive accuracy while providing the same level of constraint satisfaction.

\begin{table}[h!]
    \centering
    \caption{Engineering case study on a chemical process simulation (extractive distillation) with nonlinear thermodynamics.}
    \begin{tabular}{cccccc}
    \toprule
        Method & MSE & Max eq. & Mean eq. & Inf. time [s] & Train. time [s] \\
        \midrule
        MLP & $(7.6 \pm 2.1)\times10^{-8}$ & $0.003 \pm 0.001$ & $(2 \pm 0)\times10^{-4}$ & $(1.0\pm 0.0) \times10^{-4}$ & $105 \pm 1$\\
        KKT-Hardnet & $(1.1 \pm 0.0)\times10^{-4}$ & $<10^{-7}$ & $<10^{-7}$ & $0.29\pm 0.04$ & $38,631 \pm 440$\\
        KKT-Hardnet-v2 & $(2.0 \pm 0.0)\times10^{-6}$ & $<10^{-7}$ & $<10^{-7}$ & $0.005\pm 0.001$ & $415 \pm 4$\\
        ENFORCE & $(1.2 \pm 0.1)\times10^{-7}$ & $<10^{-7}$ & $<10^{-7}$ & $0.006 \pm 0.001$ & $391 \pm 2$ \\ 
    \bottomrule
    \end{tabular}
    \label{tab:case_study_column}
\end{table}

\subsubsection{Pooling problem (nonlinear equality and inequality constraints)}
\begin{figure}
    \centering
    \includegraphics[width=0.7\linewidth]{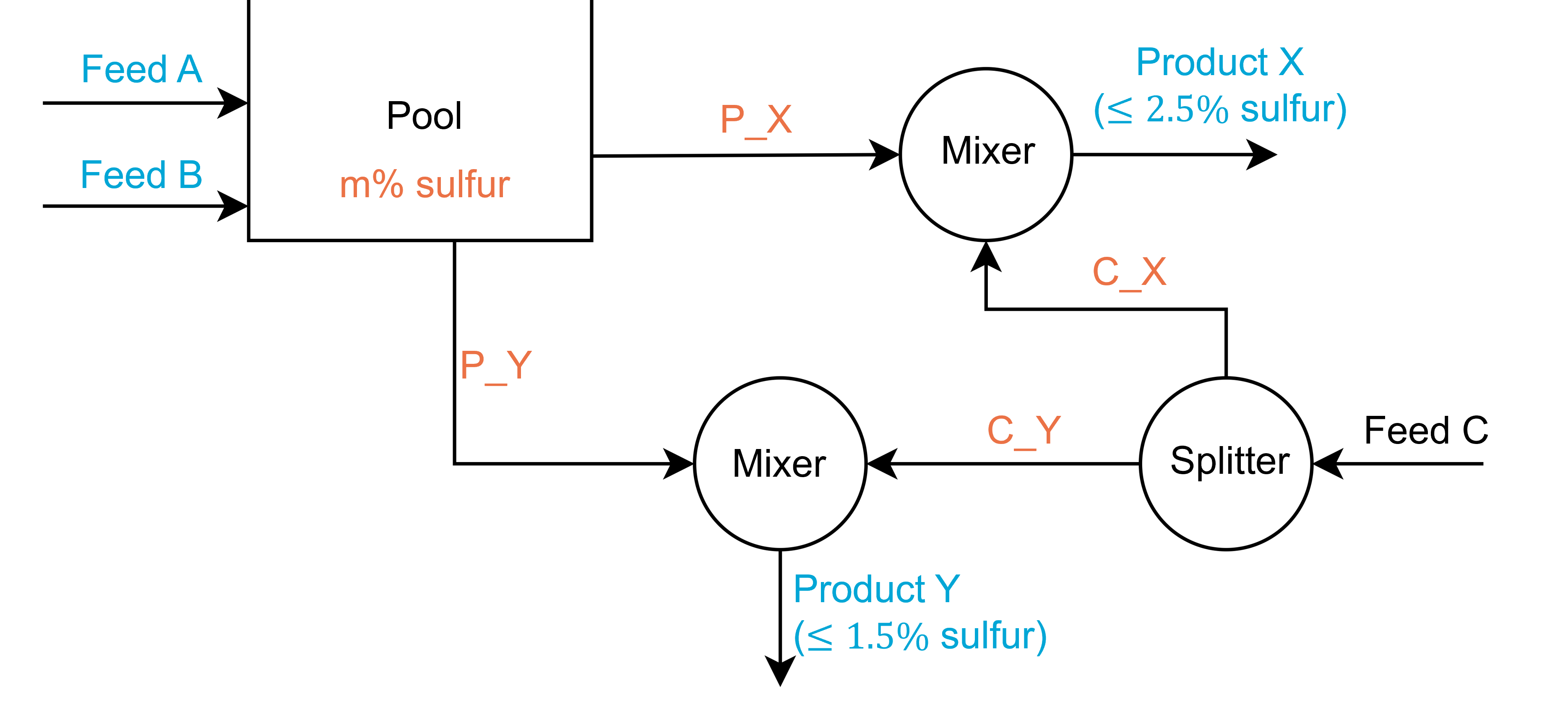}
    \caption{Pooling flowsheet representation adapted from~\cite{Iftakher2025_PhysicsInformedNeural}. In blue and orange are denoted the input and output variables, respectively. The system is subject to inequality constraints on product specifications and equality constraints enforcing material balances in the pool.}
    \label{fig:pool_flowsheet}
\end{figure}

As a second engineering case study, we consider the pooling problem originally described by Floudas et al.~\cite{Floudas1999_HandbookTestProblems} and used as a benchmark by Iftakher et al.~\cite{Iftakher2025_PhysicsInformedNeural}. The process (in Figure~\ref{fig:pool_flowsheet}) consists of a pooling unit, a splitter, and two mixers that combine three feed streams (A,B, and C) to produce two product streams (X and Y). This benchmark includes both nonlinear equality and inequality constraints: the equality constraints arise from material balances and sulfur conservation in the pool, while the inequality constraints enforce sulfur-content specifications on the final products. Following the benchmark introduced by Iftakher et al.~\cite{Iftakher2025_PhysicsInformedNeural}, the inputs are the feed and product flowrates, and the outputs are the pooled-stream sulfur fraction together with the internal flow variables. We use the dataset and problem setup provided by the authors; details on the constraints and data generation are reported in~\cite{Iftakher2025_PhysicsInformedNeural}.

As shown in Table~\ref{tab:case_study_pooling}, ENFORCE enforces both nonlinear equality and inequality constraints up to the prescribed tolerance, while improving the MSE with respect to the unconstrained MLP. Compared with the original KKT-Hardnet, ENFORCE is about $12\times$ faster at inference and about $15\times$ faster during training, while retaining the same level of constraint satisfaction.
KKT-Hardnet-v2 significantly improves over the original implementation and achieves the lowest MSE in this benchmark, with training time shorter than ENFORCE. However, ENFORCE remains faster at inference and recovers stricter feasibility for both equality and inequality constraints. Although outside the scope of this work, an accelerated \texttt{jax}-based implementation of ENFORCE could further improve its computational efficiency.

\begin{table}[h!]
    \centering
    \caption{Engineering case study on the pooling problem introduced by Floudas et al.~\cite{Floudas1999_HandbookTestProblems} and used as a benchmark by Iftakher et al.~\cite{Iftakher2025_PhysicsInformedNeural}. The problem involves both nonlinear equality and inequality constraints.}
    \begin{tabular}{cccccccc}
    \toprule
        Method & MSE & Max eq. & Max ineq. & Inf. time [s] & Train. time [s] \\
        \midrule
        MLP & $64\pm3$ & $7.4\pm2.2$ &  $3.4\pm1.5$  & $(2\pm 0) \times10^{-4}$ & $579 \pm 1$ \\
        KKT-Hardnet & $54\pm 1$ & $<10^{-7}$ & $<10^{-7}$ & $0.66\pm0.07$ & $61,768 \pm 1,155$ \\
        KKT-Hardnet-v2 & $48\pm 10$ & $(0.3\pm 2.7) \times10^{-2}$ & $(0.3\pm 2.8) \times10^{-2}$ & $0.18\pm0.42$ & $3,406 \pm 66$ \\
        ENFORCE & $61\pm3$ & $<10^{-7}$  &  $<10^{-7}$ & $0.057\pm0.007$ & $4,041 \pm 209$ \\
    \bottomrule
    \end{tabular}
    \label{tab:case_study_pooling}
\end{table}

\section{Conclusions}
We propose ENFORCE, a hard-constrained learning architecture that combines a neural network backbone with an adaptive-depth neural projection module (AdaNP) to enforce nonlinear equality and inequality constraints in the predictions. Equality nonlinear $\mathcal{C}^1$ constraints $c(x,y)=0$ are enforced up to a prescribed tolerance $\varepsilon$, and exactly in the affine-in-$y$ case. We generalize the approach to nonlinear inequality constraints $g(x,y)\le 0$ via a Fischer--Burmeister reformulation in an extended output space. For affine constraint sets, we prove that the resulting projection mapping is non-expansive (1-Lipschitz), ensuring stable gradient propagation. For nonlinear constraints, we provide a local convergence analysis under standard regularity assumptions.
The feasibility-recovery module AdaNP is not inherently tied to neural network backbones and could be coupled with other differentiable predictors, offering a route toward constraint-aware machine learning modeling.\\
The method is demonstrated on (i) function fitting with nonlinear equality and inequality constraints, (ii) scalable nonlinear parametric optimization problems with equality constraints, and (iii) two real-world engineering case studies with nonlinear equalities and inequalities. In the illustrative function-fitting benchmarks, ENFORCE achieves $\varepsilon$-feasibility by construction and eliminates systematic violations that persist under penalty-based training. In the optimization benchmarks, ENFORCE returns feasible solutions to numerical tolerance while providing substantial inference speedups over a nonlinear programming solver (up to $25\times$ in our tested instances) and improved objective values relative to the DC3 baseline. In the engineering case studies, ENFORCE recovers feasibility to numerical tolerance and offers favorable efficiency trade-offs, matching the feasibility guarantees of the state-of-the-art KKT-HardNet while reducing inference and training time (up to $50\times$ and $1000\times$, respectively) at a modest accuracy trade-off.\\

\paragraph{Limitations}
The guarantees of ENFORCE for nonlinear constraints are local: AdaNP relies on the accuracy of the linearization and on a well-conditioned constraint Jacobian in the neighborhood of a feasible point. If the backbone prediction is far from the constraint manifold, if LICQ fails, or if the Jacobian is severely ill-conditioned, AdaNP may require many iterations and may not reach the target tolerance within the imposed depth cap. In addition, the computational and memory cost of each projection step scales cubically with the number of enforced equality constraints ($\mathcal{O}(N_C^3)$) due to the Gram-matrix inversion; this can become the dominant cost when hundreds (or more) of equality constraints must be enforced per instance. For inequality constraints, the Fischer--Burmeister reformulation increases the dimension of the extended output space and may introduce additional numerical sensitivity; in practice we mitigate such issues with stabilized reformulations and eigenvalue-clamping safeguards, but extreme scale disparities can still increase projection cost.


%
\section*{Statements and Declarations}

\subsection*{Funding}
This research is supported by Shell Global Solutions International B.V., for which we express sincere gratitude.

\subsection*{Competing interests}
The authors declare that they have no known competing financial interests or personal relationships that could have appeared to in influence the work reported in this paper.

\subsection*{Author contributions} Giacomo Lastrucci: Writing -- original draft, Visualization, Validation, Software, Methodology, Investigation, Formal analysis, Data curation, Conceptualization. Artur M. Schweidtmann: Writing -- review \& editing, Supervision, Resources, Project administration, Methodology, Investigation, Funding acquisition, Conceptualization, Formal analysis. \subsection*{Data availability} The data used in this study are publicly available. All datasets are linked and can be downloaded from the code repository: \url{https://github.com/process-intelligence-research/ENFORCE}. \subsection*{Code availability} The source code of ENFORCE, together with the scripts required to reproduce the numerical experiments, is publicly available at: \url{https://github.com/process-intelligence-research/ENFORCE}. Furthermore the package is published in PyPI and can be installed with \texttt{pip install enforce-nn}.

\nolinenumbers
\bibliographystyle{unsrt}  
\bibliography{ENFORCE}  

@misc{tensorflow2015-whitepaper,
title={{TensorFlow}: Large-Scale Machine Learning on Heterogeneous Systems},
url={http://tensorflow.org/},
note={Software available from tensorflow.org},
author={
    Mart\'{\i}n~Abadi and
    Ashish~Agarwal and
    Paul~Barham and
    Eugene~Brevdo and
    Zhifeng~Chen and
    Craig~Citro and
    Greg~S.~Corrado and
    Andy~Davis and
    Jeffrey~Dean and
    Matthieu~Devin and
    Sanjay~Ghemawat and
    Ian~Goodfellow and
    Andrew~Harp and
    Geoffrey~Irving and
    Michael~Isard and
    Yangqing Jia and
    Rafal~Jozefowicz and
    Lukasz~Kaiser and
    Manjunath~Kudlur and
    Josh~Levenberg and
    Dan~Man\'{e} and
    Rajat~Monga and
    Sherry~Moore and
    Derek~Murray and
    Chris~Olah and
    Mike~Schuster and
    Jonathon~Shlens and
    Benoit~Steiner and
    Ilya~Sutskever and
    Kunal~Talwar and
    Paul~Tucker and
    Vincent~Vanhoucke and
    Vijay~Vasudevan and
    Fernanda~Vi\'{e}gas and
    Oriol~Vinyals and
    Pete~Warden and
    Martin~Wattenberg and
    Martin~Wicke and
    Yuan~Yu and
    Xiaoqiang~Zheng},
  year={2015},
}

@misc{paszke2019pytorchimperativestylehighperformance,
      title={PyTorch: An Imperative Style, High-Performance Deep Learning Library}, 
      author={Adam Paszke and Sam Gross and Francisco Massa and Adam Lerer and James Bradbury and Gregory Chanan and Trevor Killeen and Zeming Lin and Natalia Gimelshein and Luca Antiga and Alban Desmaison and Andreas Köpf and Edward Yang and Zach DeVito and Martin Raison and Alykhan Tejani and Sasank Chilamkurthy and Benoit Steiner and Lu Fang and Junjie Bai and Soumith Chintala},
      year={2019},
      eprint={1912.01703},
      archivePrefix={arXiv},
      primaryClass={cs.LG},
      url={https://arxiv.org/abs/1912.01703}, 
}

@misc{jax2018_github,
  author = {James Bradbury and Roy Frostig and Peter Hawkins and Matthew James Johnson and Chris Leary and Dougal Maclaurin and George Necula and Adam Paszke and Jake Vander{P}las and Skye Wanderman-{M}ilne and Qiao Zhang},
  title = {{JAX}: composable transformations of {P}ython+{N}um{P}y programs},
  url = {http://github.com/jax-ml/jax},
  version = {0.3.13},
  year = {2018},
}

@Article{Raissi2019_Physicsinformedneural,
  author     = {M. Raissi and P. Perdikaris and G.E. Karniadakis},
  journal    = {Journal of Computational Physics},
  title      = {Physics-informed neural networks: A deep learning framework for solving forward and inverse problems involving nonlinear partial differential equations},
  year       = {2019},
  month      = {feb},
  pages      = {686--707},
  volume     = {378},
  comment    = {Constrained included as penalty term in the loss function using automatic differentiation},
  doi        = {https://doi.org/10.1016/j.jcp.2018.10.045},
  publisher  = {Elsevier {BV}},
  readstatus = {read},
}

@Article{Karniadakis2021_Physicsinformedmachine,
  author     = {George Em Karniadakis and Ioannis G. Kevrekidis and Lu Lu and Paris Perdikaris and Sifan Wang and Liu Yang},
  journal    = {Nature Reviews Physics},
  title      = {Physics-informed machine learning},
  year       = {2021},
  month      = {may},
  number     = {6},
  pages      = {422--440},
  volume     = {3},
  doi        = {10.1038/s42254-021-00314-5},
  publisher  = {Springer Science and Business Media {LLC}},
  ranking    = {rank4},
  readstatus = {read},
}

@Article{Beucler2019_EnforcingAnalyticConstraints,
  author     = {Beucler, Tom and Pritchard, Michael and Rasp, Stephan and Ott, Jordan and Baldi, Pierre and Gentine, Pierre},
  journal    = {Physical Review Letters},
  title      = {Enforcing Analytic Constraints in Neural-Networks Emulating Physical Systems},
  year       = {2019},
  issn       = {1079-7114},
  month      = mar,
  number     = {9},
  pages      = {098302},
  volume     = {126},
  copyright  = {arXiv.org perpetual, non-exclusive license},
  doi        = {https://doi.org/10.1103/PhysRevLett.126.098302},
  publisher  = {arXiv},
  readstatus = {read},
}

@Article{Chen2024_PhysicsInformedNeural,
  author    = {Chen, Hao and Flores, Gonzalo E. Constante and Li, Can},
  journal   = {Computers \&  Chemical Engineering},
  title     = {Physics-Informed Neural Networks with Hard Linear Equality Constraints},
  year      = {2024},
  issn      = {0098-1354},
  month     = oct,
  pages     = {108764},
  volume    = {189},
  copyright = {arXiv.org perpetual, non-exclusive license},
  doi       = {https://doi.org/10.1016/j.compchemeng.2024.108764},
  publisher = {arXiv},
}

@Article{Amos2017_OptNetDifferentiableOptimization,
  author    = {Amos, Brandon and Kolter, J. Zico},
  journal   = {Proceedings of the 34th International Conference on Machine Learning, Sydney, Australia, PMLR 70, 2017},
  title     = {OptNet: Differentiable Optimization as a Layer in Neural Networks},
  year      = {2017},
  comment   = {They develop differentiable optimization layers for quadratic problems (quadratic objective and linear eq/ineq). The main contribution is a computationally cheap wayof obtaining the gradients in the backward pass. However, they have to solve a constrained convex optimization problem at every forward pass (they develop a batched interior point method).},
  copyright = {arXiv.org perpetual, non-exclusive license},
  doi       = {https://doi.org/10.48550/arXiv.1703.00443},
  publisher = {arXiv},
}

@Article{Donti2021_DC3learningmethod,
  author    = {Donti, Priya L. and Rolnick, David and Kolter, J. Zico},
  journal   = {International Conference on Learning Representations},
  title     = {DC3: A learning method for optimization with hard constraints},
  year      = {2021},
  copyright = {Creative Commons Attribution 4.0 International},
  doi       = {https://doi.org/10.48550/arXiv.2104.12225},
  publisher = {arXiv},
}

@Article{Mukherjee2024_developmentsteadystate,
  author    = {Mukherjee, Angan and Bhattacharyya, Debangsu},
  journal   = {Computers \&  Chemical Engineering},
  title     = {On the development of steady-state and dynamic mass-constrained neural networks using noisy transient data},
  year      = {2024},
  issn      = {0098-1354},
  month     = aug,
  pages     = {108722},
  volume    = {187},
  comment   = {Only mass constrained!!},
  doi       = {https://doi.org/10.1016/j.compchemeng.2024.108722},
  publisher = {Elsevier BV},
}

@Article{Konstantinov2023_NewComputationallySimple,
  author    = {Konstantinov, A. V. and Utkin, L. V.},
  journal   = {Doklady Mathematics},
  title     = {A New Computationally Simple Approach for Implementing Neural Networks with Output Hard Constraints},
  year      = {2023},
  issn      = {1531-8362},
  month     = dec,
  number    = {S2},
  pages     = {S233--S241},
  volume    = {108},
  comment   = {Very mathematical paper. They claim to do hard constraint with CONVEX constraint. They demonstrate with linear and quadratic constraints. They claim that the method is very computationally cheap. Apparently, they mainly talk about INEQUALITY constraints. Then they achieve equalities by using auxiliary variables.},
  doi       = {https://doi.org/10.1134/S1064562423701077},
  publisher = {Pleiades Publishing Ltd},
}

@Article{Frerix2020_HomogeneousLinearInequality,
  author    = {Frerix, Thomas and Niesner, Matthias and Cremers, Daniel},
  journal   = {2020 IEEE/CVF Conference on Computer Vision and Pattern Recognition Workshops (CVPRW)},
  title     = {Homogeneous Linear Inequality Constraints for Neural Network Activations},
  year      = {2020},
  month     = jun,
  pages     = {3229--3234},
  volume    = {521},
  comment   = {They enforce linear INEQUALITY constraints of the form Ax>=b.
They have some interesting perspective on constrained generative models (like variational autoencoders).},
  doi       = {10.1109/cvprw50498.2020.00382},
  publisher = {IEEE},
}

@Article{Chen2021_Theoryguidedhard,
  author    = {Chen, Yuntian and Huang, Dou and Zhang, Dongxiao and Zeng, Junsheng and Wang, Nanzhe and Zhang, Haoran and Yan, Jinyue},
  journal   = {Journal of Computational Physics},
  title     = {Theory-guided hard constraint projection (HCP): A knowledge-based data-driven scientific machine learning method},
  year      = {2021},
  issn      = {0021-9991},
  month     = nov,
  pages     = {110624},
  volume    = {445},
  comment   = {Very interesting, they use same concept as KKT-hPINN but on differential equations (discretized). They use projection. They claim that is exacteven if they satisfy a discretized version (that is, an approximation by definition). However, they can sample MORE collocation points and CONTROL the approximation degree (smaller delta). This is not doable for our taylor method.},
  doi       = {https://doi.org/10.1016/j.jcp.2021.110624},
  publisher = {Elsevier BV},
}

@Article{Schweidtmann2021_Obeyvaliditylimits,
  author    = {Schweidtmann, Artur M. and Weber, Jana M. and Wende, Christian and Netze, Linus and Mitsos, Alexander},
  journal   = {Optimization and Engineering},
  title     = {Obey validity limits of data-driven models through topological data analysis and one-class classification},
  year      = {2021},
  issn      = {1573-2924},
  month     = may,
  number    = {2},
  pages     = {855--876},
  volume    = {23},
  doi       = {https://doi.org/10.1007/s11081-021-09608-0},
  publisher = {Springer Science and Business Media LLC},
}

@Article{Wang2024_LinSATNetPositiveLinear,
  author    = {Wang, Runzhong and Zhang, Yunhao and Guo, Ziao and Chen, Tianyi and Yang, Xiaokang and Yan, Junchi},
  journal   = {In Proceedings of the 40th International Conference on Machine Learning (ICML'23)},
  title     = {LinSATNet: The Positive Linear Satisfiability Neural Networks},
  year      = {2024},
  copyright = {Creative Commons Attribution 4.0 International},
  doi       = {https://doi.org/10.48550/arXiv.2407.13917},
  publisher = {arXiv},
}

@Article{Kotary2021_EndEndConstrained,
  author    = {Kotary, James and Fioretto, Ferdinando and Van Hentenryck, Pascal and Wilder, Bryan},
  journal   = {Proceedings of the Thirtieth International Joint Conference on Artificial Intelligence (IJCAI-21)},
  title     = {End-to-End Constrained Optimization Learning: A Survey},
  year      = {2021},
  copyright = {Creative Commons Attribution 4.0 International},
  doi       = {https://doi.org/10.48550/arXiv.2103.16378},
  publisher = {arXiv},
}

@Article{Anil2018_SortingoutLipschitz,
  author    = {Anil, Cem and Lucas, James and Grosse, Roger},
  journal   = {Proceedings of the 36th International Conference on Machine Learning, Long Beach, California, PMLR 97, 2019},
  title     = {Sorting out Lipschitz function approximation},
  year      = {2018},
  copyright = {arXiv.org perpetual, non-exclusive license},
  doi       = {https://doi.org/10.48550/arXiv.1811.05381},
  publisher = {arXiv},
}

@Article{Agrawal2019_DifferentiableConvexOptimization,
  author    = {Agrawal, Akshay and Amos, Brandon and Barratt, Shane and Boyd, Stephen and Diamond, Steven and Kolter, Zico},
  journal   = {33rd Conference on Neural Information Processing Systems (NeurIPS 2019), Vancouver, Canada},
  title     = {Differentiable Convex Optimization Layers},
  year      = {2019},
  comment   = {Extension to OptNet (Amos and Kolter, 2017). Please see comments there. Here they basically expand to any convex program (not only quadratic).},
  copyright = {arXiv.org perpetual, non-exclusive license},
  doi       = {https://doi.org/10.48550/arXiv.1910.12430},
  publisher = {arXiv},
}

@Article{Min2024_HardConstrainedNeural,
  author    = {Min, Youngjae and Sonar, Anoopkumar and Azizan, Navid},
  journal   = {ArXiv preprint},
  title     = {Hard-Constrained Neural Networks with Universal Approximation Guarantees},
  year      = {2024},
  copyright = {Creative Commons Attribution 4.0 International},
  doi       = {https://doi.org/10.48550/arXiv.2410.10807},
  publisher = {arXiv},
}

@Article{Balestriero2022_POLICEProvablyOptimal,
  author    = {Balestriero, Randall and LeCun, Yann},
  journal   = {arXiv preprint},
  title     = {POLICE: Provably Optimal Linear Constraint Enforcement for Deep Neural Networks},
  year      = {2022},
  copyright = {Creative Commons Attribution 4.0 International},
  doi       = {https://doi.org/10.48550/arXiv.2211.01340},
  publisher = {arXiv},
}

@Article{Tordesillas2023_RAYENImpositionHard,
  author    = {Tordesillas, Jesus and How, Jonathan P. and Hutter, Marco},
  journal   = {ArXiv preprint},
  title     = {RAYEN: Imposition of Hard Convex Constraints on Neural Networks},
  year      = {2023},
  copyright = {arXiv.org perpetual, non-exclusive license},
  doi       = {https://doi.org/10.48550/arXiv.2307.08336},
  publisher = {arXiv},
}

@Article{Cybenko1989_Approximationsuperpositionssigmoidal,
  author    = {Cybenko, G.},
  journal   = {Mathematics of Control, Signals, and Systems},
  title     = {Approximation by superpositions of a sigmoidal function},
  year      = {1989},
  issn      = {1435-568X},
  month     = dec,
  number    = {4},
  pages     = {303--314},
  volume    = {2},
  doi       = {https://doi.org/10.1007/BF02551274},
  publisher = {Springer Science and Business Media LLC},
}

@Article{Hornik1989_Multilayerfeedforwardnetworks,
  author    = {Hornik, Kurt and Stinchcombe, Maxwell and White, Halbert},
  journal   = {Neural Networks},
  title     = {Multilayer feedforward networks are universal approximators},
  year      = {1989},
  issn      = {0893-6080},
  month     = jan,
  number    = {5},
  pages     = {359--366},
  volume    = {2},
  doi       = {https://doi.org/10.1016/0893-6080(89)90020-8},
  publisher = {Elsevier BV},
}

@Article{Rahaman2018_SpectralBiasNeural,
  author    = {Rahaman, Nasim and Baratin, Aristide and Arpit, Devansh and Draxler, Felix and Lin, Min and Hamprecht, Fred A. and Bengio, Yoshua and Courville, Aaron},
  journal   = {ArXiv},
  title     = {On the Spectral Bias of Neural Networks},
  year      = {2018},
  copyright = {arXiv.org perpetual, non-exclusive license},
  doi       = {https://doi.org/10.48550/arXiv.1806.08734},
  publisher = {arXiv},
}

@Article{Wang2021_Learningsolutionoperator,
  author    = {Wang, Sifan and Wang, Hanwen and Perdikaris, Paris},
  title     = {Learning the solution operator of parametric partial differential equations with physics-informed DeepOnets},
  year      = {2021},
  journal   = {ArXiv},
  copyright = {arXiv.org perpetual, non-exclusive license},
  doi       = {https://doi.org/10.48550/arXiv.2103.10974},
  publisher = {arXiv},
}

@Article{Bronstein2017_GeometricDeepLearning,
  author    = {Bronstein, Michael M. and Bruna, Joan and LeCun, Yann and Szlam, Arthur and Vandergheynst, Pierre},
  journal   = {IEEE Signal Processing Magazine},
  title     = {Geometric Deep Learning: Going beyond Euclidean data},
  year      = {2017},
  issn      = {1558-0792},
  month     = jul,
  number    = {4},
  pages     = {18--42},
  volume    = {34},
  doi       = {https://doi.org/10.1109/MSP.2017.2693418},
  publisher = {Institute of Electrical and Electronics Engineers (IEEE)},
}

@Article{Wu2021_ComprehensiveSurveyGraph,
  author    = {Wu, Zonghan and Pan, Shirui and Chen, Fengwen and Long, Guodong and Zhang, Chengqi and Yu, Philip S.},
  journal   = {IEEE Transactions on Neural Networks and Learning Systems},
  title     = {A Comprehensive Survey on Graph Neural Networks},
  year      = {2021},
  issn      = {2162-2388},
  month     = jan,
  number    = {1},
  pages     = {4--24},
  volume    = {32},
  doi       = {https://doi.org/10.1109/TNNLS.2020.2978386},
  publisher = {Institute of Electrical and Electronics Engineers (IEEE)},
}

@Article{LeCun1989_BackpropagationAppliedHandwritten,
  author    = {LeCun, Y. and Boser, B. and Denker, J. S. and Henderson, D. and Howard, R. E. and Hubbard, W. and Jackel, L. D.},
  journal   = {Neural Computation},
  title     = {Backpropagation Applied to Handwritten Zip Code Recognition},
  year      = {1989},
  issn      = {1530-888X},
  month     = dec,
  number    = {4},
  pages     = {541--551},
  volume    = {1},
  doi       = {https://doi.org/10.1162/neco.1989.1.4.541},
  publisher = {MIT Press},
}

@Article{Wang2020_WhenwhyPINNs,
  author    = {Wang, Sifan and Yu, Xinling and Perdikaris, Paris},
  journal   = {ArXiv},
  title     = {When and why PINNs fail to train: A neural tangent kernel perspective},
  year      = {2020},
  copyright = {arXiv.org perpetual, non-exclusive license},
  doi       = {https://doi.org/10.48550/arXiv.2007.14527},
  publisher = {arXiv},
}

@Article{Wang2020_Understandingmitigatinggradient,
  author     = {Wang, Sifan and Teng, Yujun and Perdikaris, Paris},
  journal   = {ArXiv},
  title      = {Understanding and mitigating gradient pathologies in physics-informed neural networks},
  year       = {2020},
  copyright  = {arXiv.org perpetual, non-exclusive license},
  doi        = {https://doi.org/10.48550/arXiv.2001.04536},
  publisher  = {arXiv},
  ranking    = {rank5},
  readstatus = {read},
}

@Article{Brosowsky2020_SampleSpecificOutput,
  author    = {Brosowsky, Mathis and Dünkel, Olaf and Slieter, Daniel and Zöllner, Marius},
  journal   = {Proceedings of the AAAI Conference on Artificial Intelligence, 35(8), 6812-6821},
  title     = {Sample-Specific Output Constraints for Neural Networks},
  year      = {2020},
  comment   = {They do constraint (convex polytopes, a.k.a., linear) on the output. So, they are essentially inequality constraints. They need additional inputs to the NN to enforce the constraints. The additional input is some parameter characterizing the constraints.},
  copyright = {arXiv.org perpetual, non-exclusive license},
  doi       = {https://doi.org/10.48550/arXiv.2003.10258},
  publisher = {arXiv},
}

@Article{Waechter2005_implementationinteriorpoint,
  author    = {Wächter, Andreas and Biegler, Lorenz T.},
  journal   = {Mathematical Programming},
  title     = {On the implementation of an interior-point filter line-search algorithm for large-scale nonlinear programming},
  year      = {2005},
  issn      = {1436-4646},
  month     = apr,
  number    = {1},
  pages     = {25--57},
  volume    = {106},
  doi       = {10.1007/s10107-004-0559-y},
  publisher = {Springer Science and Business Media LLC},
}

@Article{Manek2020_LearningStableDeepa,
  author    = {Manek, Gaurav and Kolter, J. Zico},
  journal   = {33rd Conference on Neural Information Processing Systems (NeurIPS 2019), Vancouver, Canada},
  title     = {Learning Stable Deep Dynamics Models},
  year      = {2020},
  copyright = {arXiv.org perpetual, non-exclusive license},
  doi       = {https://doi.org/10.48550/arXiv.2001.06116},
  publisher = {arXiv},
}

@Article{Xu2021_Artificialintelligencepowerful,
  author    = {Xu, Yongjun and Liu, Xin and Cao, Xin and Huang, Changping and Liu, Enke and Qian, Sen and Liu, Xingchen and Wu, Yanjun and Dong, Fengliang and Qiu, Cheng-Wei and Qiu, Junjun and Hua, Keqin and Su, Wentao and Wu, Jian and Xu, Huiyu and Han, Yong and Fu, Chenguang and Yin, Zhigang and Liu, Miao and Roepman, Ronald and Dietmann, Sabine and Virta, Marko and Kengara, Fredrick and Zhang, Ze and Zhang, Lifu and Zhao, Taolan and Dai, Ji and Yang, Jialiang and Lan, Liang and Luo, Ming and Liu, Zhaofeng and An, Tao and Zhang, Bin and He, Xiao and Cong, Shan and Liu, Xiaohong and Zhang, Wei and Lewis, James P. and Tiedje, James M. and Wang, Qi and An, Zhulin and Wang, Fei and Zhang, Libo and Huang, Tao and Lu, Chuan and Cai, Zhipeng and Wang, Fang and Zhang, Jiabao},
  journal   = {The Innovation},
  title     = {Artificial intelligence: A powerful paradigm for scientific research},
  year      = {2021},
  issn      = {2666-6758},
  month     = nov,
  number    = {4},
  pages     = {100179},
  volume    = {2},
  doi       = {https://doi.org/10.1016/j.xinn.2021.100179},
  publisher = {Elsevier BV},
}

@Article{Wang2023_Scientificdiscoveryage,
  author    = {Wang, Hanchen and Fu, Tianfan and Du, Yuanqi and Gao, Wenhao and Huang, Kexin and Liu, Ziming and Chandak, Payal and Liu, Shengchao and Van Katwyk, Peter and Deac, Andreea and Anandkumar, Anima and Bergen, Karianne and Gomes, Carla P. and Ho, Shirley and Kohli, Pushmeet and Lasenby, Joan and Leskovec, Jure and Liu, Tie-Yan and Manrai, Arjun and Marks, Debora and Ramsundar, Bharath and Song, Le and Sun, Jimeng and Tang, Jian and Veličković, Petar and Welling, Max and Zhang, Linfeng and Coley, Connor W. and Bengio, Yoshua and Zitnik, Marinka},
  journal   = {Nature},
  title     = {Scientific discovery in the age of artificial intelligence},
  year      = {2023},
  issn      = {1476-4687},
  month     = aug,
  number    = {7972},
  pages     = {47--60},
  volume    = {620},
  doi       = {https://doi.org/10.1038/s41586-023-06221-2},
  publisher = {Springer Science and Business Media LLC},
}

@Article{Schweidtmann2018_DeterministicGlobalOptimization,
  author    = {Schweidtmann, Artur M. and Mitsos, Alexander},
  journal   = {Journal of Optimization Theory and Applications},
  title     = {Deterministic Global Optimization with Artificial Neural Networks Embedded},
  year      = {2018},
  issn      = {1573-2878},
  month     = oct,
  number    = {3},
  pages     = {925--948},
  volume    = {180},
  doi       = {https://doi.org/10.1007/s10957-018-1396-0},
  publisher = {Springer Science and Business Media LLC},
}

@Article{Mize2019_BestPracticesEstimating,
  author    = {Mize, Trenton},
  journal   = {Sociological Science},
  title     = {Best Practices for Estimating, Interpreting, and Presenting Nonlinear Interaction Effects},
  year      = {2019},
  issn      = {2330-6696},
  pages     = {81--117},
  volume    = {6},
  doi       = {10.15195/v6.a4},
  publisher = {Society for Sociological Science},
}

@Book{Nicolis1995_IntroductionNonlinearScience,
  author    = {Nicolis, G.},
  publisher = {Cambridge University Press},
  title     = {Introduction to Nonlinear Science},
  year      = {1995},
  isbn      = {9781139170802},
  month     = jun,
  doi       = {https://doi.org/10.1017/CBO9781139170802},
}

@Article{Fletcher2002_Nonlinearprogrammingpenalty,
  author    = {Fletcher, Roger and Leyffer, Sven},
  journal   = {Mathematical Programming},
  title     = {Nonlinear programming without a penalty function},
  year      = {2002},
  issn      = {1436-4646},
  month     = jan,
  number    = {2},
  pages     = {239--269},
  volume    = {91},
  doi       = {https://doi.org/10.1007/s101070100244},
  publisher = {Springer Science and Business Media LLC},
}

@Article{Fletcher2002_GlobalConvergenceFilter,
  author    = {Fletcher, Roger and Leyffer, Sven and Toint, Philippe L.},
  journal   = {SIAM Journal on Optimization},
  title     = {On the Global Convergence of a Filter--SQP Algorithm},
  year      = {2002},
  issn      = {1095-7189},
  month     = jan,
  number    = {1},
  pages     = {44--59},
  volume    = {13},
  doi       = {https://doi.org/10.1137/S105262340038081X},
  publisher = {Society for Industrial & Applied Mathematics (SIAM)},
}

@Article{Pfrommer2020_ContactNetsLearningDiscontinuousa,
  author    = {Pfrommer, Samuel and Halm, Mathew and Posa, Michael},
  journal   = {Conference on Robot Learning 2020},
  title     = {ContactNets: Learning Discontinuous Contact Dynamics with Smooth, Implicit Representations},
  year      = {2020},
  copyright = {arXiv.org perpetual, non-exclusive license},
  doi       = {https://doi.org/10.48550/arXiv.2009.11193},
  publisher = {arXiv},
}

@Article{Erichson2019_PhysicsinformedAutoencoders,
  author    = {Erichson, N. Benjamin and Muehlebach, Michael and Mahoney, Michael W.},
  journal   = {Second Workshop on Machine Learning and the Physical Sciences (NeurIPS 2019), Vancouver, Canada},
  title     = {Physics-informed Autoencoders for Lyapunov-stable Fluid Flow Prediction},
  year      = {2019},
  copyright = {arXiv.org perpetual, non-exclusive license},
  doi       = {https://doi.org/10.48550/arXiv.1905.10866},
  publisher = {arXiv},
}

@Article{Schweidtmann2024_reviewperspectivehybrida,
  author    = {Schweidtmann, Artur M. and Zhang, Dongda and von Stosch, Moritz},
  journal   = {Digital Chemical Engineering},
  title     = {A review and perspective on hybrid modeling methodologies},
  year      = {2024},
  issn      = {2772-5081},
  month     = mar,
  pages     = {100136},
  volume    = {10},
  doi       = {https://doi.org/10.1016/j.dche.2023.100136},
  publisher = {Elsevier BV},
}

@Article{Kingma2014_AdamMethodStochastic,
  author    = {Kingma, Diederik P. and Ba, Jimmy},
  journal   = {3rd International Conference for Learning Representations, San Diego, 2015},
  title     = {Adam: A Method for Stochastic Optimization},
  year      = {2014},
  copyright = {arXiv.org perpetual, non-exclusive license},
  doi       = {https://doi.org/10.48550/arXiv.1412.6980},
  publisher = {arXiv},
}

@Article{Stoian2024_HowRealisticIs,
  author    = {Stoian, Mihaela Cătălina and Dyrmishi, Salijona and Cordy, Maxime and Lukasiewicz, Thomas and Giunchiglia, Eleonora},
  journal   = {Published as a conference paper at ICLR 2024},
  title     = {How Realistic Is Your Synthetic Data? Constraining Deep Generative Models for Tabular Data},
  year      = {2024},
  copyright = {arXiv.org perpetual, non-exclusive license},
  doi       = {https://doi.org/10.48550/arXiv.2402.04823},
  publisher = {arXiv},
}

@Article{Giunchiglia2021_MultiLabelClassification,
  author    = {Giunchiglia, Eleonora and Lukasiewicz, Thomas},
  journal   = {Journal of Artificial Intelligence Research},
  title     = {Multi-Label Classification Neural Networks with Hard Logical Constraints},
  year      = {2021},
  issn      = {1076-9757},
  month     = nov,
  pages     = {759--818},
  volume    = {72},
  doi       = {https://doi.org/10.1613/jair.1.12850},
  publisher = {AI Access Foundation},
}

@Article{Lastrucci2025_PicardKKThPINN,
  author     = {Lastrucci, Giacomo and Karia, Tanuj and Gromotka, Zoë and Schweidtmann, Artur M.},
  journal    = {Systems and Control Transactions},
  title      = {Picard-KKT-hPINN: Enforcing Nonlinear Enthalpy Balances for Physically Consistent Neural Networks},
  year       = {2025},
  issn       = {2818-4734},
  month      = {Jul},
  pages      = {1718--1723},
  volume     = {4},
  booktitle  = {Proceedings of the 35th European Symposium on Computer Aided Process Engineering (ESCAPE 35)},
  collection = {ESCAPE 35},
  doi        = {https://doi.org/10.69997/sct.108423},
  publisher  = {PSE Press},
  series     = {ESCAPE 35},
  url        = {https://psecommunity.org/wp-content/plugins/wpor/includes/file/2506/LAPSE-2025.0428-1v1.pdf},
}

@Book{Nocedal2006_NumericalOptimization,
  author    = {Jorge Nocedal and Stephen J. Wright},
  publisher = {Springer},
  title     = {Numerical Optimization},
  year      = {2006},
  address   = {New York, NY},
  edition   = {2},
  isbn      = {978-0-387-30303-1},
  series    = {Springer Series in Operations Research and Financial Engineering},
  doi       = {10.1007/978-0-387-40065-5},
  pages     = {XXII, 664},
  url       = {https://doi.org/10.1007/978-0-387-40065-5},
}

@Article{DiVito2024_LearningSolveDifferential,
  author    = {Di Vito, Vincenzo and Mohammadian, Mostafa and Baker, Kyri and Fioretto, Ferdinando},
  journal   = {arXiv preprint},
  title     = {Learning To Solve Differential Equation Constrained Optimization Problems},
  year      = {2024},
  copyright = {Creative Commons Attribution 4.0 International},
  doi       = {https://doi.org/10.48550/arXiv.2410.01786},
  publisher = {arXiv},
}

@book{burden2005numerical,
  title     = {Numerical Analysis},
  author    = {Burden, Richard L. and Faires, J. Douglas},
  year      = {2005},
  edition   = {8},
  publisher = {Thomson Brooks/Cole},
  address   = {Belmont, CA},
  isbn      = {9780534404994}
}

@Article{ESamadi2022_trainingstrategyhybrid,
  author    = {E. Samadi, Moein and Kiefer, Sandra and Fritsch, Sebastian Johaness and Bickenbach, Johannes and Schuppert, Andreas},
  journal   = {PLOS ONE},
  title     = {A training strategy for hybrid models to break the curse of dimensionality},
  year      = {2022},
  issn      = {1932-6203},
  month     = sep,
  number    = {9},
  pages     = {e0274569},
  volume    = {17},
  doi       = {https://doi.org/10.1371/journal.pone.0274569},
  editor    = {Chen, Chi-Hua},
  publisher = {Public Library of Science (PLoS)},
}

@InProceedings{Fischer2019_DL2TrainingQuerying,
  author    = {Fischer, Marc and Balunovic, Mislav and Drachsler-Cohen, Dana and Gehr, Timon and Zhang, Ce and Vechev, Martin},
  booktitle = {Proceedings of the 36th International Conference on Machine Learning},
  title     = {{DL}2: Training and Querying Neural Networks with Logic},
  year      = {2019},
  editor    = {Chaudhuri, Kamalika and Salakhutdinov, Ruslan},
  month     = {09--15 Jun},
  pages     = {1931--1941},
  publisher = {PMLR},
  series    = {Proceedings of Machine Learning Research},
  volume    = {97},
  url       = {https://proceedings.mlr.press/v97/fischer19a.html},
}

@Article{Tao2023_ArchitecturePreservingProvable,
  author    = {Tao, Zhe and Nawas, Stephanie and Mitchell, Jacqueline and Thakur, Aditya V.},
  title     = {Architecture-Preserving Provable Repair of Deep Neural Networks},
  journal   = {arXiv preprint},
  year      = {2023},
  copyright = {Creative Commons Attribution Non Commercial No Derivatives 4.0 International},
  doi       = {https://doi.org/10.48550/arXiv.2304.03496},
  publisher = {arXiv},
}

@InProceedings{Tao2024_ProvableEditingDeep,
  author    = {Tao, Zhe and Thakur, Aditya V.},
  booktitle = {Advances in Neural Information Processing Systems},
  title     = {Provable Editing of Deep Neural Networks using Parametric Linear Relaxation},
  year      = {2024},
  url       = {https://openreview.net/forum?id=IGhpUd496D},
}

@Article{Iftakher2025_PhysicsInformedNeural,
  author    = {Iftakher, Ashfaq and Golder, Rahul and Roy, Bimol Nath and Faruque Hasan, M.M.},
  journal   = {Computers \& Chemical Engineering},
  title     = {Physics-informed neural networks with hard nonlinear equality and inequality constraints},
  year      = {2026},
  issn      = {0098-1354},
  month     = jan,
  pages     = {109418},
  volume    = {204},
  copyright = {arXiv.org perpetual, non-exclusive license},
  doi       = {https://doi.org/10.1016/j.compchemeng.2025.109418},
  publisher = {Elsevier BV},
}

@Book{Floudas1999_HandbookTestProblems,
  author    = {Floudas, Christodoulos A. and Pardalos, Pãnos M. and Adjiman, Claire S. and Esposito, William R. and Gümüş, Zeynep H. and Harding, Stephen T. and Klepeis, John L. and Meyer, Clifford A. and Schweiger, Carl A.},
  publisher = {Springer US},
  title     = {Handbook of Test Problems in Local and Global Optimization},
  year      = {1999},
  isbn      = {9781475730401},
  doi       = {https://doi.org/10.1007/978-1-4757-3040-1},
  issn      = {1571-568X},
  journal   = {Nonconvex Optimization and Its Applications},
}

\linenumbers
\appendix

\section*{Appendix}
\section{Projection operator and geometry}
\label{app:proj_oper_geom}
This appendix collects (i) the closed-form expression of the \emph{linearized} projection used in a neural projection (NP) layer, and (ii) short geometric remarks clarifying what is and is not guaranteed in the nonlinear case.

\subsection{Closed-form neural projection}
\label{app:clos_form}
We derive here the closed-form expression defining a neural projection layer in Def.~\ref{def:NPlayer} (Section~\ref{subsec:adanp}). Given the linearized projection problem in Eq.\eqref{eq:linearized_projection}, we can define the Lagrangian function as:

\begin{equation}
\label{eq:linearized_lagrangian}
\mathcal{L}(x, y, \lambda) = \frac{1}{2} (y - \hat{y})^T (y - \hat{y}) + \lambda^T\left( c(x,\hat{y}) + \left.J_yc\right|_{x, \hat{y}} (y - \hat{y})\right)
\end{equation}

Then, a local optimum can be found by solving the KKT conditions (i.e., primal and dual feasibility):

\begin{equation}
\label{eq:linearized_feasibility}
\begin{aligned}
&\nabla_y\mathcal{L} = (y - \hat{y}) + \left.J^T_yc\right|_{x, \hat{y}} \lambda = 0 \\
&c(x,\hat{y}) + \left.J_yc\right|_{x, \hat{y}} (y - \hat{y}) = 0 
\end{aligned}
\end{equation}

To simplify the notation, we define the linear system as:

\begin{equation}
\label{eq:simplified_system}
\begin{aligned}
(1) \quad &(y - \hat{y}) + B^T \lambda = 0 \\
(2) \quad &By-v = 0 
\end{aligned}
\end{equation}

where:
\begin{equation}
\label{eq:definitionBv}
\begin{aligned}
B &= \left.J_yc\right|_{x, \hat{y}} \in \mathbb{R}^{N_C \times N_O}\\
v &= \left.J_yc\right|_{x, \hat{y}} \hat{y} - c(x, \hat{y}) \in \mathbb{R}^{N_C}
\end{aligned}
\end{equation}

Solving the system, we obtain a closed form for the neural projection layer:




\begin{equation}
\label{eq:closed_form}
\tilde{y}=(I - B^T(BB^T)^{-1}B) \hat{y}+ B^T(BB^T)^{-1}v
\end{equation}



\subsection{Geometric remarks: affine vs.\ nonlinear constraints}
\label{app:geom_remarks}

The results in Proposition~\ref{prop:nonexpansive_affine} concern the \emph{Euclidean projection onto an affine constraint set} (constraints affine in $y$). In that case the feasible set is a closed convex affine subspace and the (metric) projection is single-valued and non-expansive.

For nonlinear constraints, AdaNP does \emph{not} compute the metric projection onto the nonlinear manifold
\[
\mathcal{M}(x) := \{y\in\mathbb{R}^{N_O}: c(x,y)=0\}.
\]
Instead, each NP layer projects onto a \emph{locally linearized} constraint set,
\[
\mathcal{M}_{\text{lin}}(x,\hat{y}) := \{y: c(x,\hat{y}) + J_y c(x,\hat{y})(y-\hat{y}) = 0\},
\]
which is an affine subspace. Consequently:

\begin{itemize}
    \item \textbf{Per-step well-posedness.} Under the standard full-row-rank condition on $B=\left.J_yc\right|_{x,\hat{y}}$ (LICQ for the linearization), the linearized QP has a unique minimizer given by Eq.~\eqref{eq:closed_form}.
    \item \textbf{No global uniqueness claim for the nonlinear manifold.} The (true) metric projection onto $\mathcal{M}(x)$ may fail to be unique far from the manifold, and even locally uniqueness requires additional geometric regularity (e.g., positive reach / tubular neighborhood conditions). We do not rely on global uniqueness of the metric projection in the method or in the theory.
    \item \textbf{No ``no-worse'' guarantee relative to ground truth.} For nonlinear manifolds, projecting $\hat{y}$ onto a local linearization can move the prediction away from the (unknown) ground-truth $y^*$, even if $y^*\in\mathcal{M}(x)$. Thus, AdaNP does not guarantee monotone improvement in task error; this motivates the activation heuristic in Algorithm~\ref{algorithm:adanp_activation}.
\end{itemize}

\subsection{Gradient flow through an affine projection layer}
\label{app:affine_grad_flow}
This short note complements Proposition~\ref{prop:nonexpansive_affine}. If the feasible set is affine in $y$, the Euclidean projection $\mathcal{P}$ is (almost everywhere) differentiable with Jacobian equal to an orthogonal projector $P$ satisfying $\|P\|_2 \le 1$. Therefore, for any scalar loss $\ell$ depending on $\tilde{y}=\mathcal{P}(\hat{y})$,
\[
\nabla_{\hat{y}}\ell = J_{\hat{y}}\mathcal{P}^\top \nabla_{\tilde{y}}\ell
\quad\Rightarrow\quad
\|\nabla_{\hat{y}}\ell\| \le \|J_{\hat{y}}\mathcal{P}\|_2 \, \|\nabla_{\tilde{y}}\ell\|
\le \|\nabla_{\tilde{y}}\ell\|.
\]
This bound holds for the affine case addressed by Proposition~\ref{prop:nonexpansive_affine}; it does not imply global non-expansiveness for nonlinear constraints.

\section{Convergence analysis of AdaNP}
\subsection{Deviation from Newton's method}
\label{app:deviation_newton}
To support the discussion raised in Section~\ref{subsec:adanp}, we show how our method deviates from standard Newton's method for solving nonlinear KKT conditions. Given a nonlinear program:

\begin{equation}
\label{eq:nlp}
\begin{aligned}
\tilde{y} = &\operatorname*{arg\,min}_{y} \frac{1}{2} ||y - \hat{y}||^2 \\
\text{s.t.} & \quad  c(x, y) = 0
\end{aligned}
\end{equation}

With associated Lagrangian function:

\begin{equation}
\label{eq:nlp_lagrangian}
\mathcal{L}(x, y, \lambda) = \frac{1}{2} (y - \hat{y})^T (y - \hat{y}) + \lambda^T c(x,y)
\end{equation}

The primal and dual feasibility can be derived as:

\begin{equation}
\label{eq:nlp_feasibility}
\begin{aligned}
&\nabla_y\mathcal{L} = (y - \hat{y}) + J^T_y c(x,y) \lambda = 0 \\
&c(x,y) = 0 
\end{aligned}
\end{equation}

Linearizing the system according to Newton's iteration at $y=y_0$ results in:

\begin{equation}
\begin{aligned}
&(y_0 - \hat{y}) + J^T_y c|_{y_0, \lambda_0} \lambda_0 + (y-y_0) + \lambda^T \left. \mathbf{H}_y c \right|_{y_0, \lambda_0} (y-y_0) + \left.J^T_y c\right|_{y_0, \lambda_0} (\lambda-\lambda_0) = 0 \\
&c(x, y_0) + J_y c|_{y_0, \lambda_0} (y-y_0) = 0 
\end{aligned}
\end{equation}

Thus, assuming to center the linearization in the neural network prediction, i.e., $y_0 = \hat{y}$, and choosing $\lambda_0 = 0$:

\begin{equation}
\begin{aligned}
& (y-\hat{y}) + \lambda^T \left. \mathbf{H}_y c \right|_{\hat{y}, 0} (y-\hat{y}) + \left. J^T_y c \right|_{\hat{y}, 0} \lambda = 0 \\
&c(x, \hat{y}) + \left. J_y c\right|_{\hat{y}, 0} (y-\hat{y}) = 0 
\end{aligned}
\end{equation}

We can conclude that, essentially, our NP layer solves a similar linear system (Eq.\eqref{eq:linearized_feasibility}) which does not comprise the term $\lambda^T \left. \mathbf{H}_y c \right|_{\hat{y}, 0} (y-\hat{y})$, hence avoiding the computation of the Hessian tensor $\mathbf{H}_yc$. Here, we note some similarity with Gauss-Newton methods used to solve least square problems~\cite{Nocedal2006_NumericalOptimization}.

\subsection{Local convergence rate}
\label{app:local_convergence_rate}
The projection operator $\mathcal{P}$ solves an SQP subproblem of the form:
\begin{equation}
\label{eq:incomplete_SQP_step}
\begin{aligned}
    \min_{y} & \frac{1}{2} (y - \hat{y})^T \bar{H} (y - \hat{y}) \\
    \text{s.t.} & \quad  J_yc \, (y - \hat{y}) + c(x, \hat{y}) = 0,
\end{aligned}
\end{equation}

which approximates the original nonlinear program:
\begin{equation}
\label{eq:complete_SQP_step}
\begin{aligned}
    \min_{y} & \frac{1}{2} (y - \hat{y})^T I (y - \hat{y}) \\
    \text{s.t.} & \quad  c(x,y) = 0
\end{aligned}
\end{equation}

The Hessian $\bar{H}$, as observed in the deviation from Newton's method~\ref{app:deviation_newton}, does not include the second-order derivatives of the constraints that would appear in the full Lagrangian Hessian. The resulting method is often called \textit{Gauss-Newton SQP step}, since the way the constraints derivatives are dropped reminds of the Gauss-Newton method for nonlinear least squares~\cite{Nocedal2006_NumericalOptimization}.\\
Supported by SQP theory~\cite{Nocedal2006_NumericalOptimization}, conditions for local convergence can be derived. We assume $y^*$ to be a local solution to the original nonlinear program~(Eq.\ref{eq:weighted_nonlinear_projection}) at which the following conditions hold~\cite{Nocedal2006_NumericalOptimization}:
\begin{enumerate}
    \item[H1] The objective function and the constraints are twice differentiable in a neighborhood of $y^*$ with Lipschitz continuous second derivatives.
    \item[H2] The linear independence constraint qualification (LICQ) holds at $y^*$. Then, the KKT conditions are satisfied for a vector of Lagrangian multipliers $\lambda^*$.
    \item[H3] The second-order sufficient conditions (SOSC) hold at ($y^*$, $\lambda^*$).
\end{enumerate}
The KKT conditions for the original nonlinear program are defined as:
\begin{equation}
    F(z) = \begin{bmatrix} \nabla_y\mathcal{L}(y,\lambda) \\ c(y) \end{bmatrix}, \quad\text{with}\quad  z = \begin{bmatrix} y \\ \lambda \end{bmatrix},
\end{equation}
and are satisfied by a vector $z^* = \begin{bmatrix} y^* &\lambda ^*\end{bmatrix}^T$.\\
We define the Jacobian of the KKT conditions of the original nonlinear program (Eq.\eqref{eq:complete_SQP_step}) in a neighborhood of the local solution as:
\begin{equation}
    J^{(k)}=\begin{bmatrix}
        \nabla_{yy}^2\mathcal{L}^{(k)} & J_c^T \\ J_c & 0
    \end{bmatrix}
\end{equation}

We assume that LICQ and SOQC hold also in the neighborhood of $z^*$ (H2 and H3), hence the Jacobian at iteration $k$, $J^{(k)}$, is non-singular and thus invertible.

The deviation of the projection operator $\mathcal{P}$ from the complete SQP step can be expressed through a matrix $E$ holding the second-order derivatives of the constraints:
\begin{equation}
    E=\begin{bmatrix}
        \sum_i\lambda^{(k)}_i\nabla^2c_i(y^{(k)}) & 0 \\ 0 & 0
    \end{bmatrix},
\end{equation}
such that $J^{(k)}=\bar{J} + E$, with $\bar{J}$ being the Jacobian of the KKT conditions associated with the problem in Eq.\eqref{eq:incomplete_SQP_step}.\\
At iteration $k$, we define the residual $r^{(k)} = F(z^{(k)})$, the error $e^{(k)} = z^{(k)} - z^*$ and solve for the Newton's step $s^{(k)}$:
\begin{equation}
    \begin{aligned}
        &\bar{J}s^{(k)} = -r^{(k)} \quad \quad \text{(QP solve, Newton step)}\\
        & z^{(k+1)} = z^{(k)} + s^{(k)} \\
        & e^{(k+1)} = e^{(k)} + s^{(k)}    
    \end{aligned}
\end{equation}
Since $F(z)$ is twice continuously differentiable, using Taylor expansion:
\begin{equation}
    F(z^{(k)}) = J^{(k)}e^{(k)} + r^{(k)}, \quad \quad \text{with} \quad r^{(k)} = \mathcal{O}(||e^{(k)}||^2)
\end{equation}
Thus, a QP step can be expressed as:
\begin{equation}
    \bar{J}s^{(k)} = -r^{(k)} = -F(z^{(k)}) = -J^{(k)}e^{(k)} - r^{(k)}
\end{equation}
From the definition of the Jacobian $\bar{J}$ and given that $J^{(k)}$ is invertible:
\begin{equation}
    (J^{(k)}-E)s^{(k)}  = -J^{(k)}e^{(k)} - r^{(k)}
\end{equation}
\begin{equation}
    ((J^{(k)})^{-1}J^{(k)}-(J^{(k)})^{-1}E)s^{(k)}  = -(J^{(k)})^{-1}J^{(k)}e^{(k)} - (J^{(k)})^{-1} r^{(k)}
\end{equation}
\begin{equation}
    (I-M)s^{(k)}  = -e^{(k)} - (J^{(k)})^{-1}r^{(k)},
\end{equation}
\begin{equation}
    s^{(k)}  = -(I-M)^{-1}(e^{(k)} + (J^{(k)})^{-1}r^{(k)}),
\end{equation}

with $M = (J^{(k)})^{-1}E$. \\
Thus, in the neighborhood of the solution, the error at iteration $k+1$ can be expressed as:

\begin{equation}
    e^{(k+1)} = z^{(k+1)}- z^* = e^{(k)} + s^{(k)}  = e^{(k)} -(I-M)^{-1}(e^{(k)} + (J^{(k)})^{-1}r^{(k)})
\end{equation}

Rearranging:

\begin{equation}
\begin{aligned}
    e^{(k+1)} &=  (I-(I-M)^{-1}) e^{(k)} - (I-M)^{-1}(J^{(k)})^{-1}r^{(k)} \\
    &= (I-M)^{-1}((I-M)-I) e^{(k)} - (I-M)^{-1}(J^{(k)})^{-1}r^{(k)} \\
    &= -(I-M)^{-1}M e^{(k)} - (I-M)^{-1}(J^{(k)})^{-1}r^{(k)}
\end{aligned}
\end{equation}

Banach's lemma then gives:
\begin{equation}
    ||(I-M)^{-1}|| \leq \frac{1}{1-||M||} = \frac{1}{1-\rho} = C_0
\end{equation}

Then we can estimate the error:

\begin{equation}
    ||e^{(k+1)}|| \leq C_0( ||M|| \: ||e^{(k)}|| + ||(J^{(k)})^{-1}|| \: ||r^{(k)}||)
\end{equation}

Since $r^{(k)} = O(||e^{(k)}||^2)$, $\exists \:C_1>0 : ||r^{(k)}||\leq C_1||e^{(k)}||^2$, then:

\begin{equation}
    ||e^{(k+1)}|| \leq C_0||M|| \: ||e^{(k)}|| + C_0 ||(J^{(k)})^{-1}|| \:C_1 ||e^{(k)}||^2)
\end{equation}

\textbf{Remark 1} We can conclude that, in the neighborhood of the solution:
\begin{itemize}
    \item If $M = 0$, the linear term vanishes and yields quadratic convergence, i.e., when the constraints are affine and thus the second order derivative of the constraints vanish ($\nabla^2c_i=0$).
    \item If $M\neq0$ but $||M||<1$, it is guaranteed strictly linear convergence with rate $||M||$, plus a higher-order correction.
    \item If $||M|| \ge 1$, the Gauss–Newton step alone may not converge. Second-order corrections (or using the full Lagrangian Hessian) are then required.
\end{itemize}

\textbf{Remark 2}
Among state-of-the-art methods for constrained learning, Newton-based completion approaches can achieve quadratic local convergence of the \emph{completion solve} under standard regularity assumptions. However, their differentiable completion mapping involving Newton solves may lead to instabilities and degraded performances~\cite{Beucler2019_EnforcingAnalyticConstraints}.

\section{Conditioning and stability analysis}
\label{app:conditioning_stability}

This appendix discusses how the feasibility-recovery modules used in constrained learning affect stability during training and inference. We distinguish between (i) \emph{affine constraint sets}, where the Euclidean projection is globally non-expansive, and (ii) \emph{nonlinear constraints}, where AdaNP performs projections onto \emph{local linearizations} and no global non-expansiveness guarantee is available.

\subsection{Affine constraint sets: non-expansiveness and gradient flow}
\label{app:affine_nonexpansive}

In the affine-in-$y$ case, the feasible set can be written as
\[
\mathcal{M}=\{y\in\mathbb{R}^{N_O}: By=v\},
\]
with $B\in\mathbb{R}^{N_C\times N_O}$ full row rank. The Euclidean projection of $\hat{y}$ onto $\mathcal{M}$ admits the closed form (cf.\ Appendix~\ref{app:clos_form})
\begin{equation}
\label{eq:affine_proj_closed_form}
\tilde{y}= \mathcal{P}(\hat{y}) = P\hat{y} + q,
\qquad
P := I - B^\top(BB^\top)^{-1}B,
\qquad
q := B^\top(BB^\top)^{-1}v.
\end{equation}
The matrix $P$ is an orthogonal projector onto $\ker(B)$, hence $\|P\|_2 = 1$. This is consistent with the non-expansiveness property established in Proposition~\ref{prop:nonexpansive_affine}.

\paragraph{Implication for backpropagation.}
Let $\ell(\tilde{y})$ be a scalar loss with $\tilde{y}=\mathcal{P}(\hat{y})$. Since $J_{\hat{y}}\mathcal{P}=P$, the chain rule yields
\begin{equation} 
\label{eq:affine_grad_flow} 
\nabla_{\hat{y}}\ell = P^\top \nabla_{\tilde{y}}\ell = P\nabla_{\tilde{y}}\ell, \qquad \Rightarrow\qquad \|\nabla_{\hat{y}}\ell\| \le \|P\|_2 \,\|\nabla_{\tilde{y}}\ell\| \le \|\nabla_{\tilde{y}}\ell\|. 
\end{equation}
Thus, in the affine case, the projection layer does not amplify gradients in operator norm.

\subsection{Nonlinear constraints: local well-posedness and conditioning}
\label{app:nonlinear_conditioning}

For nonlinear constraints, each NP layer applies a projection onto a locally linearized constraint set with Jacobian $B = J_y c(x,\hat{y})$. The resulting step involves solving a linear system with matrix $BB^\top$.

The numerical sensitivity of this operation depends on the conditioning of $B$. In particular,
\begin{equation}
\|(BB^\top)^{-1}\|_2 = \frac{1}{\sigma_{\min}(B)^2},
\end{equation}
so small singular values of $B$ may lead to amplification of numerical errors.

This reflects the local nature of the method: stability depends on both proximity to the constraint manifold and the conditioning of the constraint Jacobian. This is consistent with the local convergence guarantees discussed in Appendix~\ref{app:local_convergence_rate}.

\subsection{Comparison with predict-and-complete layers}
\label{app:nullspace_conditioning}

Both AdaNP and predict-and-complete approaches rely on the constraint Jacobian and can be sensitive to ill-conditioning. The key difference is \emph{where} the inverse-Jacobian sensitivity enters the computational graph.

\paragraph{Predict-and-complete (null-space) mapping.}
Predict-and-complete methods partition the output as $y=(y_F,y_C)$, where $y_F\in\mathbb{R}^{N_O-N_C}$ are free variables predicted by a network and $y_C\in\mathbb{R}^{N_C}$ are obtained by solving
\[
c(x,y_F,y_C)=0.
\]
Under the implicit function theorem assumptions, a local completion mapping $y_C=\varphi_x(y_F)$ exists and is differentiable, with Jacobian
\begin{equation}
\label{eq:ift_jacobian}
J_{y_F}\varphi_x(y_F) = -\left(J_{y_C}c(x,y)\right)^{-1} J_{y_F}c(x,y),
\end{equation}
evaluated at $y=(y_F,\varphi_x(y_F))$. Consequently, a local sensitivity bound is
\begin{equation}
\label{eq:nullspace_lip_bound}
\|J_{y_F}\varphi_x\|_2
\le
\|(J_{y_C}c)^{-1}\|_2 \,\|J_{y_F}c\|_2
=
\frac{\|J_{y_F}c\|_2}{\sigma_{\min}(J_{y_C}c)}.
\end{equation}
Thus, as $\sigma_{\min}(J_{y_C}c)\to 0$, the completion mapping can become arbitrarily sensitive: small perturbations in $y_F$ can induce large changes in $y_C$. This sensitivity is \emph{structural} in the sense that it is intrinsic to the completion parameterization whenever the chosen block $J_{y_C}c$ is poorly conditioned.

\paragraph{AdaNP projection steps.}
In AdaNP, each NP layer solves a linearly constrained projection subproblem that involves the matrix inverse $(BB^\top)^{-1}$ with $B=J_yc(x,\hat{y})$. Hence, per-step numerical sensitivity also increases when $\sigma_{\min}(B)$ is small (cf.\ Appendix~\ref{app:nonlinear_conditioning}). In contrast to predict-and-complete, this inverse appears inside a \emph{correction step} rather than defining a completion map. In practice, AdaNP admits mitigation mechanisms (e.g., depth adaptation, early stopping at $d_{\max}$, displacement regularization, and acceptance/rejection of projection steps during training), and its stated guarantees are local.

\paragraph{Backpropagation implication.}
In predict-and-complete layers, the network output $y_F$ is passed through an explicit completion map $y_C=\varphi_x(y_F)$. Consequently, backpropagation to the backbone necessarily includes the factor $J_{y_F}\varphi_x = -(J_{y_C}c)^{-1}J_{y_F}c$, so ill-conditioning of $J_{y_C}c$ can directly amplify or suppress gradients. In AdaNP, the backbone output $\hat y$ is instead transformed by one or more projection steps $\tilde y=\mathcal{P}(\hat y)$, and gradients backpropagate through $J_{\hat y}\mathcal{P}(\hat y)$. For affine constraint sets, $J_{\hat y}\mathcal{P}$ is an orthogonal projector with operator norm $\le 1$, hence the projection step cannot amplify gradients. For nonlinear constraints, no global non-expansiveness is claimed and the conditioning depends on the local Jacobian; in practice we mitigate this via depth adaptation and step acceptance during training.

\section{Implementation details}
\label{app:implementation_details}

\subsection{Batch local projection}
\label{app:batch_local_projection}
The computationally most expensive operation in the neural projection layer is the matrix inversion $(B B^T)^{-1}$ (Def.~\ref{def:NPlayer} in Section~\ref{subsec:adanp}), which has a complexity of $\mathcal{O}(N^3)$. At inference time, $N=N_C=N_{Eq} + N_{Ineq}$, since $B \in \mathbb{R}^{N_C \times N_O}$ and $B B^T \in \mathbb{R}^{N_C \times N_C}$. For moderate numbers of constraints (e.g., $N_C<10^3$), the cost of inverting the matrix $BB^\top \in \mathbb{R}^{N_C \times N_C}$ remains manageable on modern hardware.
During training, with batch size $BS$, this operation is performed independently for each sample, leading to an apparent cost of $\mathcal{O}(BS \times N_C^3)$. In practice, we implement these computations using batched linear algebra routines. Specifically, we construct a \textit{rank-3} tensor $\mathbf{B} \in \mathbb{R}^{BS \times N_C \times N_O}$ to store the local Jacobians, and a \textit{rank-2} tensor $V \in \mathbb{R}^{BS \times N_C}$ for the corresponding vectors $\mathbf{v}$.
Modern deep learning libraries support batched matrix factorizations and solves, allowing these operations to be executed in parallel across the batch dimension. As a result, the computational cost scales with $N_C^3$ per batch element, while benefiting from hardware parallelism in practice. To invert the batch of matrices, we use the Cholesky factorization algorithm~\cite{burden2005numerical}.\\
Given these conditions and the capabilities of current hardware, the neural projection operation remains computationally efficient even during training when the number of constraints is in the order of a few hundred. Moreover, the complexity of this method is \textit{equivalent} to other state-of-the-art methods such as DC3~\cite{Donti2021_DC3learningmethod} and KKTHardnet~\cite{Iftakher2025_PhysicsInformedNeural}, where each Newton's step in the completion algorithm requires the inversion of a batch of $(N_C \times N_C)$ matrices.

\subsection{Memory footprint}
\label{app:memory_footprint}
For each neural projection layer, the AdaNP module creates four 3-dimensional tensors:
\begin{itemize}
    \item Tensor $\mathbf{B}$ of shape $(BS, N_C, N_O)$
    \item Tensor $\mathbf{b}$ of shape $(BS, N_C, 1)$
    \item Tensor $\mathbf{B^*}$ of shape $(BS, N_O, N_O)$
    \item Tensor $\mathbf{b^*}$ of shape $(BS, N_O, 1)$
\end{itemize}

Assuming 32-bit floating-point representation (i.e., 4 bytes per element), the memory required to store these tensors for a single projection step is:
\begin{equation}
M_{\mathcal{P}} = 4 \cdot BS \cdot (N_O+1) \cdot (N_C + N_O) \quad \text{(bytes)}
\end{equation}

\begin{figure}[t]
    \centering
        \includegraphics[height=7cm]{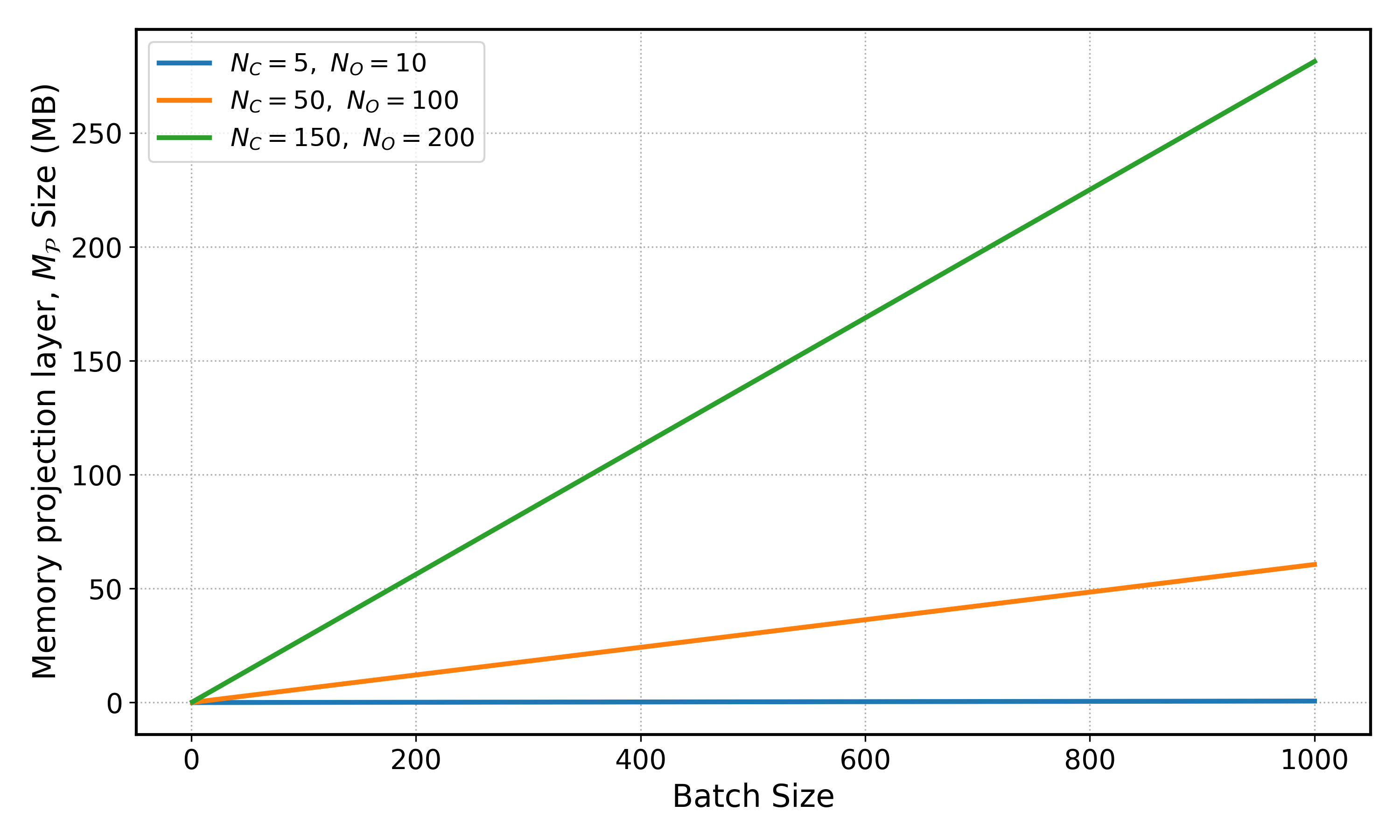}
        \caption{Memory footprint of a single projection layer $\mathcal{P}$. $N_C$ is the number of constraints, and $N_O$ is the dimension of the neural network output. The memory usage scales linearly with the batch size, with the growth rate determined by the number of predicted variables and constraints. For a batch size of 200, the memory requirement is on the order of tens of megabytes.}
        \label{fig:memory_P}
\end{figure}

\begin{figure}[t]
        \centering
        \includegraphics[height=7cm]{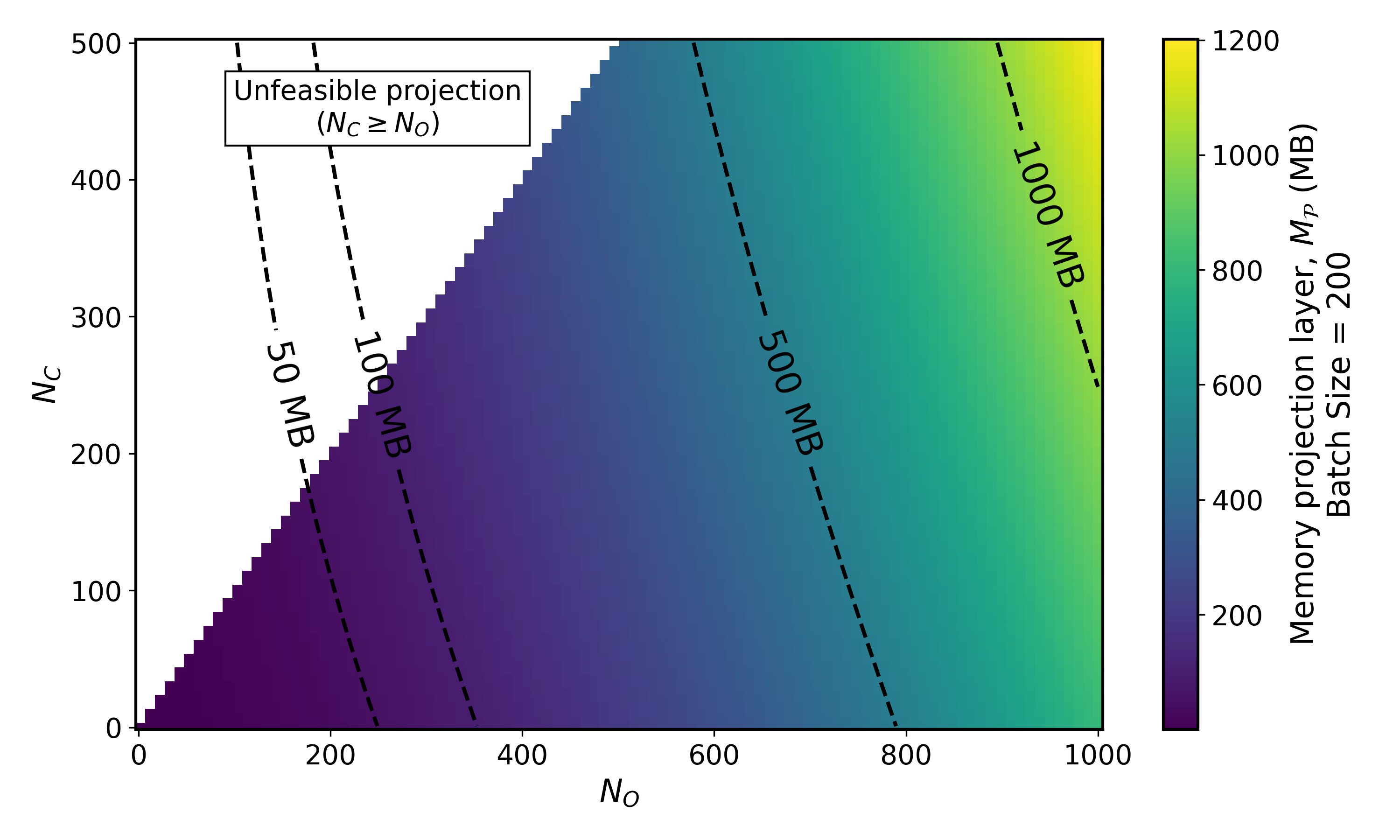}
        \caption{Memory footprint for a fixed batch size of 200 samples. $N_C$ is the number of constraints, and $N_O$ is the dimension of the neural network output. For large-scale tasks involving thousands of variables and constraints, the AdaNP module can become memory-intensive, with each projection layer requiring more than 1 GB of memory.}
        \label{fig:memory_heatmap}
\end{figure}

Figures~\ref{fig:memory_P} and~\ref{fig:memory_heatmap} illustrate how the memory requirement for a single projection layer is affected by variations in batch size, the number of predicted variables, and the number of constraints.

The total memory usage of the \textit{unrolled} AdaNP depends on the mode of operation:

\begin{itemize}
    \item During training, tensors are retained at each of the $n$ projection steps (i.e., for gradient computation), resulting in:
    \begin{equation}
    M_{\text{train}} = 4 \cdot n \cdot BS \cdot (N_O+1) \cdot (N_C + N_O) 
    \end{equation}

    \item During inference, tensors are not retained across steps, and the peak memory usage is:
    \begin{equation}
    M_{\text{infer}} = 4 \cdot BS \cdot (N_O+1) \cdot N_O
    \end{equation}
\end{itemize}

\begin{figure}[t]
    \centering
    \includegraphics[height=7cm]{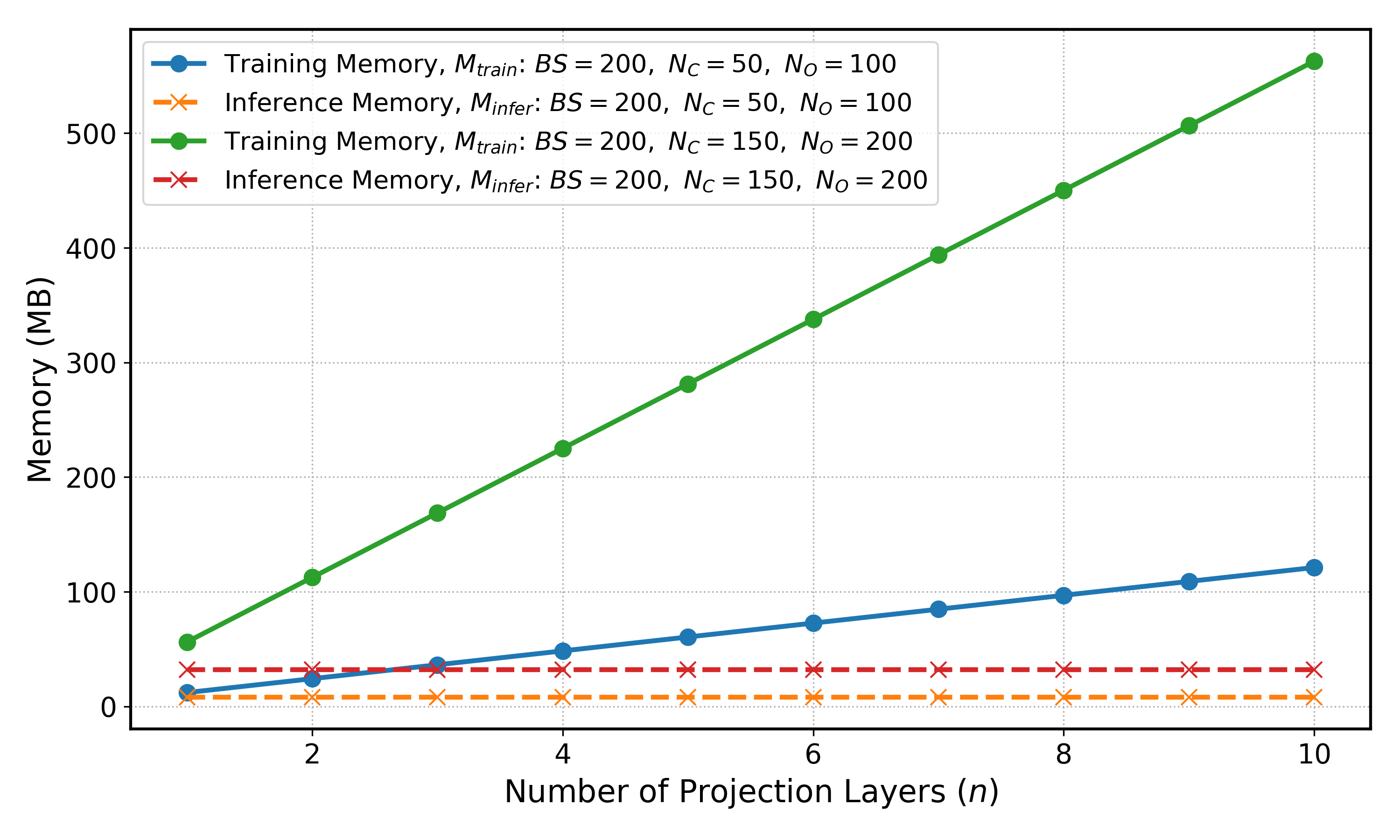}
    \caption{Memory usage during training and inference as a function of the number of projection layers $n$, for two problem scales. $N_C$ is the number of constraints, and $N_O$ is the dimension of the neural network output. Results are shown for a fixed batch size $BS = 200$. Smaller-scale setup: $N_C = 50$, $N_O = 100$; larger-scale setup: $N_C = 150$, $N_O = 200$. Training memory increases linearly with $n$, with significantly higher usage in larger-scale problems. Inference memory remains nearly constant across all configurations.}
    \label{fig:memory_AdaNP}
\end{figure}

\section{Additional experiments}
\label{app:additional_exp}

\subsection{Zero-set not locally \texorpdfstring{$C^{1,1}$}{C1,1}}
We provide an experiment involving a simple regression task with a constraint whose zero set is not locally $C^{1,1}$. Specifically, we consider regressing the functions $y_1 = \sin(x)$ and $y_2 = \sqrt[3]{\sin^2(x)}$, subject to the cusp constraint $c(y_1, y_2) = y_1^2 - y_2^3$. This constraint has a singular point at the origin, where the gradient is non-Lipschitz. We train the model on 100 data points sampled in $[-2, 2]$ and test on 100,000 points. ENFORCE achieves constraint violations below $10^{-6}$ across all test samples, including near the singularity.
This result suggests that, empirically, ENFORCE can handle constraints that are not locally $C^{1,1}$.\\

\subsection{Scaling analysis for nonconvex problems with linear and nonlinear constraints}
In this section, we present detailed results from the scaling analysis conducted on the two classes of optimization problems presented in Section~\ref{subsec:constr_opt_prob}: (i) nonconvex problems with linear equality constraints, and (ii) nonconvex problems with nonlinear equality constraints. The following tables report key performance metrics of ENFORCE across varying problem sizes, including different numbers of constraints and optimization variables, and compare them with alternative deep learning-based methods and a traditional large-scale nonlinear programming solver such as IPOPT. Table~\ref{tab:scaling_metrics_linear} and  Table~\ref{tab:nonconvex_nonlinear_comparison} report the resulting metrics for the linearly constrained case and the nonlinearly constrained case, respectively.

\begin{table}[t]
\centering
\caption{Scaling experiments on a nonconvex optimization problem with linear equality constraints (Eq.\eqref{eq:lin_opt_prob_learning}) evaluating performance across varying numbers of constraints ($N_C$) and variables ($N_O$). ENFORCE consistently predicts feasible and near-optimal solutions, outperforming alternative deep learning-based methods. DC3 is trained for a greater number of epochs than the other methods, until convergence is reached.}
\label{tab:scaling_metrics_linear}
\begin{tabular}{llccc}
\toprule
Constraints & ($N_C$)
  & \multicolumn{1}{c}{$50$}
  & \multicolumn{1}{c}{$70$}
  & \multicolumn{1}{c}{$150$} \\
Variables   & ($N_O$)
  & \multicolumn{1}{c}{$100$}
  & \multicolumn{1}{c}{$100$}
  & \multicolumn{1}{c}{$200$} \\
\midrule
Method      & Metric         &       &       &       \\
\midrule
IPOPT       & Obj. value     & $-11.11\pm0.00$ & $-4.84\pm0.00$  & $-10.64\pm0.00$ \\
            & Max eq.        & $0.00\pm0.00$   & $0.00\pm0.00$   & $0.00\pm0.00$   \\
            & Mean eq.       & $0.00\pm0.00$   & $0.00\pm0.00$   & $0.00\pm0.00$   \\
            & Inf. time [s]  & $0.095\pm0.033$ & $0.13\pm0.04$   & $0.379\pm0.060$ \\
            & Tr. time [min] & --              & --              & --              \\
            & Epochs         & --              & --              & --              \\
\midrule
MLP         & Obj. value     & $-27.43\pm0.00$ & $-27.43\pm0.00$ & $-52.99\pm0.01$ \\
            & Max eq.        & $24.65\pm0.08$  & $24.89\pm0.12$  & $45.38\pm0.56$  \\
            & Mean eq.       & $7.32\pm0.00$   & $7.19\pm0.00$   & $9.14\pm0.02$   \\
            & Inf. time [s]  & $0.001\pm0.000$ & $0.001\pm0.001$ & $0.001\pm0.001$ \\
            & Tr. time [min] & $8.87\pm0.18$   & $8.89\pm0.11$   & $9.01\pm0.25$   \\
            & Epochs         & $1000$          & $1000$          & $1000$          \\
\midrule
Soft        & Obj. value     & $-10.10\pm0.31$ & $-1.86\pm0.17$  & $1.28\pm0.32$   \\
($\lambda_c=5$) 
            & Max eq.        & $0.53\pm0.04$   & $0.79\pm0.08$   & $1.45\pm0.43$   \\
            & Mean eq.       & $0.03\pm0.00$   & $0.06\pm0.00$   & $0.08\pm0.00$   \\
            & Inf. time [s]  & $0.002\pm0.000$ & $0.001\pm0.000$ & $0.001\pm0.000$ \\
            & Tr. time [min] & $10.69\pm0.51$  & $10.72\pm0.43$  & $10.91\pm0.46$  \\
            & Epochs         & $1000$          & $1000$          & $1000$          \\
\midrule
Soft        & Obj. value     & $-10.69\pm0.01$ & $-4.18\pm0.03$  & $-8.18\pm0.18$  \\
($\lambda_c=1$) 
            & Max eq.        & $0.54\pm0.05$   & $0.86\pm0.05$   & $1.47\pm0.41$   \\
            & Mean eq.       & $0.05\pm0.00$   & $0.08\pm0.00$   & $0.09\pm0.00$   \\
            & Inf. time [s]  & $0.001\pm0.000$ & $0.001\pm0.000$ & $0.001\pm0.001$ \\
            & Tr. time [min] & $10.69\pm0.52$  & $10.70\pm0.46$  & $10.85\pm0.49$  \\
            & Epochs         & $1000$          & $1000$          & $1000$          \\
\midrule
Soft        & Obj. value     & $-12.05\pm0.00$ & $-6.82\pm0.01$  & $-13.55\pm0.02$ \\
($\lambda_c=0.1$)
            & Max eq.        & $2.09\pm0.03$   & $2.51\pm0.08$   & $2.17\pm0.12$   \\
            & Mean eq.       & $0.36\pm0.00$   & $0.43\pm0.00$   & $0.35\pm0.00$   \\
            & Inf. time [s]  & $0.001\pm0.001$ & $0.001\pm0.000$ & $0.001\pm0.001$ \\
            & Tr. time [min] & $10.62\pm0.61$  & $10.73\pm0.52$  & $10.88\pm0.54$  \\
            & Epochs         & $1000$          & $1000$          & $1000$          \\
\midrule
DC3         & Obj. value     & $-10.31\pm10.07$& $-2.76\pm0.06$  & $-6.27\pm0.07$  \\
            & Max eq.        & $0.00\pm0.00$   & $0.00\pm0.00$   & $0.00\pm0.00$   \\
            & Mean eq.       & $0.00\pm0.00$   & $0.00\pm0.00$   & $0.00\pm0.00$   \\
            & Inf. time [s]  & $0.003\pm0.000$ & $0.002\pm0.000$ & $0.004\pm0.000$ \\
            & Tr. time [min] & $22.96\pm3.73$  & $20.57\pm8.30$  & $25.18\pm8.63$  \\
            & Epochs         & $3500$          & $3500$          & $3500$          \\
\midrule
\textbf{ENFORCE}
            & Obj. value     & $-11.50\pm0.01$ & $-4.86\pm0.00$  & $-10.59\pm0.00$ \\
            & Max eq.        & $0.00\pm0.00$   & $0.00\pm0.00$   & $0.00\pm0.00$   \\
            & Mean eq.       & $0.00\pm0.00$   & $0.00\pm0.00$   & $0.00\pm0.00$   \\
            & Inf. time [s]  & $0.008\pm0.001$ & $0.010\pm0.001$ & $0.016\pm0.002$ \\
            & Tr. time [min] & $12.79\pm0.03$  & $12.72\pm0.04$  & $13.91\pm0.07$  \\
            & Epochs         & $1000$          & $1000$          & $1000$          \\
\bottomrule
\end{tabular}
\end{table}

\begin{table}[t]
\centering
\caption{Scaling experiments on a nonconvex optimization problem with nonlinear equality constraints (Eq.\eqref{eq:opt_prob_learning}) evaluating performance across varying numbers of constraints ($N_C$) and variables ($N_O$). ENFORCE consistently predicts feasible and near-optimal solutions. In the simplest setting, it outperforms the nonlinear programming solver IPOPT in terms of solution quality.}
\label{tab:scaling_metrics_nonlinear}
\begin{tabular}{llccccc}
\toprule
Constraints & ($N_C$)
  & \multicolumn{1}{c}{$10$}
  & \multicolumn{1}{c}{$30$}
  & \multicolumn{1}{c}{$50$}
  & \multicolumn{1}{c}{$70$}
  & \multicolumn{1}{c}{$150$} \\
Variables & ($N_O$)
  & \multicolumn{1}{c}{$100$}
  & \multicolumn{1}{c}{$100$}
  & \multicolumn{1}{c}{$100$}
  & \multicolumn{1}{c}{$100$}
  & \multicolumn{1}{c}{$200$} \\
\midrule
Method & Metric & & & & & \\
\midrule
IPOPT  & Obj. value           & $-26.27\pm0.00$ & $-21.81\pm0.00$ & $-18.05\pm0.00$ & $-11.69\pm0.00$ & $-29.45\pm0.00$ \\
       & Max Eq.              & $0.00\pm0.00$   & $0.00\pm0.00$   & $0.00\pm0.00$   & $0.00\pm0.00$   & $0.00\pm0.00$   \\
       & Avg. Eq.             & $0.00\pm0.00$   & $0.00\pm0.00$   & $0.00\pm0.00$   & $0.00\pm0.00$   & $0.00\pm0.00$   \\
       & Inf. time [s]        & $0.094\pm0.032$ & $0.244\pm0.132$ & $0.268\pm0.125$ & $0.401\pm0.167$ & $3.40\pm1.40$   \\
       & Tr. time [min]    & --              & --              & --              & --              & --              \\
       & Epochs               & --              & --              & --              & --              & --              \\
\midrule
MLP     & Obj. value           & $-27.43\pm0.00$ & $-27.43\pm0.00$ & $-27.43\pm0.00$ & $-27.43\pm0.00$ & $-53.07\pm0.00$ \\
        & Max eq.              & $214.95\pm0.10$ & $317.76\pm0.06$ & $317.14\pm0.01$ & $317.97\pm0.73$ & $497.38\pm4.64$ \\
        & Mean eq.             & $59.49\pm0.03$  & $72.10\pm0.01$  & $69.63\pm0.00$  & $70.57\pm0.01$  & $118.39\pm0.07$ \\
        & Inf. time [s]        & $0.001\pm0.001$ & $0.001\pm0.001$ & $0.001\pm0.001$ & $0.001\pm0.001$ & $0.002\pm0.001$ \\
        & Tr. time [min]    & $7.1\pm0.3$     & $7.8\pm0.0$     & $10.1\pm3.1$    & $9.4\pm0.3$     & $10.6\pm0.3$    \\
        & Epochs               & $1000$          & $1000$          & $1000$          & $1000$          & $1000$          \\
\midrule
Soft   & Obj. value           & $462.93\pm29.91$ 
                            & $>10^{3}$
                            & $>10^{5}$
                            & $>10^{5}$
                            & $>10^{5}$ \\
($\lambda_c=5$)
        & Max eq.              & $16.31\pm1.00$ & $71.27\pm3.58$ & $72.97\pm0.74$ & $73.63\pm4.56$ & $79.23\pm3.81$ \\
        & Mean eq.             & $2.31\pm0.10$  & $16.19\pm0.12$ & $16.81\pm0.03$ & $16.71\pm0.03$ & $16.72\pm0.07$ \\
        & Inf. time [s]        & $0.001\pm0.000$ & $0.001\pm0.000$ & $0.001\pm0.001$ & $0.002\pm0.001$ & $0.001\pm0.001$ \\
        & Tr. time [min]    & $12.4\pm0.1$   & $12.6\pm0.3$   & $12.8\pm0.6$   & $14.0\pm0.6$   & $15.1\pm0.7$   \\
        & Epochs               & $1000$          & $1000$          & $1000$          & $1000$          & $1000$          \\
\midrule
Soft   & Obj. value           & $61.93\pm3.31$
                            & $>10^{4}$
                            & $>10^{4}$
                            & $>10^{4}$
                            & $>10^{4}$ \\
($\lambda_c=1$)
        & Max eq.              & $15.67\pm1.43$ & $70.39\pm3.46$ & $72.18\pm2.21$ & $73.80\pm5.78$ & $79.30\pm3.70$ \\
        & Mean eq.             & $2.16\pm0.06$  & $16.21\pm0.14$ & $16.81\pm0.01$ & $16.71\pm0.04$ & $16.68\pm0.06$ \\
        & Inf. time [s]        & $0.001\pm0.000$ & $0.002\pm0.000$ & $0.002\pm0.001$ & $0.001\pm0.001$ & $0.001\pm0.000$ \\
        & Tr. time [min]    & $12.5\pm0.1$   & $12.6\pm0.3$   & $11.0\pm0.6$   & $13.9\pm0.6$   & $15.0\pm0.6$   \\
        & Epochs               & $1000$          & $1000$          & $1000$          & $1000$          & $1000$          \\
\midrule
Soft   & Obj. value           & $-18.29\pm1.21$
                            & $>10^{3}$
                            & $>10^{3}$
                            & $>10^{3}$
                            & $>10^{3}$ \\
($\lambda_c=0.1$)
        & Max eq.              & $16.18\pm0.55$ & $70.37\pm5.08$ & $76.56\pm2.35$ & $75.36\pm2.70$ & $78.35\pm1.95$ \\
        & Mean eq.             & $2.05\pm0.17$  & $16.35\pm0.08$ & $16.86\pm0.05$ & $16.77\pm0.07$ & $16.63\pm0.01$ \\
        & Inf. time [s]        & $0.001\pm0.000$ & $0.001\pm0.000$ & $0.002\pm0.001$ & $0.001\pm0.001$ & $0.002\pm0.001$ \\
        & Tr. time [min]    & $11.5\pm0.1$   & $12.5\pm0.4$   & $11.9\pm0.1$   & $13.9\pm0.7$   & $14.9\pm0.8$   \\
        & Epochs               & $1000$          & $1000$          & $1000$          & $1000$          & $1000$          \\
\midrule
\textbf{ENFORCE} & Obj. value           & $-26.37\pm0.00$ & $-21.48\pm0.01$ & $-16.68\pm0.01$ & $-7.75\pm0.03$  & $-27.77\pm0.02$ \\
        & Max eq.              & $0.00\pm0.00$   & $0.00\pm0.00$   & $0.00\pm0.00$   & $0.00\pm0.00$   & $0.00\pm0.00$   \\
        & Mean eq.             & $0.00\pm0.00$   & $0.00\pm0.00$   & $0.00\pm0.00$   & $0.00\pm0.00$   & $0.00\pm0.00$   \\
        & Inf. time [s]        & $0.013\pm0.002$ & $0.023\pm0.002$ & $0.030\pm0.005$ & $0.049\pm0.009$ & $0.14\pm0.08$   \\
        & Tr. time [min]    & $25.3\pm0.1$    & $29.4\pm0.1$    & $35.8\pm0.4$    & $49.0\pm0.9$    & $69.4\pm23.1$   \\
        & Epochs               & $1000$          & $1000$          & $1000$          & $1000$          & $1000$          \\
\bottomrule
\end{tabular}
\end{table}

\end{document}